%% file: 0-master.tex
\documentclass[twoside,11pt]{article}

\usepackage[T1]{fontenc}
\usepackage{wasysym}
\usepackage{cite}
\usepackage[bottom]{footmisc}
\usepackage{times,latexsym}
\usepackage[hyphens]{url}
\usepackage[T1]{fontenc}
\usepackage{tikz}
\usepackage{svg}
\usepackage{multicol}
\usepackage{enumitem}
\usepackage{booktabs}
\usepackage{amssymb}
\usepackage{array}\newcolumntype{P}[1]{>{\centering\arraybackslash}p{#1}}
\usepackage{makecell}
\usepackage{moresize}
\usepackage{appendix}
\usepackage{refstyle}\usepackage{float}
\usepackage{jair, theapa, rawfonts}

\definecolor{pink}{rgb}{0.858, 0.188, 0.478}

\ShortHeadings{Recent Advances in Natural Language Inference: A Survey}
{Storks, Gao, \& Chai}
\firstpageno{1}

\begin{document}

\title{Recent Advances in Natural Language Inference:\\ A Survey of Benchmarks, Resources, and Approaches}
\author{\name Shane Storks \email sstorks@umich.edu \\
       \addr Computer Science and Engineering\\
       University of Michigan\\
       \vspace{5pt}
       Ann Arbor, MI 48109-2121 USA\\
       \name Qiaozi Gao \email gaoqiaoz@msu.edu \\
       \addr Computer Science and Engineering\\
       Michigan State University\\
       \vspace{5pt}
       East Lansing, MI 48824 USA\\
       \name Joyce Y. Chai \email chaijy@umich.edu \\
       \addr Computer Science and Engineering\\
       University of Michigan\\
       Ann Arbor, MI 48109-2121 USA}

\maketitle
\begin{abstract}
In the NLP community, recent years have seen a surge of research activities that address machines' ability to perform deep language understanding which goes beyond what is explicitly stated in text, rather relying on reasoning and knowledge of the world. 
Many benchmark tasks and datasets have been created to support the development and evaluation of such natural language inference ability. 
As these benchmarks become instrumental and a driving force for the NLP research community, this paper aims to provide an overview of recent benchmarks, relevant knowledge resources, and state-of-the-art learning and inference approaches in order to support a better understanding of this growing field. 
\end{abstract}

\input{1-intro.tex}
\input{2-benchmarks.tex}
\input{3-resources.tex}
\input{4-approaches.tex}

\input{5-otherbenchmark.tex}
\input{6-discussion.tex}
\input{7-acknowledgements.tex}
\input{8-appendix.tex}

\bibliographystyle{theapa}
\bibliography{main}
\end{document}

%% file: 1-intro.tex
\section{Introduction}\label{intro}

\textcolor{black}{We humans use a variety of knowledge and reasoning to help understand meanings of language. For example, consider these sentences from \citeA{minskyCommonsensebasedInterfaces2000}: ``Jack needed some money, so he went and shook his piggy bank. He was disappointed when it made no sound.'' From this, it is not difficult for us to understand that Jack did not find any money, and because of that, Jack was having a negative emotion. What made us come to this conclusion, which was not explicitly stated in the passage, is the knowledge we have about the world 
and the underlying reasoning process, often called \textit{commonsense thought} \cite{minskyCommonsensebasedInterfaces2000} or \textit{commonsense reasoning} \cite{davisCommonsenseReasoningCommonsense2015}, that allows us to connect pieces of knowledge to reach the new conclusion. We know that a \textit{piggy bank} is a container that holds coins (not pigs) and that coins are money which are made of metal. Since metal is a hard solid, the coins will make a sound when shaken inside of a container such as a piggy bank; if there is no sound, then there are no coins. It is also likely that we can predict that as piggy banks are typically possessed by children, there is a good chance that Jack is a child. Alternatively, these predictions may be derived from similar events we have experienced as children, and allow us to make similar conclusions by analogy \cite{minskyCommonsensebasedInterfaces2000}.
While this kind of knowledge and reasoning comes so naturally to human readers, it is notoriously difficult for machines. Despite significant advances in natural language processing in the last several decades, machines are still far away from having this type of {\em natural language inference (NLI)} ability. }

\begin{figure}
  \centering
  \includegraphics[width=1\textwidth]{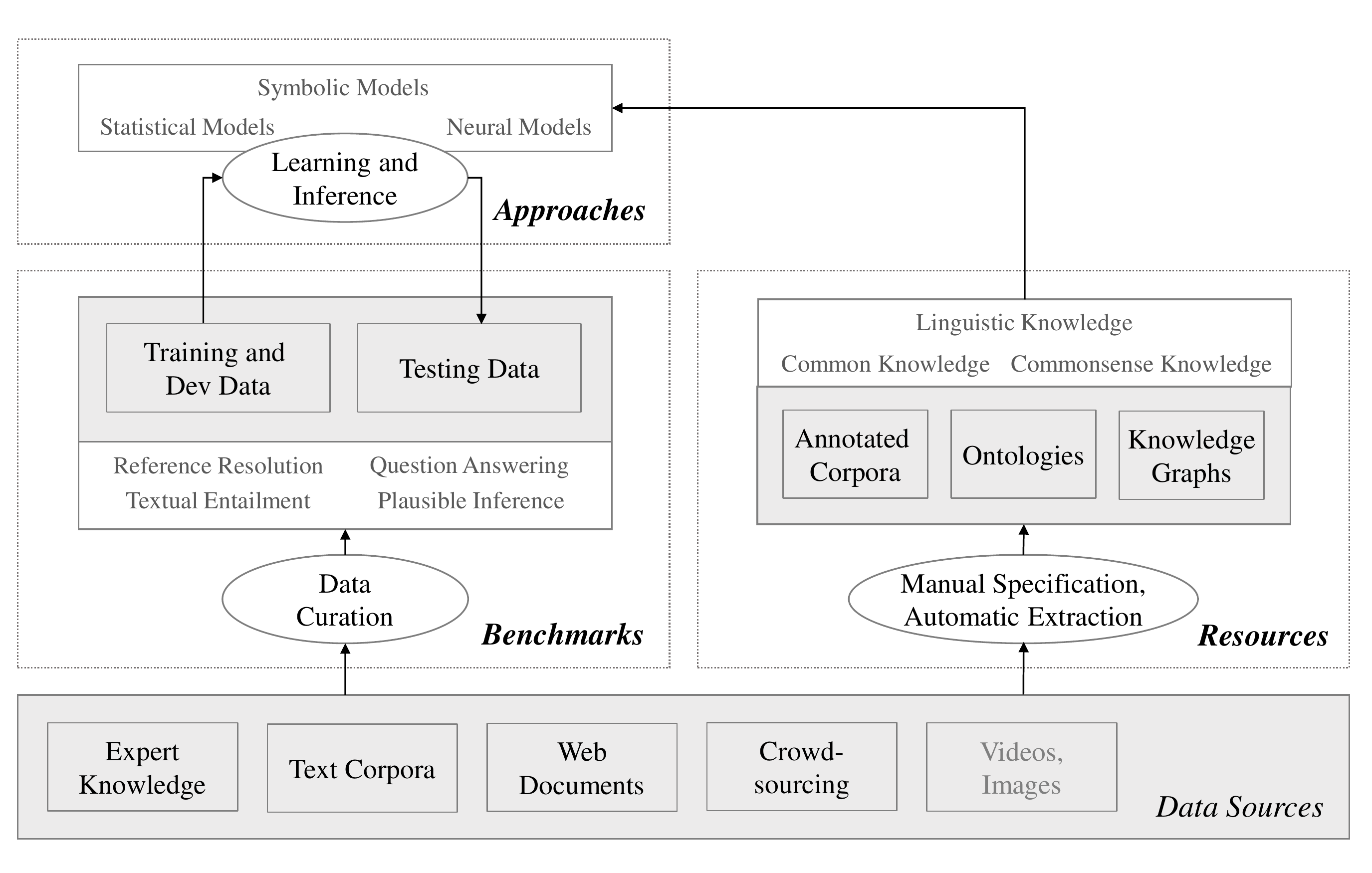}
  \caption[Main research efforts in natural language inference from the NLP community occur in three areas: benchmarks and tasks, knowledge resources, and learning and inference approaches.]{Main research efforts in natural language inference from the NLP community occur in three areas: benchmarks and tasks, knowledge resources, and learning and inference approaches.\footnotemark }
  \label{fig:overall}
\end{figure}
\footnotetext{Although videos and images are often used for creating various benchmarks, they are not the focus of this paper.}

To address this problem, recent years have seen a surge of research activities on NLI: {\em machines' capability of deep understanding of language that goes beyond what is explicitly expressed, rather relying on new conclusions inferred from knowledge about how the world works}.\footnote{While NLI is often a term used to refer to textual entailment tasks, e.g., by \citeA{bowmanLargeAnnotatedCorpus2015}, we use the term here to refer to a broader sense of inference which includes but is not restricted to the textual entailment task formulation.}
Many benchmark datasets have been created to help develop and evaluate NLI algorithms and models. These datasets have drawn significant attention from the research community, and many learning and inference approaches have been developed. To facilitate quantitative evaluation and encourage broader participation, various leaderboards for benchmarks have been set up and maintained. As these benchmarks become instrumental and a driving force for the community, this paper aims to provide an overview of recent advances in NLI focusing on existing tasks and benchmarks, knowledge resources, and learning and inference approaches, and discuss current limitations and future opportunities. We hope this paper will provide an entry point to those who are not familiar with but interested in pursuing research in this quickly evolving topic area.

\textcolor{black}{The NLP community has a long history of creating benchmarks to facilitate algorithm development and evaluation for various language processing tasks. Many resources to support those approaches to the benchmarks have also been proposed. It is neither possible nor intended for this paper to provide a comprehensive discussion of all benchmarks and resources proposed by the NLP community. This survey focuses on the ongoing research effort in benchmarks, resources, and approaches specifically toward natural language inference (NLI), as shown in Figure~\ref{fig:overall}.} 

\subsection{Scope of Benchmark Datasets}

\textcolor{black}{There has been a long line of work toward creating benchmark datasets to evaluate machines' progress on language processing tasks, and facilitate development of new approaches to the tasks. Many early efforts focused on collecting large-scale annotations to support data-driven approaches for low-level tasks such as POS tagging \cite{marcusBuildingLargeAnnotated1993} and named entity recognition \cite{grishmanMessageUnderstandingConference61996}. Other early benchmarks, such as the Message Understanding Conference (MUC) benchmarks for information extraction \cite{Sundheim:1993:TIE:1072017.1072023} and the automatic content extraction (ACE) benchmarks for coreference resolution and relation extraction~\cite{doddington-etal-2004-automatic}, were mainly designed to explore the information found in linguistic knowledge and context.
In part motivated by earlier works from \citeA{mccarthyExampleNaturalLanguage1976} and \citeA{lehnertHumanComputationalQuestion1977}, other benchmarks were created to focus on reading comprehension through tasks like question answering \cite{hirschmanDeepReadReading1999,voorheesTREC8QuestionAnswering2000}. These early reading comprehension benchmarks often required only one sentence of context at a time to answer questions\cite{kociskyNarrativeQAReadingComprehension2018}, possible through shallow linguistic approaches. Although it may be the case that some type of outside knowledge or reasoning may be required to reach an oracle performance, as all of these earlier benchmarks primarily targeted approaches that only applied linguistic context, they are outside the scope of this survey.}

\textcolor{black}{Recent years have seen a shift in benchmark tasks which are beyond the use of linguistic context, aiming to require a deeper understanding to solve the tasks. For instance, consider this question from the Winograd Schema Challenge \cite{levesqueWinogradSchemaChallenge2011}:``The trophy would not fit in the brown suitcase because it was too big. What was too big?'' To answer this question, linguistic constraints will not be able to resolve whether \textit{it} refers to \textit{the trophy} or \textit{the brown suitcase}. Only based on external knowledge (i.e., the fact that an object must be bigger than another object in order to contain it) is it possible to resolve the pronoun \textit{it} and answer the question correctly. 
Higher-level tasks, such as question answering, may appear the same as in earlier benchmarks. However, successfully solving these tasks will also require machines to go beyond linguistic context, and rely on reasoning and knowledge that is not explicitly stated in text. In this survey, we attempt to provide an overview of recent benchmarks like these.}  

These benchmarks vary in terms of the scope of tasks, for example, from the specific and focused reference resolution task in the Winograd Schema Challenge \cite{levesqueWinogradSchemaChallenge2011} to broader tasks such as textual entailment, e.g., the RTE Challenges \cite{daganPASCALRecognisingTextual2005}. While some benchmarks only focus on one specific task, others are comprised of a variety of tasks, e.g., GLUE~\cite{wangGLUEMultiTaskBenchmark2018}. Some benchmarks target a specific type of knowledge, e.g., commonsense social psychology in Event2Mind \cite{rashkinEvent2MindCommonsenseInference2018}, while others intend to address a variety of types of knowledge, e.g., the Story Cloze Test \cite{mostafazadehCorpusClozeEvaluation2016}, which requires a wide range of facts about everyday life. How benchmark tasks are formulated also differs among existing benchmarks. Some are in the form of multiple-choice questions, requiring a binary decision or a selection from a candidate list, while others are more open-ended. Different characteristics of benchmarks serve different goals, and a critical question is what criteria to consider in order to create a benchmark that can support technology development and measurable outcome of research progress on deep reasoning abilities. 

Section~\ref{sec:benchmarks} gives a detailed account of existing benchmarks and their common characteristics. It also summarizes important criteria to consider in building these benchmarks, including factors to consider in task formulation, techniques to avoid problematic data bias, data collection methods, and factors affecting overall benchmark complexity.

\subsection{Scope of Knowledge Resources}\label{sec:knowledgereasoning}

\textcolor{black}{Humans perform natural language inference based on a vast amount of external knowledge about language and the world. To support machines' inference ability, compiled knowledge resources become important. It is an unsolved problem to comprehensively taxonomize all of the human knowledge and reasoning required to perform general inference \cite{davisCommonsenseReasoningCommonsense2015}. For the purpose of this survey, though, we attempt to provide an overview of three types of knowledge resources: {\em linguistic knowledge}, {\em common knowledge}, and {\em commonsense knowledge}. Although modern approaches often rely on large text corpora to implicitly incorporate external knowledge without regard to what types of knowledge are required, we think it would be helpful to distinguish different types of knowledge to facilitate better understanding of the nature of NLI tasks.}

\paragraph{Linguistic knowledge.}
\textcolor{black}{In order to comprehend human language, machines first need \textit{linguistic knowledge}, i.e., knowledge about the language. This includes an understanding of word meanings, grammar, syntax, semantics, and discourse structure. Having linguistic knowledge gives a human or machine the basic capabilities of understanding language, and is a required property of virtually any NLP system, even those not created for NLI tasks. }

\paragraph{Common knowledge.}
\textcolor{black}{\textit{Common knowledge} refers to well-known facts about the world that are often explicitly stated, e.g., "canine distemper is a domestic animal disease" \cite{cambriaIsanetteCommonCommon2011a}. 
This kind of knowledge is often referred to in human communication \cite{cambriaIsanetteCommonCommon2011a}.
Some types of common knowledge may be domain-specific, e.g., the knowledge required to answer questions on an elementary-level science test \cite{sugawaraPrerequisiteSkillsReading2017}. While domain-specific knowledge is obviously useful for domain-specific applications, much of this knowledge may not be needed for general-purpose communication with humans.}

\paragraph{Commonsense knowledge.}
\textcolor{black}{\textit{Commonsense knowledge}, on the other hand, is typically unstated, as it is considered obvious to most humans \cite{cambriaIsanetteCommonCommon2011a}, and consists of universally accepted beliefs about the world \cite{nunbergPositionPaperCommonsense1987}. \citeA{davisCommonsenseReasoningCommonsense2015} demonstrate this: ``if you see a six-foot-tall person holding a two-foot-tall person in his arms, and you are told they are father and son, you do not have to ask which is which.'' Commonsense knowledge provides a deeper understanding of language. While it is rarely referred to in language, humans rely on it in communication \cite{cambriaIsanetteCommonCommon2011a}, as it is required to reach a common ground \cite{chaiLanguageActionInteractive2018}. Commonsense knowledge consists of everyday assumptions about the world \cite{zangSurveyCommonsenseKnowledge2013}, and is generally learned through one's own experience with the world, but can also be inferred by generalizing over common knowledge \cite{speerAnalogySpaceReducingDimensionality2008}. While common knowledge can vary by region, culture, and other factors, we expect that commonsense knowledge should be roughly typical to all humans \cite{davisLogicalFormalizationsCommonsense2017}.}

\textcolor{black}{Two particularly important domains of commonsense knowledge are intuitive physics, i.e., humans' basic understanding of physical interactions in the world, and intuitive psychology, i.e., humans' basic understanding of human emotion, behavior, and motives \cite{darpaMachineCommonSense2018}. Also respectively referred to as na{\"i}ve physics and psychology \cite{davisCommonsenseReasoningCommonsense2015} or physical and social commonsense \cite{sakaguchi2019winogrande}, these are considered fundamental as they have been shown to develop in humans as infants, even before language develops \cite{baillargeon2004infants,songCanInfantsAttribute2005}, and humans are capable of several kinds of intuitive physical and psychological inference at only eighteen months of age \cite{darpaMachineCommonSense2018}. These types of knowledge allow us to make on-the-fly predictions about everyday physical and social interactions. Intuitive physics allows us to infer, for example, the potential consequences when dropping a glass of water on the floor, i.e., the glass may shatter, water will almost certainly spill on the floor, and the floor could become slippery. From intuitive psychology, we can further infer that dropping the glass was likely an accident, and the person who dropped the glass will be angry, upset, or embarrassed.}

Section~\ref{sec:resources} gives an introduction to existing knowledge resources, and several recent efforts in building such resources to facilitate natural language inference.

\subsection{Scope of Approaches}
To tackle the challenging benchmark tasks, many computational models have been developed. These range from earlier symbolic and statistical approaches to recent approaches based on deep neural networks. Especially for recent neural approaches, various architectures have been proposed that model context of language, take advantage of external data or knowledge resources, and achieve the state-of-the-art performance, and in some cases, achieve near or above human performance. 
\textcolor{black}{However, due to the unexplainability of these recent approaches, as well as statistical biases recently found in benchmark datasets \cite{schwartzEffectDifferentWriting2017,gururanganAnnotationArtifactsNatural2018,nivenProbingNeuralNetwork2019}, it remains a subject of debate 
in terms of how much progress we have made in enabling natural language inference ability. }
Section~\ref{sec:approaches} summarizes recent learning and inference approaches, and discusses their performance, limitations, and trends.

%% file: 2-benchmarks.tex
\section{Benchmarks and Tasks}\label{sec:benchmarks}

\textcolor{black}{Alan Turing perhaps is the first one who proposed an experiment, widely-known as the Turing Test \cite{turingComputingMachineryIntelligence1950}, to evaluate whether a machine possessed intelligence. The formulation of the Turing Test, however, has been criticized for encouraging machines to deceive humans \cite{levesqueWinogradSchemaChallenge2011}, not providing feedback on a continuous scale to allow for incremental development \cite{ortizWhyWeNeed2016}, and being impractical in several ways \cite{harnadMindsMachinesTuring2000}. Benchmark datasets have thus been used as alternatives to the Turing Test, providing training and testing data, evaluation frameworks, and continuous numerical feedback for various language tasks.}
For the remainder of this section, we first give an overview of recent benchmarks that are designed to evaluate machine's inference ability in natural language understanding. Then, we summarize important considerations in creating these benchmarks and lessons learned from ongoing research.

\subsection{An Overview of Existing Benchmarks}\label{sec:existing benchmarks}
\begin{figure}
  \centering
  \includegraphics[width=1\textwidth]{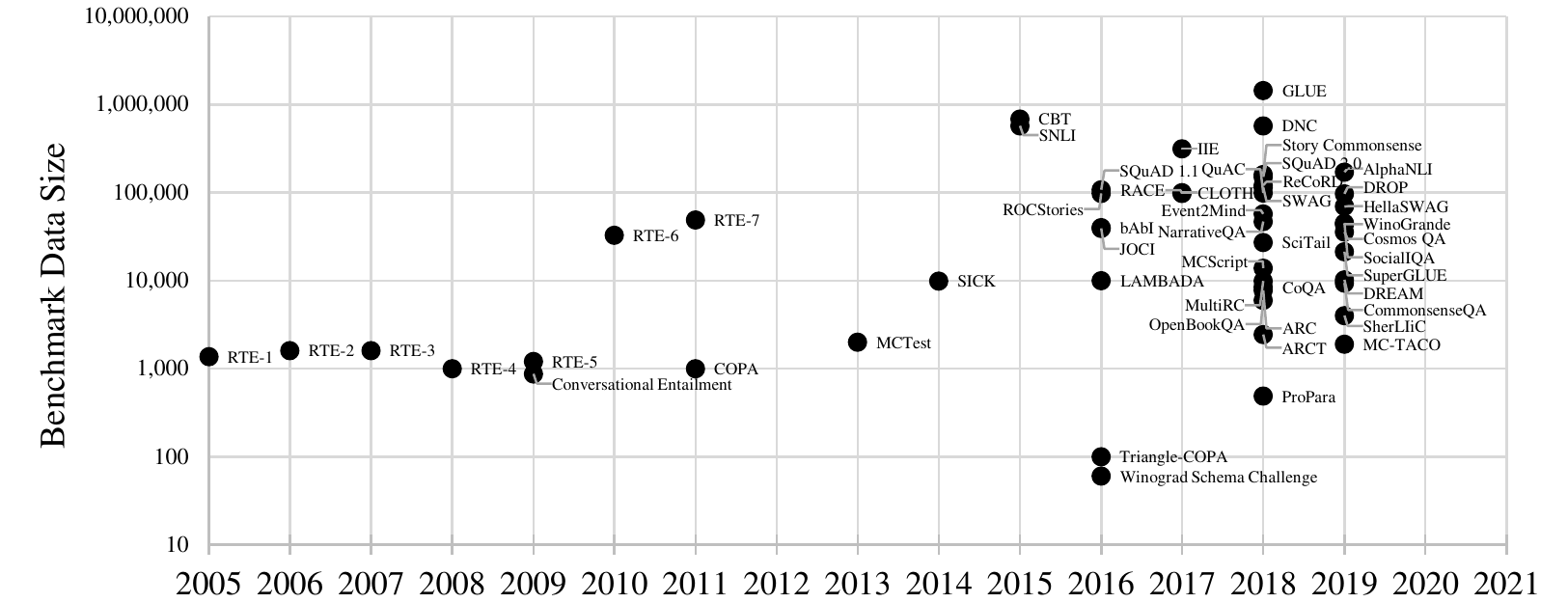}
  \caption{Since the early 2000s, there has been a surge of benchmark tasks geared toward natural language inference. In 2018, we saw the creation of more benchmarks of larger sizes than ever before.}
  \label{fig:timeline}
\end{figure}

Since the Recognizing Textual Entailment (RTE) Challenges introduced by \citeA{daganPASCALRecognisingTextual2005} in the early 2000s, there has been an explosion of challenging NLI benchmarks being created. Figure~\ref{fig:timeline} shows a trend of this growth among the benchmarks introduced in this section. The RTE challenges, which were inspired by difficulties encountered in semantic processing for machine translation, question answering, and information extraction \cite{daganPASCALRecognisingTextual2005}, had been the dominant reasoning tasks for many years. Recently, however, there has been a surge in the variety of benchmarks. 
As shown in Figure~\ref{fig:timeline}, the sizes for most benchmarks in earlier days, e.g., from 2005 to 2015, are relatively small. Most benchmarks developed after 2015 have a much larger number of instances. Some of them have more than 100,000 instances, which make the application of deep learning approaches possible. 

Next, we give a review of widely used benchmarks, introduced by the following groupings: reference resolution, question answering, textual entailment, plausible inference, psychological reasoning, and multiple tasks. These groupings are not necessarily exclusive, but they highlight the recent trends in the field. While all benchmarks' training and/or development data are available for free download, it is important to note that test data are often not distributed publicly so that testing of systems can occur privately in a standard, unbiased way.

\subsubsection{Reference Resolution}
Reference resolution is the process of identifying a referent, typically a linguistic mention in a span of text, that a particular expression, e.g., a pronoun or phrase, refers to. 
This process can be significantly complicated by ambiguities which arise from the presence of multiple entities in a sentence with pronouns, creating a need for external knowledge, e.g., commonsense knowledge, to inform decisions \cite{levesqueWinogradSchemaChallenge2011}. 
Since such challenging examples for reference resolution are typically handcrafted or handpicked \cite{morgensternPlanningExecutingEvaluating2016,rahmanResolvingComplexCases2012,gender-bias-in-coreference-resolution}, there exists a small magnitude of data for this skill, and it remains an unsolved problem \cite{rahmanResolvingComplexCases2012,davisFirstWinogradSchema2017}. Consequently, there is a need for more benchmarks here. In the following paragraphs, we introduce the few benchmarks that currently exist for reference resolution.

\paragraph{Winograd Schema Challenge.}
The classic reference resolution benchmark is the Winograd Schema Challenge (WSC), inspired by \citeA{winogradUnderstandingNaturalLanguage1972}, originally proposed by \citeA{levesqueWinogradSchemaChallenge2011}, later developed by \citeA{levesqueWinogradSchemaChallenge2012}, and actualized by \citeA{morgensternWinogradSchemaChallenge2015} and \citeA{morgensternPlanningExecutingEvaluating2016}. In this challenge, systems are presented with questions about sentences known as Winograd schemas. To answer a question, a system must disambiguate a pronoun whose referent may be one of two entities, and can be changed by replacing a single word in the sentence. Consequently, linguistic context is typically not useful in disambiguating the pronoun, and outside knowledge is required. The first Challenge dataset consisted of just 60 testing examples \cite{morgensternPronounDisambiguationProblems2016}, but there are more available online.\footnote{See \url{https://cs.nyu.edu/faculty/davise/papers/WinogradSchemas/WS.html}, \\\url{https://commonsensereasoning.org/winograd.html}, and \\\url{https://commonsensereasoning.org/disambiguation.html}.} Additionally, \citeA{rahmanResolvingComplexCases2012} curated nearly 1,000 similar pronoun resolution problems to aid in training systems for the WSC; this set of problems is included within a later-mentioned dataset, and thus is not introduced separately. Further information about the Winograd Schema Challenge is available at \url{https://commonsensereasoning.org/winograd.html}.

\paragraph{WinoGrande.}
\textcolor{black}{Inspired by biases found in the original Winograd Schema Challenge \cite{trichelair2018reasonable}, \citeA{sakaguchi2019winogrande} created WinoGrande, a large-scale, adversarial version of the problem focusing primarily on intuitive physics and psychology. It consists of nearly 44,000 examples, but maintains the same level of difficulty, having low baseline performance but high human performance. This is the first high-quality, large-scale dataset of adversarial Winograd schemas, and should encourage the application of deep neural networks to the problem to improve machine performance on reference resolution tasks. WinoGrande can be downloaded at \url{https://mosaic.allenai.org/projects/winogrande}.}

\paragraph{Other reference resolution benchmarks.}
Other reference resolution benchmarks include the basic and compound coreference subtasks within bAbI \cite{westonAICompleteQuestionAnswering2016}, the definite pronoun resolution problems by \citeA{rahmanResolvingComplexCases2012} within Inference is Everything \cite{whiteInferenceEverythingRecasting2017}, the Winograd NLI subtask within GLUE \cite{wangGLUEMultiTaskBenchmark2018}, and the gendered anaphora-based Winogender task originally by \citeA{gender-bias-in-coreference-resolution} within DNC \cite{poliak2018emnlp-DNC}. The full benchmarks to which these subtasks belong are introduced in Section~\ref{sec:multiple tasks}, while in Figure~\ref{fig:coreference examples}, we list several examples of challenging reference resolution problems across the existing benchmarks, where external knowledge, such as the knowledge that predators eat their prey, is useful in disambiguating pronouns. 

\textcolor{black}{It is important to note that all of these benchmarks evaluate the resolution of entity references rather than event references. Benchmarks toward event coreference are limited, and those which do exist typically require only linguistic context to disambiguate event references \cite{doddington-etal-2004-automatic,bejanLinguisticResourceDiscovering2008}. On the other hand, since the linguistic structures which refer to specific events are much more flexible than those for entities \cite{chenGraphbasedEventCoreference2009}, disambiguating event references may be more challenging than the existing benchmarks for entity reference resolution. As more progress is made, we anticipate event reference resolution benchmarks which require outside knowledge and reasoning will be created in the near future.}

\begin{figure}
\noindent\hrulefill
\begin{multicols}{2}
\begin{enumerate}[label=\textbf{(\Alph*)}]
\item \textbf{Winograd Schema Challenge \cite{levesqueWinogradSchemaChallenge2011}} \newline
The trophy would not fit in the brown suitcase because it was too big. What was too big?
\begin{enumerate}[label=\alph*.]
    \item \textbf{The trophy}
    \item The suitcase
\end{enumerate}

The trophy would not fit in the brown suitcase because it was too small. What was too small?
\begin{enumerate}[label=\alph*.]
    \item The trophy
    \item \textbf{The suitcase}
\end{enumerate}

\item \textbf{Winogender \cite{gender-bias-in-coreference-resolution}} \newline
The paramedic performed CPR on the passenger even though she knew it was too late. Who knew it was too late?
\begin{enumerate}[label=\alph*.]
    \item \textbf{The paramedic}
    \item The passenger
\end{enumerate}

\item \textbf{\cite{rahmanResolvingComplexCases2012}} \newline
Lions eat zebras because they are predators. Who are predators?
\begin{enumerate}[label=\alph*.]
    \item \textbf{Lions}
    \item Zebras
\end{enumerate}
\end{enumerate}
\end{multicols}
\noindent\hrulefill
\caption{Examples from existing reference resolution benchmark tasks. Answers in bold.}
\label{fig:coreference examples}
\end{figure}

\subsubsection{Question Answering}\label{sec:qa}
Instead of providing a focused language processing task like reference resolution, many benchmarks instead provide a more comprehensive mix of language processing and reasoning skills within a single task.
\textcolor{black}{Question answering (QA) is one such comprehensive task, particularly the recent formulation of the task providing a passage, and requiring a system to answer questions about it to demonstrate its comprehension of the passage (also referred to as machine reading comprehension). Today, QA is a fairly well-established area in NLP, and there are many existing benchmarks for QA. However, not all of them require external knowledge and reasoning to solve \cite{ostermannMCScriptNovelDataset2018}, so many can be solved only based on linguistic context without a deep understanding.} Some QA benchmarks that do require external knowledge are MCScript \cite{ostermannMCScriptNovelDataset2018}, CoQA \cite{reddyCoQAConversationalQuestion2018}, and OpenBookQA \cite{mihaylovCanSuitArmor2018}.
Examples from these benchmarks are listed in Figure~\ref{fig:qa examples}, where the commonsense knowledge that diapers are typically thrown away, and the common knowledge that steel is a conductor and that Democrats and Republicans compete in American elections, is particularly useful in answering questions. Other QA benchmarks require advanced reasoning processes beyond the text. 
SQuAD 2.0 is one such example, as it includes unanswerable questions about passages \cite{rajpurkarKnowWhatYou2018}, which require a deep reasoning over all knowledge given in the passage to identify. In the following paragraphs, we introduce all surveyed QA benchmarks.

\begin{figure}
\noindent\hrulefill
\begin{multicols}{2}
\begin{enumerate}[label=\textbf{(\Alph*)}]
\item \textbf{MCScript \cite{ostermannMCScriptNovelDataset2018}} \newline
Did they throw away the old diaper?
\begin{enumerate}[label=\alph*.]
\item \textbf{Yes, they put it into the bin.}
\item No, they kept it for a while.
\end{enumerate}

\item \textbf{OpenBookQA \cite{mihaylovCanSuitArmor2018}} \newline
Which of these would let the most heat travel through?
\begin{enumerate}[label=\alph*.]
\item a new pair of jeans.
\item \textbf{a steel spoon in a cafeteria.}
\item a cotton candy at a store.
\item a calvin klein cotton hat.
\end{enumerate}

\textit{Evidence:} Metal is a thermal conductor.
\newline\newline
\item \textbf{CoQA \cite{reddyCoQAConversationalQuestion2018}} \newline
The Virginia governor's race, billed as the marquee battle of an otherwise anticlimactic 2013 election cycle, is shaping up to be a foregone conclusion. Democrat Terry McAuliffe, the longtime political fixer and moneyman, hasn't trailed in a poll since May. Barring a political miracle, Republican Ken Cuccinelli will be delivering a concession speech on Tuesday evening in Richmond. In recent ...

Who is the democratic candidate?\newline
\textbf{Terry McAuliffe}\newline
\textit{Evidence}: Democrat Terry McAuliffe

Who is his opponent?\newline
\textbf{Ken Cuccinelli}\newline
\textit{Evidence}: Republican Ken Cuccinelli

\end{enumerate}
\end{multicols}
\noindent\hrulefill
\caption{Examples from QA benchmarks which require inference through outside knowledge. Answers in bold.}
\label{fig:qa examples}
\end{figure}

\paragraph{MCTest.}
\textcolor{black}{MCTest by \citeA{richardsonMCTestChallengeDataset2013} is a collection of 500 fictional stories and 2,000 multiple-choice questions which require outside knowledge and various types of reasoning to answer. 
More than half of the questions require more than one sentence to answer, so that systems must synthesize information from multiple locations rather than using shallow matching techniques to find the answer. The questions are open-domain with limited restriction on what will be asked, but stories and questions were designed so that they can be understood by young children, thus limiting the amount of required outside knowledge to be somewhat manageable. MCTest may be downloaded at \url{https://github.com/mcobzarenco/mctest/tree/master/data/MCTest}.}

\paragraph{RACE.}
\textcolor{black}{The Reading Comprehension Dataset from Examinations (RACE) by \citeA{laiRACELargescaleReAding2017} consists of about 98,000 multiple-choice reading comprehension questions over 28,000 passages collected from middle- and high-school English exams. As the exams are designed by expert teachers, the questions are challenging, often requiring reasoning over multiple sentences in the passage. Further, many questions and candidate choices cannot be matched directly to the passage. RACE can be requested for download at \url{https://www.cs.cmu.edu/~glai1/data/race/}. }

\paragraph{NarrativeQA.}
\textcolor{black}{NarrativeQA by \citeA{kociskyNarrativeQAReadingComprehension2018} is a collection of about 1,500 full stories and movie scripts paired with human-written summaries, along with about 47,000 human-written questions and answers based on the summaries. Two open-ended tasks are proposed: answering the questions paired with the original stories and scripts, and answering the questions paired with the summaries. As the questions were written based upon summaries of the stories and scripts, answering the questions based on the full stories and scripts is more challenging, and requires reasoning over a larger span of text. 
NarrativeQA can be downloaded at \url{https://github.com/deepmind/narrativeqa}.}

\paragraph{ARC.}
The AI2 Reasoning Challenge (ARC) from \citeA{clarkThinkYouHave2018} provides a dataset of almost 8,000 four-way multiple-choice science questions and answers, as well as a corpus of 14 million science-related sentences which are claimed to contain most of the information needed to answer the questions. As many of the questions require systems to draw information from multiple sentences in the corpus to answer correctly, it is not possible to accurately solve the task by simply searching the corpus for keywords. As such, this task encourages reasoning over a large amount of knowledge, an essential skill for NLI systems. ARC can be downloaded at \url{https://data.allenai.org/arc/}.

\paragraph{MCScript.}
The MCScript benchmark by \citeA{ostermannMCScriptNovelDataset2018} is one of few QA benchmarks which emphasizes commonsense knowledge. The dataset consists of about 14,000 2-way multiple-choice questions based on short passages, with a large proportion of its questions requiring pure commonsense knowledge to answer, and thus can be answered without the passage. Such an example is given in Figure~\ref{fig:qa examples}. Questions are conveniently labeled with whether they are answered with the provided text or commonsense. A download link to MCScript can be found at \url{https://www.sfb1102.uni-saarland.de/?page_id=2582}.

\paragraph{ProPara.}
ProPara by \citeA{mishraTrackingStateChanges2018} consists of 488 annotated paragraphs of procedural text. These paragraphs describe various processes such as photosynthesis and hydroelectric power generation in order for systems to learn object tracking in processes which involve changes of state. The authors assert that recognizing these state changes can require knowledge of physical processes, so proficiency in reasoning with such knowledge is required to perform well. Annotations of the paragraphs are in the form of a grid which describes the state of each participant in the paragraph after each sentence of the paragraph. If a system understands a paragraph in this dataset, it is said that for each entity mentioned in the paragraph, it can answer any question about whether the entity is created, destroyed, or moved, and when and where this happens. To answer all possible questions accurately, a system must produce a grid identical to the annotations for the paragraph. Thus, systems are evaluated by their performance on this task. ProPara data is linked from \url{https://data.allenai.org/propara/}.

\paragraph{MultiRC.}
MultiRC by \citeA{KCRUR18} is a reading comprehension dataset consisting of about 10,000 questions posed on over 800 paragraphs across a variety of topic domains. It differs from a traditional reading comprehension dataset in that most questions can only be answered by reasoning over multiple sentences in the accompanying paragraphs, answers are not spans of text from the paragraph, and the number of answer choices as well as the number of correct answers for each question is entirely variable. All of these features make it difficult for shallow approaches to perform well on the benchmark, and encourage a deeper understanding of the passage. Further, the benchmark includes a variety of nontrivial semantic phenomena in passages, such as coreference and causal relationships, which often require external knowledge to recognize and reason about. MultiRC can be downloaded at \url{https://cogcomp.org/multirc/}.

\paragraph{ARCT.}
\textcolor{black}{The Argument Reasoning Comprehension Task (ARCT) proposed by \citeA{habernalArgumentReasoningComprehension2018} is a different type of question answering task which provides the structures of arguments presented in online comments on news articles, requiring systems to choose one of two lexically similar warrants to justify a claim made in an argument. According to the authors, many such warrants, e.g., ``most house cats face enemies,'' are unstated, rather presupposed by arguments, and require external knowledge to infer. This formulation makes available some of the required knowledge for interpreting the presented arguments, however many of the examples require further world knowledge beyond the provided warrants \cite{nivenProbingNeuralNetwork2019}. ARCT consists of about 2,500 examples, and is available at \url{https://github.com/UKPLab/argument-reasoning-comprehension-task}.}

\paragraph{SQuAD.}
The Stanford Question Answering Dataset (SQuAD) from \citeA{rajpurkarSQuAD1000002016} was originally a set of about 100,000 questions posed on passages from Wikipedia articles, which are provided with the questions. The initial dataset did not require any external knowledge or reasoning to solve; answers were spans of text directly from the passage, and could be found based on linguistic context. To make this dataset more challenging, SQuAD 2.0 \cite{rajpurkarKnowWhatYou2018} was later released to add about 50,000 additional questions which are unanswerable given the passage. Such questions require one to reason about what can and cannot be inferred given the knowledge from the passage, which takes a deep understanding.
All SQuAD data can be downloaded at \url{https://rajpurkar.github.io/SQuAD-explorer/}.

\paragraph{CoQA.}
The Conversational Question Answering (CoQA) dataset by \citeA{reddyCoQAConversationalQuestion2018} contains passages each accompanied with a set of questions in conversational form, as well as their answers and evidence for the answers as spans of text. There are about 127,000 questions in the dataset total, where answers are abstractive but a rationale must be given as a span of text. As questions are conversational, \textcolor{black}{each question pertaining to a passage must be answered given the conversation history generated from previous questions on the passage.} Unlike other conversational QA datasets, CoQA explicitly requires external knowledge in many questions, e.g., the example given in Figure~\ref{fig:qa examples}. Further, CoQA includes out-of-domain types of questions appearing only in the test set, and unanswerable questions throughout, encouraging reasoning beyond the passage and knowledge learned in training to answer questions. CoQA data can be downloaded at \url{https://stanfordnlp.github.io/coqa/}.

\paragraph{QuAC.}
\textcolor{black}{Question Answering in Context (QuAC) by \citeA{choiQuACQuestionAnswering2018} is a similar but larger conversational QA benchmark consisting of about 98,000 questions over nearly 14,000 dialogs about nearly 9,000 passages. Answers here are spans of text, while in CoQA answers are abstractive, but rationalized by spans of text. Unlike CoQA, when generating conversations, the worker asking the questions could not see the text and thus did not know the answer, encouraging the creation of questions which cannot be answered by string matching approaches. To ensure productive conversations, the worker answering the questions provided dialog acts which gave the other worker feedback on whether to follow up the previous question or change topics. Like CoQA, QuAC includes unanswerable questions, but does not explicitly address the use of external knowledge. QuAC can be downloaded at \url{https://quac.ai/}.}

\paragraph{OpenBookQA.}
OpenBookQA by \citeA{mihaylovCanSuitArmor2018} intends to address some shortcomings of previous QA datasets. Earlier datasets often do not require external knowledge or reasoning to solve, and those that do require vast domains of knowledge which are difficult to capture. OpenBookQA contains about 6,000 4-way multiple choice science questions which may require science facts (common knowledge) or commonsense knowledge, e.g., the example in Figure~\ref{fig:qa examples}. Instead of providing no knowledge resource like MCScript \cite{ostermannMCScriptNovelDataset2018}, or a prohibitively large corpus of facts to support answering the questions like ARC \cite{clarkThinkYouHave2018}, OpenBookQA provides an ``open book'' of about 1,300 science facts to support answering the questions, each associated directly with the question(s) they apply to. For required commonsense knowledge, the authors expect that outside resources will be used in answering the questions. Instructions to download OpenBookQA can be found at \url{https://github.com/allenai/OpenBookQA}.

\paragraph{CommonsenseQA.}
CommonsenseQA by \citeA{talmorCommonsenseQAQuestionAnswering2019} is another QA benchmark which directly targets commonsense knowledge, consisting of 9,500 three-way multiple-choice questions. To ensure commonsense knowledge is used to answer questions, each question requires one to disambiguate a target concept from three connected concepts in ConceptNet, a popular commonsense knowledge graph \cite{liuConceptNetPracticalCommonsense2004}. Utilizing a large, general-purpose knowledge graph like ConceptNet ensures not only that questions target commonsense relations directly, but that the domain of commonsense knowledge required by questions is fairly comprehensive for everyday use. CommonsenseQA data can be downloaded at \url{https://www.tau-nlp.org/commonsenseqa}.

\paragraph{DREAM.}
\textcolor{black}{The Dialogue-Based Reading Comprehension Examination (DREAM) by \citeA{sunDREAMChallengeDataset2019} also targets the use of external knowledge. The benchmark consists of about 10,000 multiple-choice questions posed on about 6,500 selections of dialogue between multiple parties. Their analysis shows that the answers to most questions cannot be directly extracted from the text, and will require multi-sentence reasoning. Further, 34\% of questions explicitly require external knowledge to answer. Such questions are challenging to current neural reading comprehension systems, especially those which are not typically used for processing dialogue. DREAM can be downloaded at \url{https://dataset.org/dream/}.}

\paragraph{DROP.}
\textcolor{black}{The Discrete Reasoning Over Paragraphs (DROP) benchmark by \citeA{duaDROPReadingComprehension2019} consists of nearly 100,000 questions over about 7,000 passages from Wikipedia. Unlike previous similar benchmarks, however, the questions in DROP deliberately require systems to perform various reasoning tasks, including arithmetic (e.g., addition, subtraction), comparison, and more. This promotes a deep understanding of passages in order to use the knowledge in them for reasoning tasks. Further, the data was adversarially created, only including questions which a baseline neural system could not answer correctly. DROP can be downloaded at \url{https://allennlp.org/drop}.}

\paragraph{Cosmos QA.}
\textcolor{black}{Commonsense Machine Comprehension (Cosmos) QA by \cite{huang2019cosmos} consists of 35,600 multiple-choice reading comprehension questions, each based on an accompanying text. Questions are designed to require external knowledge and  reasoning, some questions are unanswerable, and most correct answers are not stated in the context passage. About 94\% of questions require commonsense knowledge, which is the highest proportion seen yet for any QA benchmark requiring the comprehension of a passage. Cosmos QA can be downloaded at \url{https://wilburone.github.io/cosmos/}.}

\paragraph{MC-TACO.}
\textcolor{black}{The Multiple Choice Temporal Common-sense (MC-TACO) benchmark by \citeA{zhou2019going} provides about 1,900 multiple-choice questions requiring commonsense knowledge about the temporal properties of events, e.g., their duration, frequency, and order. Such commonsense knowledge was under-emphasized in previous benchmarks. Questions can have multiple correct answers. MC-TACO can be downloaded at \url{https://leaderboard.allenai.org/mctaco/submissions/public}.}

\subsubsection{Textual Entailment}
Textual entailment is defined by \citeA{daganPASCALRecognisingTextual2005} as a directional relationship between a text and a hypothesis, where it can be said that the text \textit{entails} the hypothesis if a typical person would infer that the hypothesis is true given the text. Some benchmarks expand this task by also requiring recognition of contradiction, e.g., the fourth and fifth RTE Challenges \cite{giampiccoloFourthPASCALRecognizing2008,bentivogliFifthPASCALRecognizing2009}. Like question answering, this task requires the utilization of several simpler language processing skills, such as named entity recognition and coreference resolution. Unlike question answering, since it also requires a sense of what a typical person would infer, commonsense knowledge is inherently essential to the textual entailment task. The RTE Challenges \cite{daganPASCALRecognisingTextual2005} are the classic benchmarks for entailment, but there are now several larger benchmarks inspired by them. Examples from these benchmarks are listed in Figure~\ref{fig:rte examples}, and they require common knowledge such as the concept of a snow angel, and commonsense knowledge such as the relationship between the presence of a crowd and loneliness. In the following paragraphs, we introduce several textual entailment benchmarks in detail.

\begin{figure}
\noindent\hrulefill
\begin{multicols}{2}
\begin{enumerate}[label=\textbf{(\Alph*)}]
\item \textbf{RTE Challenge \cite{daganPASCALRecognisingTextual2005}} \newline
\textit{Text:} American Airlines began laying off hundreds of flight attendants on Tuesday, after a federal judge turned aside a union's bid to block the job losses. \newline
\textit{Hypothesis:} American Airlines will recall hundreds of flight attendants as it steps up the number of flights it operates.

\textit{Label:} \textbf{not entailment}

\item \textbf{SICK \cite{marelliSICKCureEvaluation2014}\footnotemark} \newline
\textit{Sentence 1:} Two children are lying in the snow and are drawing angels.\newline
\textit{Sentence 2:} Two children are lying in the snow and are making snow angels.

\textit{Label:} \textbf{entailment}\newline\newline\newline

\item \textbf{SNLI \cite{bowmanLargeAnnotatedCorpus2015}} \newline
\textit{Text:} A black race car starts up in front of a crowd of people. \newline
\textit{Hypothesis:} A man is driving down a lonely road.

\textit{Label:} \textbf{contradiction}

\item \textbf{MultiNLI, Telephone \cite{williamsBroadCoverageChallengeCorpus2017}} \newline
\textit{Context:} that doesn't seem fair does it	\newline
\textit{Hypothesis:} There's no doubt that it's fair.

\textit{Label:} \textbf{contradiction}

\item \textbf{SciTail \cite{khotSciTailTextualEntailment2018}} \newline
\textit{Premise:} During periods of drought, trees died and prairie plants took over previously forested regions. \newline
\textit{Hypothesis:} Because trees add water vapor to air, cutting down forests leads to longer periods of drought.

\textit{Label:} \textbf{neutral}
\end{enumerate}
\end{multicols}
\noindent\hrulefill
\caption{Examples from RTE benchmarks. Answers in bold.}
\label{fig:rte examples}
\end{figure}
\footnotetext{Example extracted from the SICK data available at \url{https://clic.cimec.unitn.it/composes/sick.html}.}

\paragraph{RTE Challenges.}
The Recognizing Textual Entailment (RTE) Challenge introduced the task of textual entailment, and aimed to evaluate a machines' acquisition of the common background knowledge and reasoning capabilities a typical human needs to determine whether one text entails another \cite{daganPASCALRecognisingTextual2005}. The inaugural Challenge provided a task where given a text and hypothesis, systems were expected to predict whether the text entailed the hypothesis. In following years, more similar Challenges took place \cite{bar-haimSecondPASCALRecognising2006,giampiccoloThirdPASCALRecognizing2007}. The fourth and fifth Challenge added a new three-way decision task where systems were additionally expected to recognize contradiction relationships between texts and hypotheses \cite{giampiccoloFourthPASCALRecognizing2008,bentivogliFifthPASCALRecognizing2009}. The main task for the sixth and seventh Challenges instead provided one hypothesis and several potential entailing sentences in a corpus \cite{bentivogliSixthPASCALRecognizing2010,bentivogliSeventhPASCALRecognizing2011}. The eighth Challenge \cite{dzikovskaSemEval2013TaskJoint2013} addressed a bit different problem which focused on classifying student responses as an effort toward providing automatic feedback in an educational setting. The first five RTE Challenge datasets consisted of around 1,000 examples each \cite{daganPASCALRecognisingTextual2005,bar-haimSecondPASCALRecognising2006,giampiccoloThirdPASCALRecognizing2007,giampiccoloFourthPASCALRecognizing2008,bentivogliFifthPASCALRecognizing2009}, while the sixth and seventh consisted of about 33,000 and 49,000 examples respectively. Data from all RTE Challenges can be downloaded at \url{https://tac.nist.gov/}.

\paragraph{Conversational Entailment.}
\textcolor{black}{The Conversational Entailment dataset, introduced by \citeA{zhang-chai-2009-know} and further analyzed by \citeA{zhang-chai-2010-towards}, is a small collection of 875 binary entailment pairs consisting of a premise from a natural language dialogue and a hypothesis that requires interpreting the dialogue. Many examples require predicting dialogue participants' beliefs, desires, and intentions, which need to be reasoned from the context of dialogue. 
The dataset can be downloaded at \url{https://github.com/lairmsu/ConversationalEntailment}.}

\paragraph{SICK.}
The Sentences Involving Compositional Knowledge (SICK) benchmark by \citeA{marelliSICKCureEvaluation2014} is a collection of almost 10,000 pairs of sentences. Two tasks are presented for the pairs: sentence relatedness and entailment. The entailment task, which is more related to our survey, is a 3-way decision task in the style of RTE-4 \cite{giampiccoloFourthPASCALRecognizing2008} and RTE-5 \cite{bentivogliFifthPASCALRecognizing2009}. SICK can be downloaded at \url{https://clic.cimec.unitn.it/composes/sick.html}.

\paragraph{SNLI.}
The Stanford Natural Language Inference (SNLI) benchmark from \citeA{bowmanLargeAnnotatedCorpus2015} contains nearly 600,000 sentence pairs, and also provides a 3-way decision task similar to the fourth and fifth RTE Challenges \cite{giampiccoloFourthPASCALRecognizing2008,bentivogliFifthPASCALRecognizing2009}. In addition to gold labels for entailment, contradiction, or neutral, the SNLI data includes five crowd judgments for the label, which may indicate a level of confidence or agreement for it. This benchmark is later expanded into MultiNLI \cite{williamsBroadCoverageChallengeCorpus2017}, which follows the same format, but includes sentences of various genres, such as telephone conversations. MultiNLI is included within the later introduced GLUE benchmark \cite{wangGLUEMultiTaskBenchmark2018}, while SNLI can be downloaded at \url{https://nlp.stanford.edu/projects/snli/}.

\paragraph{SciTail.}
SciTail by \citeA{khotSciTailTextualEntailment2018} consists of about 27,000 premise-hypothesis sentence pairs adapted from science questions into a 2-way entailment task. Unlike other entailment benchmarks, this one is primarily science-based, which may require domain-specific knowledge. SciTail can be downloaded at \url{https://data.allenai.org/scitail/}.

\paragraph{SherLIiC.}
\textcolor{black}{SherLIiC by \citeA{schmitt-schutze-2019-sherliic} is a benchmark for lexical inference in context (LIiC), i.e., a subset of textual entailment focusing on relationships between words, which \citeA{glockner-etal-2018-breaking} found that previous benchmarks lacked, and state-of-the-art systems struggled with. SherLIiC focuses particularly on entailment relationships between event and action verbs. It consists of about 960,000 inference rule candidates formed from pairs of relations extracted from a knowledge graph, each typed for better word disambiguation. Almost 4,000 of these are annotated for their correctness. Additionally, the benchmark includes several automatic calculations of similarity between premises and hypotheses. 
SherLIiC serves as a challenging testbed for NLI and textual entailment systems.
It can be downloaded at \url{https://github.com/mnschmit/SherLIiC}.}

\subsubsection{Plausible Inference}
While textual entailment benchmarks require one to draw concrete conclusions, others require hypothetical, intermediate, or uncertain conclusions based upon a limited context, defined as plausible inference by \citeA{davisCommonsenseReasoningCommonsense2015}, \textcolor{black}{and historically referred to as abductive reasoning \cite{Peirce1883,HOBBS199369}.} Performing this process requires reasoning over linguistic context and external knowledge.
Popular task formulations include cloze tasks, i.e., fill-in-the-blank tasks, and sentence and story completion. Examples from these benchmarks are listed in Figure~\ref{fig:pi examples}, and they require commonsense knowledge of everyday interactions, e.g., answering a door or proposing for marriage, and activities, e.g., making scrambled eggs or being snowed in. In the following paragraphs, we introduce all surveyed plausible inference benchmarks.

\begin{figure}
\noindent\hrulefill
\begin{multicols}{2}
\begin{enumerate}[label=\textbf{(\Alph*)}]
\item \textbf{COPA \cite{roemmeleChoicePlausibleAlternatives2011}} \newline
I knocked on my neighbor's door. What happened as result?
\begin{enumerate}[label=\alph*.]
\item \textbf{My neighbor invited me in.}
\item My neighbor left his house.
\end{enumerate} 

\item \textbf{JOCI \cite{zhangOrdinalCommonsenseInference2016}} \newline
\textit{Context:} John was excited to go to the fair \newline
\textit{Hypothesis:} The fair opens. \newline
\textit{Label:} \textbf{5 (very likely)}

\textit{Context:} Today my water heater broke \newline
\textit{Hypothesis:} A person looks for a heater. \newline
\textit{Label:} \textbf{4 (likely)}

\textit{Context:} John's goal was to learn how to draw well \newline
\textit{Hypothesis:} A person accomplishes the goal. \newline
\textit{Label:} \textbf{3 (plausible)}

\textit{Context:} Kelly was playing a soccer match for her University \newline
\textit{Hypothesis:} The University is dismantled. \newline
\textit{Label:} \textbf{2 (technically possible)}

\textit{Context:} A brown-haired lady dressed all in blue denim sits in a group of pigeons. \newline
\textit{Hypothesis:} People are made of the denim. \newline
\textit{Label:} \textbf{1 (impossible)}\newline\newline\newline\newline

\item \textbf{ROCStories \cite{mostafazadehCorpusClozeEvaluation2016}} \newline
Tom and Sheryl have been together for two years. One day, they went to a carnival together. He won her several stuffed bears, and bought her funnel cakes. When they reached the Ferris wheel, he got down on one knee.

\textit{Ending:}
\begin{enumerate}[label=\alph*.]
\item \textbf{Tom asked Sheryl to marry him.}
\item He wiped mud off of his boot.
\end{enumerate}

\item \textbf{AlphaNLI \cite{ch2019abductive}} \newline
\textit{Observation 1:} There was ten feet of snow outside. \newline
\textit{Observation 2:} In all that time I was unable to check my mail. \newline
\textit{Hypotheses:}
\begin{enumerate}[label=\alph*.]
\item \textbf{I couldn't open my door against a drift for 3 days.}
\item It took 10 minutes for the snow plow to come through.
\end{enumerate}

\item \textbf{SWAG \cite{zellersSWAGLargeScaleAdversarial2018}} \newline
He pours the raw egg batter into the pan. He
\begin{enumerate}[label=\alph*.]
\item drops the tiny pan onto a plate
\item \textbf{lifts the pan and moves it around to shuffle the eggs.}
\item stirs the dough into a kite.
\item swirls the stir under the adhesive.
\end{enumerate}

\end{enumerate}
\end{multicols}
\noindent\hrulefill
\caption{Examples from benchmarks requiring plausible inference. Answers in bold.}
\label{fig:pi examples}
\end{figure}

\paragraph{COPA.}
The Choice of Plausible Alternatives (COPA) task by \citeA{roemmeleChoicePlausibleAlternatives2011} evaluates causal reasoning between events, which requires commonsense knowledge about what usually takes place in the world. Each example provides a premise and either asks for the correct cause or effect from two choices, thus testing either backward or forward causal reasoning. COPA data, which consists of 1,000 examples total, can be downloaded at \url{https://people.ict.usc.edu/~gordon/copa.html}.

\paragraph{CBT.}
The Children's Book Test (CBT) from \citeA{hillGoldilocksPrincipleReading2015} consists of about 687,000 cloze-style questions about 20-sentence stories mined from publicly available children's books. Each question requires a system to fill a blank in a new line following the story given a set of 10 candidate words. Questions are classified into 4 tasks based on the type of the missing word to predict, which can be a named entity, common noun, verb, or preposition. To predict a named entity or common noun, an understanding of the story and the actions involving the entities in it will be required. To predict a verb or preposition, outside linguistic and common knowledge may be useful. CBT can be downloaded at \url{https://research.fb.com/downloads/babi/}.

\paragraph{ROCStories.}
ROCStories by \citeA{mostafazadehCorpusClozeEvaluation2016} is a corpus of about 50,000 five-sentence everyday life stories, containing a host of causal and temporal relationships between events, ideal for learning commonsense knowledge. Of the stories, about 3,700 are designated as test cases, which include a plausible and implausible alternate story ending for trained systems to choose between. The task of solving the ROCStories test cases is called the Story Cloze Test, a more challenging alternative to the narrative cloze task proposed by \citeA{chambersUnsupervisedLearningNarrative2008}. The most recent release of ROCStories, which can be requested at \url{https://cs.rochester.edu/nlp/rocstories/}, adds about 50,000 stories to the dataset.

\paragraph{LAMBADA.}
\textcolor{black}{The Language Modeling Broadened to Account for Discourse Aspects (LAMBADA) benchmark by \citeA {papernoLAMBADADatasetWord2016} is an open-ended cloze task which consists of about 10,000 passages from BooksCorpus \cite{zhuAligningBooksMovies2015} where a missing target word is predicted in the last sentence of each passage. This is similar to CBT \citeA{hillGoldilocksPrincipleReading2015}, but here, the missing word is constrained to always be the last word of the last sentence, there are no candidate words to choose from, and most importantly, examples were filtered by humans to ensure they were possible to guess given the context, i.e., the sentences in the passage leading up to the last sentence. Examples were further filtered to ensure that missing words could not be guessed without the context, ensuring that models attempting the dataset would need to reason over the entire paragraph to answer questions. Further, in a later study on a random sampling of 100 passages from LAMBADA, 24 were found to require external knowledge \citeA{chuBroadContextLanguage2017}. LAMBADA can be downloaded at \url{https://zenodo.org/record/2630551\#.XS5JHuhKhPZ}.}

\paragraph{JOCI.}
The JHU (John Hopkins University) Ordinal Commonsense Inference (JOCI) benchmark by \citeA{zhangOrdinalCommonsenseInference2016} consists of about 39,000 sentence pairs, each consisting of a context and hypothesis. Given these, systems must rate how likely the hypothesis is on a scale from 1 to 5, where 1 corresponds to impossible, 2 to technically possible, 3 to plausible, 4 to likely, and 5 to very likely. This task is similar to SNLI \cite{bowmanLargeAnnotatedCorpus2015} and other 3-way entailment tasks, but provides more options which essentially range between entailment and contradiction. This is fitting, considering the fuzzier nature of the plausible inference task compared to textual entailment. JOCI can be downloaded from \url{https://github.com/sheng-z/JOCI}.

\paragraph{CLOTH.}
The Cloze Test by Teachers (CLOTH) benchmark by \citeA{xieLargescaleClozeTest2017} is a collection of nearly 100,000 4-way multiple-choice cloze-style questions from middle- and high school-level English language exams, where the answer fills a blank in a given text. Each question is labeled with a type of deep reasoning it involves, where the four possible types are grammar, short-term reasoning, matching/paraphrasing, and long-term reasoning, i.e., reasoning over multiple sentences. CLOTH data can be downloaded at \url{https://www.cs.cmu.edu/~glai1/data/cloth/}.

\paragraph{SWAG.}
Situations with Adversarial Generations (SWAG) from \citeA{zellersSWAGLargeScaleAdversarial2018} is a benchmark dataset of about 113,000 beginnings of small texts each with four possible endings. Given the context each text provides, systems decide which of the four endings is most plausible in a task referred to as commonsense NLI. To ensure the benchmark was challenging to state-of-the-art systems and could not be solved through shallow approaches or by exploiting data bias, examples were filtered with a new process called adversarial filtering (explained further in Section~\ref{sec:discussion}). Examples include labels for the source of the correct ending, and ordinal labels for the likelihood of each possible ending and the correct ending. SWAG data can be downloaded at \url{https://rowanzellers.com/swag/}.

\paragraph{ReCoRD.}
The Reading Comprehension with Commonsense Reasoning (ReCoRD) benchmark by \citeA{zhangReCoRDBridgingGap2018}, similar to SQuAD \cite{rajpurkarSQuAD1000002016}, consists of questions posed on passages, particularly news articles. However, questions in ReCoRD are in cloze format, requiring more hypothetical reasoning, and many questions explicitly require commonsense knowledge and reasoning over multiple sentences to answer. Named entities are identified in the data, and are used to fill the blanks for the cloze task. The benchmark data consist of over 120,000 examples, most of which are claimed to require commonsense reasoning. ReCoRD can be downloaded at \url{https://sheng-z.github.io/ReCoRD-explorer/}.

\paragraph{HellaSWAG.}
\textcolor{black}{HellaSWAG by \citeA{zellers2019hellaswag} is a reissue of SWAG \cite{zellersSWAGLargeScaleAdversarial2018} consisting of 70,000 examples in the same format, but created from different sources, i.e., ActivityNet captions \cite{heilbronActivityNetLargescaleVideo2015} along with WikiHow texts rather than text from the Large Scale Movie Description Challenge \cite{rohrbachMovieDescription2017}. Further, negative endings are generated with a better language model, and again adversarially filtered to be challenging for state-of-the-art models that were created since SWAG's first release. HellaSWAG can be downloaded at \url{https://rowanzellers.com/hellaswag/}.}

\paragraph{AlphaNLI.}
\textcolor{black}{AlphaNLI by \citeA{ch2019abductive} is an abductive reasoning task where given two observations as an incomplete context, one must predict which of two hypothesized events is more plausible to have happened between the observations. This is similar to the Story Cloze Test \cite{mostafazadehCorpusClozeEvaluation2016}, which AlphaNLI is partially constructed from. However, rather than inferring the most plausible ending of an otherwise complete story, one must infer the most plausible explanation of a situation given two partial observations of it. As such, the benchmark closely resembles the classical formulation of abductive reasoning studied with formal logic \cite{Peirce1883,HOBBS199369}, unlike other benchmarks in this category, and performing the task requires reasoning with unstated commonsense implications. AlphaNLI consists of about 20,000 possible contexts and 200,000 hypotheses, paired to form about 170,000 examples in the training and development sets.\footnote{The authors do not report the magnitude of any of the dataset partitions, so these counts came from the publicly available partitions of the dataset.} AlphaNLI can be downloaded at \url{https://leaderboard.allenai.org/anli/submissions/get-started}.}

\subsubsection{Intuitive Psychology}\label{sec:intpsych}
An especially significant domain in plausible inference tasks is intuitive psychology, as inference of emotions and intentions through behavior is a fundamental capability of humans \cite{gordonCommonsenseInterpretationTriangle2016}. Several benchmarks touch on this topic in some examples, e.g., the marriage proposal example in ROCStories \cite{mostafazadehCorpusClozeEvaluation2016} from Figure~\ref{fig:pi examples}, but some are entirely focused here. Examples from each benchmark are listed in Figure~\ref{fig:psych examples}, and they require intuitive social and psychological commonsense knowledge such as plausible reactions to being punched or yelled at. In the following paragraphs, we introduce these benchmarks in detail.

\begin{figure}
\noindent\hrulefill
\begin{multicols}{2}
\begin{enumerate}[label=\textbf{(\Alph*)}]
\item \textbf{Triangle-COPA \cite{gordonCommonsenseInterpretationTriangle2016}} \newline
A circle and a triangle are in the house and are arguing. The circle punches the triangle. The triangle runs out of the house. Why does the triangle leave the house?\begin{enumerate}[label=\alph*.]
\item The triangle leaves the house because it wants the circle to come fight it outside.
\item \textbf{The triangle leaves the house because it is afraid of being further assaulted by the circle.}
\end{enumerate}

\item \textbf{SC \cite{rashkinModelingNaivePsychology2018}\footnotemark} \newline
Jervis has been single for a long time. \newline
He wants to have a girlfriend. \newline
One day he meets a nice girl at the grocery store.

\textit{Motivation:} \textbf{to be loved, companionship} \newline
\textit{Emotions:} \textbf{shocked, excited, hope, shy, fine, wanted}

\textit{Maslow:} \textbf{love}\newline
\textit{Reiss:} \textbf{contact, romance} \newline
\textit{Plutchik:} \textbf{joy, surprise, anticipation, trust, fear}

\item \textbf{Event2Mind \cite{rashkinEvent2MindCommonsenseInference2018}} \newline
PersonX starts to yell at PersonY

\textit{PersonX's intent:} \textbf{to express anger, to vent their frustration, to get PersonY's full attention} \newline
\textit{PersonX's reaction:} \textbf{mad, frustrated, annoyed} \newline
\textit{Other's reactions:} \textbf{shocked, humiliated, mad at PersonX}

\item \textbf{SocialIQA \cite{sapSocialIQACommonsenseReasoning2019}} \newline
Tracy accidentally pressed upon Austin in the small elevator and it was awkward. Why did Tracy do this?\begin{enumerate}[label=\alph*.]
\item get very close to Austin
\item \textbf{squeeze into the elevator}
\item get flirty with Austin
\newline \newline \newline \newline
\end{enumerate}

\end{enumerate}
\end{multicols}
\noindent\hrulefill
\caption{Examples from intuitive psychology benchmarks. Answers in bold.}
\label{fig:psych examples}
\end{figure}
\footnotetext{Example extracted from Story Commonsense development data available at \url{https://uwnlp.github.io/storycommonsense/}.}

\paragraph{Triangle-COPA.}
Triangle-COPA by \citeA{gordonCommonsenseInterpretationTriangle2016} is a variation of COPA \cite{roemmeleChoicePlausibleAlternatives2011} based on a past social psychology experiment by \citeA{heiderExperimentalStudyApparent1944}. It contains 100 examples in the format of COPA, and accompanying videos. Questions focus specifically on emotions, intentions, and other aspects of social psychology. The data also includes logical forms of the questions and alternatives, as the paper focuses on logical formalisms for psychological commonsense knowledge and reasoning. Triangle-COPA can be downloaded at \url{https://github.com/asgordon/TriangleCOPA}.

\paragraph{Story Commonsense.}
As mentioned earlier, the stories from ROCStories by \citeA{mostafazadehCorpusClozeEvaluation2016} contain sociological and psychological instances of commonsense knowledge. Motivated by classical theories of motivation and emotions from psychology, \citeA{rashkinModelingNaivePsychology2018} created the Story Commonsense benchmark containing about 160,000 annotations of the motivations and emotions of characters in ROCStories to enable more concrete reasoning in this area. In addition to the tasks of generating motivational and emotional annotations, the dataset introduces three classification tasks: one for inferring the basic human needs theorized by \citeA{maslowTheoryHumanMotivation1943}, one for inferring the human motives theorized by \citeA{reissMultifacetedNatureIntrinsic2004}, and one for inferring the human emotions theorized by \citeA{plutchikGeneralPsychoevolutionaryTheory1980}. Story Commonsense can be downloaded at \url{https://uwnlp.github.io/storycommonsense/}.

\paragraph{Event2Mind.}
In addition to motivations and emotions, systems may need to infer intentions and reactions surrounding events. To support this, \citeA{rashkinEvent2MindCommonsenseInference2018} introduce Event2Mind, a benchmark dataset of about 57,000 annotations of intentions and reactions for about 25,000 unique events extracted from other corpora, including ROCStories \cite{mostafazadehCorpusClozeEvaluation2016}. Each event involves one or two participants, and presents three tasks of predicting the primary participant's (1) intentions and (2) reactions, and predicting (3) the reactions of other participants. Event2Mind can be downloaded at \url{https://uwnlp.github.io/event2mind/}.

\paragraph{SocialIQA.}
Social Intelligence QA (SocialIQA) by  \citeA{sapSocialIQACommonsenseReasoning2019} is a multiple-choice question answering benchmark containing 45,000 questions requiring intuitive psychology and commonsense knowledge of social interactions. Each question consists of a brief context, a question about the context, and three answer options. Interestingly, the authors show that when fine-tuned on SocialIQA, deep learning models perform better when later fine-tuned on smaller-magnitude benchmarks which require commonsense knowledge such as COPA \cite{roemmeleChoicePlausibleAlternatives2011} and the Winograd Schema Challenge \cite{davisFirstWinogradSchema2017}, exceeding the state of the art on both. This suggests that some useful commonsense knowledge can be acquired from the data. More information about SocialIQA can be found at \url{https://maartensap.github.io/social-iqa/}.

\begin{table}[H]
\centering
\begin{tabular}{P{5cm}P{1.2cm}P{1.2cm}P{1.2cm}P{1.2cm}P{2.2cm}}\toprule
\thead{\textbf{Task}} & \thead{\textbf{bAbI}} & \thead{\textbf{IIE}} & \thead{\textbf{GLUE}} & \thead{\textbf{DNC}} & \thead{\textbf{SuperGLUE}} \\ \toprule
Semantic Role Labeling &  & \checkmark &  & &  \\\midrule
Relation Extraction & \checkmark  &  &  & \checkmark &  \\\midrule
Event Factuality &  &  &  & \checkmark & \\\midrule
Named Entity Recognition &  &  &  & \checkmark &  \\\midrule
Word Sense Disambiguation & & & &  & \checkmark \\\midrule
Reference Resolution & \checkmark & \checkmark & \checkmark & \checkmark & \checkmark \\\midrule
Grammaticality &  &  & \checkmark & &  \\\midrule
Lexicosyntactic Inference &  &  &  & \checkmark & \checkmark \\\midrule
Sentiment Analysis &  &  & \checkmark & \checkmark & \\\midrule
Figurative Language &  &  &  & \checkmark & \\\midrule
Sentence Similarity & & & \checkmark &  & \\\midrule
Paraphrase &  & \checkmark & \checkmark & \checkmark & \\\midrule
Sentence Completion & & & & & \checkmark \\\midrule
Textual Entailment &  &  &  \checkmark & \checkmark & \checkmark \\\midrule
Question Answering & \checkmark & & \checkmark & & \checkmark \\\midrule
\end{tabular}
\caption{Comparison of language processing tasks present in the bAbI \cite{westonAICompleteQuestionAnswering2016}, IIE \cite{whiteInferenceEverythingRecasting2017}, GLUE \cite{wangGLUEMultiTaskBenchmark2018}, DNC \cite{poliak2018emnlp-DNC}, and SuperGLUE \cite{wangSuperGLUEStickierBenchmark2019} benchmarks. Recent multi-task benchmarks focus on the recognition of a greater variety of linguistic and semantic phenomena.}
\label{tbl:multitask comparison}
\end{table}

\subsubsection{Multiple Tasks}\label{sec:multiple tasks}
\textcolor{black}{Some benchmarks consist of several focused language processing or reasoning subtasks so that a diverse set of reading comprehension skills can be learned and tested in a consistent format. These benchmarks can be used as diagnostics to determine how a model performs in different areas \cite{wangGLUEMultiTaskBenchmark2018}. Subtasks are often reframed from various pre-existing benchmarks \cite{whiteInferenceEverythingRecasting2017}. As shown in Table~\ref{tbl:multitask comparison}, many benchmark subtasks are specified by low- to high-level language processing tasks. In the following paragraphs, we introduce these benchmarks in detail.}

\paragraph{bAbI.}
The bAbI benchmark from \citeA{westonAICompleteQuestionAnswering2016} consists of 20 prerequisite tasks, each with 1,000 examples for training and 1,000 for testing. Each task presents systems with a passage, then asks a reading comprehension question. Each task also focuses on a different type of reasoning or language processing task, allowing systems to learn basic skills one at a time. Tasks are as follows:
\begin{multicols}{3}
\begin{enumerate}[nolistsep]
\item Single supporting fact
\item Two supporting facts
\item Three supporting facts
\item Two argument relations
\item Three argument relations
\item Yes/no questions
\item Counting
\item Lists/sets
\item Simple negation
\item Indefinite knowledge
\item Basic coreference
\item Conjunction
\item Compound coreference
\item Time reasoning
\item Basic deduction
\item Basic induction
\item Positional reasoning
\item Size reasoning
\item Path finding
\item Agent's motivations
\end{enumerate}
\end{multicols}

In addition to providing focused language processing tasks as previously discussed, bAbI also provides focused commonsense reasoning tasks, such as its tasks for deduction and induction, and time, positional, and size reasoning \cite{westonAICompleteQuestionAnswering2016}. 
\textcolor{black}{It is important to note that bAbI's data is synthetic and very simply structured, and the inference required for testing data can be learned mostly from the training data, so it is easily solved by current systems. Further, models trained on bAbI currently cannot generalize well to real-world, naturally-generated data \cite{dasBuildingDynamicKnowledge2019}. As such, bAbI is best used to gauge a model's ability to learn reasoning skills rather than its comprehension of natural language. A dataset by \citeA{nematzadehEvaluatingTheoryMind2018} expands bAbI with questions about theory of mind, which may make it a more realistic reasoning task.} bAbI can be downloaded at \url{https://research.fb.com/downloads/babi/}.

\paragraph{Inference is Everything.}
Inference is Everything (IIE) by \citeA{whiteInferenceEverythingRecasting2017} follows bAbI \cite{westonAICompleteQuestionAnswering2016} in creating a suite of tasks, where each task is deliberately geared toward a different language processing task: semantic proto-role labeling, paraphrase, and pronoun resolution. Each task is in classic RTE Challenge format \cite{daganPASCALRecognisingTextual2005}, i.e., given context and hypothesis texts, one must determine whether the context entails the hypothesis. Between the tasks, IIE includes about 300,000 examples, all of which are recast from previously existing datasets. IIE can be downloaded with another multi-task suite at \url{https://github.com/decompositional-semantics-initiative/DNC}.

\paragraph{GLUE.}
The General Language Understanding Evaluation (GLUE) dataset from \citeA{wangGLUEMultiTaskBenchmark2018} consists of 9 language tasks, including single-sentence binary classification and 2- or 3-way entailment comparable to the dual tasks in RTE-4 and RTE-5 \cite{giampiccoloFourthPASCALRecognizing2008,bentivogliFifthPASCALRecognizing2009}. The GLUE tasks are recast or included directly from other benchmark data and corpora:

\begin{itemize}[nolistsep]
\item Corpus of Linguistic Acceptability (CoLA) from \citeA{warstadt2018neural} 
\item Stanford Sentiment Treebank (SST-2) from \citeA{socherRecursiveDeepModels2013}
\item Microsoft Research Paraphrase Corpus (MRPC) from  \citeA{dolanAutomaticallyConstructingCorpus2005}
\item Quora Question Pairs (QQP) from \citeA{iyerFirstQuoraDataset2017}
\item Semantic Textual Similarity Benchmark (STS-B) from \citeA{cerSemEval2017TaskSemantic2017}
\item Multi-Genre Natural Language Inference (MNLI) from \citeA{williamsBroadCoverageChallengeCorpus2017}
\item Question Natural Language Inference (QNLI) recast from SQuAD 1.1 \citeA{rajpurkarSQuAD1000002016}
\item Recognizing Textual Entailment (RTE), consisting of examples from RTE-1 \cite{daganPASCALRecognisingTextual2005}, RTE-2 \cite{bar-haimSecondPASCALRecognising2006}, RTE-3 \cite{giampiccoloThirdPASCALRecognizing2007}, and RTE-5 \cite{bentivogliFifthPASCALRecognizing2009}
\item Winograd Natural Language Inference (WNLI) recast from privately shared Winograd schemas by the creators of the Winograd Schema Challenge \citeA{levesqueWinogradSchemaChallenge2011}
\end{itemize}
\vspace{1em}

Many of these tasks, though not all, require some outside knowledge and reasoning to make inferences. GLUE includes a small analysis set for diagnostic purposes, which has manual annotations of fine-grained knowledge and reasoning categories examples fall into. Overall, GLUE has over 1 million examples, which can be downloaded at \url{https://gluebenchmark.com/tasks}.

\paragraph{DNC.}
The Diverse Natural Language Inference Collection (DNC) by \citeA{poliak2018emnlp-DNC} consists of 9 textual entailment tasks requiring 7 different types of reasoning. Like IIE by \citeA{whiteInferenceEverythingRecasting2017}, data for each task follows the form of the original RTE Challenge \cite{daganPASCALRecognisingTextual2005}. \textcolor{black}{Some of the tasks within DNC cover basic language processing tasks like named entity recognition, while others cover more challenging tasks like anaphora resolution and recognition of puns, which require some outside knowledge and reasoning beyond the text.} Each task is recast from a previously existing dataset:

\begin{enumerate}[nolistsep]
\item Event Factuality, recast from UW \cite{lee2015event}, MEANTIME \cite{minardmeantime}, and \cite{neural-models-of-factuality}
\item Named Entity Recognition, recast from the Groningen Meaning Bank \cite{bos2017groningen} and the ConLL-2003 shared task \cite{TjongKimSang:2003:ICS:1119176.1119195}
\item Gendered Anaphora Resolution, recast from the Winogender dataset  \cite{gender-bias-in-coreference-resolution}
\item Lexicosyntactic Inference, recast from MegaVeridicality \cite{white_role_2018}, VerbNet \cite{schuler2005verbnet}, and VerbCorner \cite{hartshorne2013verbcorner}
\item Figurative Language, recast from puns by \citeA{D15-1284} and \citeA{miller-hempelmann-gurevych:2017:SemEval}
\item Relation Extraction, partially from FACC1 \cite{gabrilovich2013facc1}
\item Subjectivity, recast from \citeA{kotzias2015group}
\end{enumerate}
\vspace{1em}

The DNC benchmark consists of about 570,000 examples total, and can be downloaded at \url{https://github.com/decompositional-semantics-initiative/DNC}.

\paragraph{SuperGLUE.}
\textcolor{black}{SuperGLUE by \citeA{wangSuperGLUEStickierBenchmark2019} is a reissue of GLUE \cite{wangGLUEMultiTaskBenchmark2018} with about 23,000 examples over a different selection of more difficult reframed single-sentence binary classification and 2- or 3-way entailment tasks:
\begin{itemize}[nolistsep]
\item CommitmentBank (CB) from \citeA{demarneffeCommitmentBankInvestigatingProjection2019}
\item Choice of Plausible Alternatives (COPA) from \citeA{roemmeleChoicePlausibleAlternatives2011}
\item Multi-Sentence Reading Comprehension (MultiRC) from \citeA{KCRUR18}
\item Recognizing Textual Entailment (RTE), consisting of examples from RTE-1 \cite{daganPASCALRecognisingTextual2005}, RTE-2 \cite{bar-haimSecondPASCALRecognising2006}, RTE-3 \cite{giampiccoloThirdPASCALRecognizing2007}, and RTE-5 \cite{bentivogliFifthPASCALRecognizing2009}
\item Word-in-Context (WiC) from \citeA{pilehvarWiCWordinContextDataset2019}
\item Winograd Schema Challenge (WSC) from \citeA{levesqueWinogradSchemaChallenge2011} recast instead into a single-sentence binary classification task\footnote{Consists of original Winograd Schema Challenge examples \cite{morgensternPlanningExecutingEvaluating2016}, examples by the authors available online, and examples privately shared by the authors.}
\end{itemize}
\vspace{1em}}

Many of these tasks require reasoning over multiple sentences, and using external knowledge to perform language processing tasks like word-sense disambiguation and reference resolution. SuperGLUE can be downloaded at \url{https://super.gluebenchmark.com/}.

\subsection{Criteria and Considerations for Creating Benchmarks}
The goal of benchmarks is to support technology development and provide a platform to measure research progress. 
Whether this goal can be achieved depends on the nature of the benchmark. 
This section identifies the successes and lessons learned from the creation of the surveyed benchmarks, and summarizes key considerations and criteria that should guide the creation of the benchmarks, particularly in the areas of task format, evaluation schemes, data biases, data collection methods, \textcolor{black}{and benchmark complexity}.

\subsubsection{Task Format}
In creating benchmarks, determining the formulation of the problem is an important step. Among existing benchmarks, there exist a few common task formats, and while some formats are interchangeable, others are only suited for particular tasks. We provide a review of these formats, indicating the types of tasks they are suitable for.

\paragraph{Classification tasks.}
Most benchmark tasks are classification problems, where each response is a single choice from a finite number of options. These include textual entailment tasks, which most commonly require a binary or ternary decision about a pair of sentences, cloze tasks, which require a multiple-choice decision to fill in a blank, and traditional multiple-choice question answering tasks.

\vspace{5pt}
\noindent
{\em Textual entailment tasks.}
A highly popular format was originally introduced by the RTE Challenges, where given a pair of texts, i.e., a context and hypothesis, one must determine whether the context entails the hypothesis \cite{daganPASCALRecognisingTextual2005}. In the fourth and fifth RTE Challenges, this format was extended to a three-way decision problem where the hypothesis may contradict the context \cite{giampiccoloFourthPASCALRecognizing2008,bentivogliFifthPASCALRecognizing2009}. The JOCI benchmark further extends the problem to a five-way decision task, where the hypothesis text may range from impossible to very likely given the context \cite{zhangOrdinalCommonsenseInference2016}. 

While this format is typically used for textual entailment problems like the RTE Challenges, it can be used for nearly any type of inference problem. Some multi-task benchmarks have adopted the format for several different reasoning tasks, for example, the Inference is Everything \cite{whiteInferenceEverythingRecasting2017}, GLUE \cite{wangGLUEMultiTaskBenchmark2018}, and DNC \cite{poliak2018emnlp-DNC} benchmarks use the classic RTE format for most of their subtasks by automatically recasting previously existing benchmarks into it. Many of these problems deal either with reasoning processes more focused than the RTE Challenges, or more advanced reasoning than the RTE Challenges demand, showing the flexibility of the format. Some such tasks include reference resolution, recognition of puns, and question answering. Examples from these recasted subtasks are listed in Figure~\ref{fig:rte format examples}.

\begin{figure}
\noindent\hrulefill
\begin{multicols}{2}
\begin{enumerate}[label=\textbf{(\Alph*)}]
\item \textbf{GLUE, Question Answering NLI} \newline
\textit{Context:} Who was the main performer at this year's halftime show?

\textit{Hypothesis:} The Super Bowl 50 halftime show was headlined by the British rock group Coldplay with special guest performers Beyonc{\' e} and Bruno Mars, who headlined the Super Bowl XLVII and Super Bowl XLVIII halftime shows, respectively.

\textit{Label:} \textbf{entailed}

\item \textbf{IIE, Definite Pronoun Resolution} \newline
\textit{Context:} The bird ate the pie and it died.

\textit{Hypothesis:} The bird ate the pie and the bird died.

\textit{Label:} \textbf{entailed}

\item \textbf{DNC, Figurative Language} \newline
\textit{Context:} Carter heard that a gardener who moved back to his home town rediscovered his roots.

\textit{Hypothesis:} Carter heard a pun.

\textit{Label:} \textbf{entailed}

\end{enumerate}
\end{multicols}
\noindent\hrulefill
\caption{Examples from tasks within the General Language Understanding Evaluation (GLUE) benchmark by \citeA{wangGLUEMultiTaskBenchmark2018}, Inference is Everything (IIE) by \citeA{whiteInferenceEverythingRecasting2017}, and the Diverse NLI Collection (DNC) by \citeA{poliak2018emnlp-DNC}. Each task is recast from preexisting data into classic RTE format.}
\label{fig:rte format examples}
\end{figure}

\vspace{5pt}
\noindent
{\em Cloze tasks.}
Another popular format is the cloze task, originally conceived by \citeA{taylorClozeProcedureNew1953}. Such a task typically involves the deletion of one or more words in a text, essentially requiring one to fill in the blank, usually from a set of choices, but not always, e.g., LAMBADA, which is an open-ended cloze task \cite{papernoLAMBADADatasetWord2016}. 
\textcolor{black}{This format expands upon the traditional language modeling task, i.e., the task of predicting the next word in a sequence of words, inspired by the need to calculate a probability distribution of sequences in language for problems like speech recognition \cite{bahlMaximumLikelihoodApproach1983} and machine translation \cite{brownStatisticalApproachMachine1990}.} 
The cloze format has been used in several recent NLI benchmarks, including CBT \cite{hillGoldilocksPrincipleReading2015}, the Story Cloze Test for the ROCStories benchmark \cite{mostafazadehCorpusClozeEvaluation2016}, CLOTH \cite{xieLargescaleClozeTest2017}, SWAG \cite{zellersSWAGLargeScaleAdversarial2018}, and ReCoRD \cite{zhangReCoRDBridgingGap2018}. These recent benchmarks provide anywhere from two to ten options to fill in the blank, and range from requiring the prediction of a single word to parts of sentences and entire sentences. Examples of cloze tasks are listed in Figure~\ref{fig:cloze examples}.

\begin{figure}
\noindent\hrulefill
\begin{multicols}{2}
\begin{enumerate}[label=\textbf{(\Alph*)}]
\item \textbf{CBT \cite{hillGoldilocksPrincipleReading2015}}
\tiny
\begin{enumerate}[label=\arabic* ,itemindent=0pt,leftmargin=0pt,itemsep=0em]
\item Mr. Cropper was opposed to our hiring you .
\item Not , of course , that he had any personal objection to you , but he is set against female teachers , and when a Cropper is set there is nothing on earth can change him .
\item He says female teachers ca n't keep order .
\item He 's started in with a spite at you on general principles , and the boys know it .
\item They know he 'll back them up in secret , no matter what they do , just to prove his opinions .
\item Cropper is sly and slippery , and it is hard to corner him . ''
\item `` Are the boys big ? ''
\item queried Esther anxiously .
\item `` Yes .
\item Thirteen and fourteen and big for their age .
\item You ca n't whip 'em -- that is the trouble .
\item A man might , but they 'd twist you around their fingers .
\item You 'll have your hands full , I 'm afraid .
\item But maybe they 'll behave all right after all . ''
\item Mr. Baxter privately had no hope that they would , but Esther hoped for the best .
\item She could not believe that Mr. Cropper would carry his prejudices into a personal application .
\item This conviction was strengthened when he overtook her walking from school the next day and drove her home .
\item He was a big , handsome man with a very suave , polite manner .
\item He asked interestedly about her school and her work , hoped she was getting on well , and said he had two young rascals of his own to send soon .
\item Esther felt relieved .
\item She thought that Mr. \_\_\_\_\_ had exaggerated matters a little .

\textit{Blank:} \textbf{Baxter}, Cropper, Esther, course, fingers, manner, objection, opinions, right, spite
\end{enumerate}
\normalsize

\item \textbf{ROCStories \cite{mostafazadehCorpusClozeEvaluation2016}} \newline
Karen was assigned a roommate her first year of college. Her roommate asked her to go to a nearby city for a concert. Karen agreed happily. The show was absolutely exhilarating.

\textit{Ending:}
\begin{enumerate}[label=\alph*.]
\item \textbf{Karen became good friends with her roommate.}
\item Karen hated her roommate.
\end{enumerate}

\item \textbf{CLOTH \cite{xieLargescaleClozeTest2017}} \newline
She pushed the door open and found nobody there. "I am the \_\_\_\_\_ to arrive." She thought and came to her desk.
\begin{enumerate}[label=\alph*.]
\item last
\item second
\item third
\item \textbf{first}
\end{enumerate}

\item \textbf{SWAG \cite{zellersSWAGLargeScaleAdversarial2018}} \newline
On stage, a woman takes a seat at the piano. She
\begin{enumerate}[label=\alph*.]
\item sits on a bench as her sister plays with the doll.
\item smiles with someone as the music plays.
\item is in the crowd, watching the dancers.
\item \textbf{nervously sets her fingers on the keys.}
\end{enumerate}

\item \textbf{ReCoRD \cite{zhangReCoRDBridgingGap2018}} \newline
... Daniela Hantuchova knocks Venus Williams out of Eastbourne 6-2 5-7 6-2 ...

\textit{Query:} Hantuchova breezed through the first set in just under 40 minutes after breaking Williams' serve twice to take it 6-2 and led the second 4-2 before \_\_\_\_\_ hit her stride.\newline
\textbf{Venus Williams}

\end{enumerate}
\end{multicols}
\noindent\hrulefill
\caption{Examples from cloze tasks. Answers in bold.}
\label{fig:cloze examples}
\end{figure}

\vspace{5pt}
\noindent
{\em Traditional multiple-choice tasks.}
If not in entailment or cloze form, classification tasks tend to be formulated as traditional multiple-choice questions. Benchmarks which use this format include COPA \cite{roemmeleChoicePlausibleAlternatives2011}, the Winograd Schema Challenge \cite{davisWinogradSchemaChallenge2018a}, and MCScript \cite{ostermannMCScriptNovelDataset2018}. Two-way and four-way decision questions are the most common among the surveyed multiple-choice benchmarks.

\paragraph{Open-ended tasks.}
On the other hand, some benchmarks require more open-ended responses rather than providing a small list of alternatives to choose from. Answers may be restricted to spans of a given text, e.g., SQuAD \cite{rajpurkarSQuAD1000002016,rajpurkarKnowWhatYou2018} or QuAC \cite{choiQuACQuestionAnswering2018}. They may be less restricted to a subset of a large number of category labels, e.g., the Maslow, Reiss, and Plutchik tasks in Story Commonsense \cite{rashkinModelingNaivePsychology2018}. And of course, they may be purely open-ended, e.g., Event2Mind \cite{rashkinEvent2MindCommonsenseInference2018}, NarrativeQA \cite{kociskyNarrativeQAReadingComprehension2018}, or DROP \cite{duaDROPReadingComprehension2019}. 
Examples of these open-ended formats are listed in Figure~\ref{fig:oe examples}.

\begin{figure}
\noindent\hrulefill
\begin{multicols}{2}
\begin{enumerate}[label=\textbf{(\Alph*)}]
\item \textbf{SQuAD 2.0 \cite{rajpurkarKnowWhatYou2018}} \newline
In meteorology, precipitation is any product of the condensation of atmospheric water vapor that falls under gravity. The main forms of precipitation include drizzle, rain, sleet, snow, graupel and hail... Precipitation forms as smaller droplets coalesce via collision with other rain drops or ice crystals within a cloud. Short, in-tense periods of rain in scattered locations are called "showers".

What causes precipitation to fall? \newline\textbf{gravity}

\item \textbf{bAbI \cite{westonAICompleteQuestionAnswering2016}} \newline
The kitchen is north of the hallway.\newline
The bathroom is west of the bedroom.\newline
The den is east of the hallway.\newline
The office is south of the bedroom.

How do you go from den to kitchen? \newline
\textbf{west, north}

\item \textbf{SC \cite{rashkinModelingNaivePsychology2018}\footnotemark} \newline
Valerie was getting ready for a formal dance. She had been preparing for hours. As she was ready to leave, her acrylic nail broke. She snapped off all of her faux nails.

\textit{Maslow:} \textbf{esteem, stability}\newline
\textit{Reiss:} \textbf{status, approval, order} \newline
\textit{Plutchik:} \textbf{surprise, sadness, disgust, anger}

\item \textbf{DROP \cite{duaDROPReadingComprehension2019}} \newline
That year, his Untitled (1981), a painting of a haloed, black-headed man with a bright red skeletal body, depicted amid the artists signature scrawls, was sold by Robert Lehrman for \$16.3 million, well above its \$12 million high estimate.

\textit{Question:} How many more dollars was the Untitled (1981) painting sold for than the 12 million dollar estimation? \newline
\textbf{4300000}

\end{enumerate}
\end{multicols}
\noindent\hrulefill
\caption{Examples from open-ended response tasks. Answers in bold.}
\label{fig:oe examples}
\end{figure}
\footnotetext{Example extracted from Story Commonsense test data available at \url{https://uwnlp.github.io/storycommonsense/}}

\subsubsection{Evaluation Schemes}\label{sec:eval schemes}
As mentioned earlier, the Turing Test \cite{turingComputingMachineryIntelligence1950} has long been criticized by AI researchers as it does not seem to truly evaluate machine intelligence. As described by \citeA{ortizWhyWeNeed2016}, there is a critical need for new intelligence benchmarks to support incremental development and evaluation of AI techniques. 
These benchmarks should not merely provide a pass or fail grade, rather they should provide feedback on a continuous scale which enables both incremental development and comparison of approaches. One key consideration for these benchmarks is informative evaluation metrics that are objective and easy to calculate. These metrics can be used to compare different approaches and compare machine performance against human performance. 

\paragraph{Evaluation metrics.}
Choice of evaluation metrics is highly dependent on the type of task, and thus so is the difficulty of calculating them. 
Classification tasks often use exact-match accuracy if correct answers or class labels are evenly distributed through benchmark data. If this is not the case, common practice is to additionally present F-measure as an evaluation metric \cite{wangGLUEMultiTaskBenchmark2018}. The precision and recall may also be presented, however the F-measure (which considers both in its calculation) is much more common in the recent surveyed benchmarks. Multiple-choice and classification task formats such as RTE, cloze, and traditional multiple-choice can all use these metrics.

More open-ended tasks are by nature more difficult to evaluate, but they can still be objective and informative. Tasks like SQuAD \cite{rajpurkarSQuAD1000002016} or QuAC \cite{choiQuACQuestionAnswering2018}, where answers can only be spans of a provided text, can be evaluated similarly to multiple-choice tasks. Exact-match accuracy and F-measure are used as evaluation metrics on both of these benchmarks, where the collection of tokens in the predicted and true spans (excluding punctuation and articles) are compared. Where answers are instead a subset of a large group of category labels, e.g., the Maslow, Reiss, and Plutchik tasks in Story Commonsense, evaluation is similar, but for these benchmarks particularly, precision and recall are additionally calculated \cite{rashkinModelingNaivePsychology2018}.

Where multiple purely open-ended responses are given, as in Event2Mind \cite{rashkinEvent2MindCommonsenseInference2018}, evaluation is more difficult. Event2Mind particularly uses the average cross-entropy and the "recall @ 10," i.e., the percentage of times human-produced ground truth labels fall within the top 10 generated predictions from a system. In bAbI \cite{westonAICompleteQuestionAnswering2016}, LAMBADA \cite{papernoLAMBADADatasetWord2016}, and CoQA \cite{reddyCoQAConversationalQuestion2018}, where open-ended responses are compared to a single ground truth answer, exact-match accuracy and F-measure are used. These benchmarks are able to use such exact evaluation measures because responses are short. In bAbI particularly, correct responses are limited to one word or lists of words, and in LAMBADA, responses are only a single word to fill a blank. Such restrictions are essential for such simple and accurate evaluation of open-ended responses. \textcolor{black}{For longer generated responses, e.g., as required in NarrativeQA \cite{kociskyNarrativeQAReadingComprehension2018}, evaluation metrics for machine translation like BLEU \cite{papineniBleuMethodAutomatic2002} or Meteor \cite{denkowskiMeteorAutomaticMetric2011}, or similar metrics for text summarization like ROUGE \cite{lin:2004:ACLsummarization}, are often used to compare answers with gold responses. However, many such metrics have been found to not correlate well with human judgments of generated text \cite{belz-reiter-2006-comparing,elliott-keller-2014-comparing}, so they should be interpreted with caution.}

\paragraph{Baseline performance.}
Benchmarks provide common datasets and experimental setups for researchers to compare different approaches. When a new benchmark is first released, \textcolor{black}{results from simple baseline approaches are often reported. Ideally, the benchmark should be designed in a way that a simple baseline would achieve poor performance to ensure that it cannot be solved by trivial means, instead requiring a more advanced approach.}
For classification problems, simple baseline approaches are often calculated by random choice, choosing the majority class from the training data, or choosing the alternative with the highest overlap in n-grams with the question or provided text \cite{richardsonMCTestChallengeDataset2013}. Examples of these baseline approaches can be found in the Story Cloze Test baselines, which include most of these approaches and more \cite{mostafazadehCorpusClozeEvaluation2016}. For open-ended problems, shallow lexical approaches (e.g., using language models) may again be used, as in the bAbI benchmark\cite{westonAICompleteQuestionAnswering2016}. 
\textcolor{black}{When releasing a new benchmark, results from one or more state-of-the-art models for similar existing benchmarks will also typically be reported as baselines. 
Later, as more advanced approaches are developed for the benchmark, they are often used as state-of-the-art baselines on the next generation of benchmarks. This cycle continually boosts baseline performance over time, and drives the development of new and improved models as well as new benchmarks. }

\paragraph{Human performance measurement.}
To evaluate the progress of machine intelligence, human performance on benchmark tasks is often measured to provide a reference point. \textcolor{black}{Human performance may be measured in a few different ways. A typical way is to select several human judges, which may be crowdsourced, to solve a random subset of the benchmark data. Performance may be averaged across judges, e.g., as done for CBT \cite{hillGoldilocksPrincipleReading2015}, or calculated through a majority vote of all judges, e.g., for SWAG \cite{zellersSWAGLargeScaleAdversarial2018}. Human performance can also be measured through annotator agreement, e.g., as done for ROCStories \cite{mostafazadehCorpusClozeEvaluation2016}. An ``expert'' human performance measurement may be additionally provided as a ceiling, typically by testing researchers involved with creating the dataset. This was done, for example, by the authors of RACE~\cite{laiRACELargescaleReAding2017} and SWAG. } 

\textcolor{black}{In many benchmarks, one measure for computational models is to assess how close it comes to or whether it exceeds human performance. When a system exceeds human performance, this seems to suggest that the system has acquired the skills and knowledge required by the benchmark. However, due to the unexplainability of many neural models, this is often difficult to prove. Further, data biases have recently been found in many popular benchmarks \cite{gururanganAnnotationArtifactsNatural2018}, allowing both simple approaches \cite{schwartzEffectDifferentWriting2017} and state-of-the-art neural systems \cite{nivenProbingNeuralNetwork2019} to exploit the biases and achieve artificially high performance. 
This puts these unexplainable systems under higher scrutiny.}

\subsubsection{Data Biases}\label{sec:bias}
When creating benchmarks, one challenge is the bias of data unintentionally introduced to the benchmark. For example, in the first release of the Visual Question Answering (VQA) benchmark \cite{agrawalVQAVisualQuestion2017}, researchers found that machine learning models were learning several statistical biases in the data, and could answer up to 48\% of questions in the validation set without seeing the image \cite{manjunathaExplicitBiasDiscovery2018}. This artificially high system performance is problematic, as it is not accredit to the underlying technology. Here, we summarize several key dimensions of biases encountered in previous NLI research. Some of these (e.g., class label distributions) are easier to avoid, while others (e.g., hidden correlation biases) are more difficult to address.

\paragraph{Label distribution bias.} Class label distribution bias is the easiest to avoid. \textcolor{black}{In multiple-choice problems, ordinal positions of correct answers should be entirely randomized so that each possible choice appears in benchmark data in a uniform distribution. For other classification tasks where class labels have specific meanings, e.g., textual entailment, ensuring balance of class labels must occur earlier, i.e., when creating examples. Care should be taken to ensure an equal number of examples are generated for each class.} When class labels are balanced, a majority-class baseline will score as low as possible on the task. While binary-choice tasks should have a 50\% majority class baseline, the MegaVeridicality subtask within DNC \cite{poliak2018emnlp-DNC} has a 67\% majority-class baseline due to unevenly distributed class labels, leaving significantly less room for incremental improvement than benchmarks with lower-performing simple baselines.

\paragraph{Question type bias.}
For benchmarks involving question answering, previous work has made effort to balance the types of questions, especially if questions are generated by crowdsourcing. This will ensure a broad domain of knowledge and reasoning required to solve the benchmark. A fairly simple method to keep a balance of question types is to calculate the distribution of the first words of each question, as the creators of CoQA \cite{reddyCoQAConversationalQuestion2018} and CommonsenseQA \cite{talmorCommonsenseQAQuestionAnswering2019} did. One could also manually label and analyze a random sample of questions with categories relating to the types of knowledge or reasoning required, or have crowd workers perform the labeling if an expert is not essential. Examples of this are shown by the creators of SQuAD 2.0 \cite{rajpurkarKnowWhatYou2018} and ARC \cite{clarkThinkYouHave2018}. To entirely avoid question type biases, implementing a standard set of questions for all provided texts may be beneficial. ProPara does this for all entities in its procedural paragraphs \cite{mishraTrackingStateChanges2018}, limiting questions about each entity to whether it is created, destroyed, or moved during the paragraph, and when and where this happens. \citeA{manjunathaExplicitBiasDiscovery2018} suggest that biases can further be avoided in VQA benchmarks by forcing questions to require a particular skill (e.g., telling time or subtracting) to be answered. This rule of thumb can be applicable for textual benchmarks as well, e.g., in the DROP benchmark \cite{duaDROPReadingComprehension2019}, which requires discrete reasoning skills like arithmetic to answer questions about passages.

\paragraph{Superficial correlation bias.}
The kind of biases most difficult to discern and avoid perhaps are those caused by accidental correlations between features of answers and questions. One example of this is gender bias, which NLI systems are particularly vulnerable to when trained on biased data. \citeA{gender-bias-in-coreference-resolution} highlight this problem in coreference resolution, showing that systems trained on gender-biased data perform worse in gender pronoun disambiguation. For example, consider the problem from their Winogender dataset in Figure~\ref{fig:coreference examples}: "The paramedic performed CPR on the passenger even though she knew it was too late." In determining who \textit{she} is, systems trained on gender-biased training data may be more likely, for example, to incorrectly choose \textit{the passenger} rather than \textit{the paramedic} due to male gender pronouns appearing in training data more commonly in the context of this occupation than female gender pronouns. To avoid this, gender pronouns should appear equally frequently among other words in training data, especially those related to occupations and activities. Similar gender biases are identified in Event2Mind data, which are derived from movie scripts \cite{rashkinEvent2MindCommonsenseInference2018}.

When authoring natural language data (e.g., generating questions or hypotheses), some human stylistic artifacts such as predictable sentence structure, presence of certain linguistic phenomena, and vocabulary use can also cause these superficial correlation biases. \textcolor{black}{This is most often the case if data is authored by crowd workers \cite{gururanganAnnotationArtifactsNatural2018}, but similar biases have been found in expert-authored data as well \cite{trichelair2018reasonable}.} In the Story Cloze Test \cite{mostafazadehCorpusClozeEvaluation2016}, systems are presented a plausible and implausible ending to the story, and must choose which ending is plausible. However, previous work \citeA{schwartzEffectDifferentWriting2017} has shown that the Story Cloze Test can be solved with up to 72.4\% accuracy by only looking at the two possible endings. They do this by exploiting human writing style biases in the possible endings rather than performing a more realistic reasoning. For example, they find that negative language is used more commonly in the wrong ending (e.g., ``hates''), and the correct ending is more likely to use enthusiastic language (e.g., ``!''). An example of a biased negative ending is seen in Figure~\ref{fig:cloze examples}. \citeA{sharmaTacklingStoryEnding2018} have begun work to update the benchmark data and remove these biases.

While generating the Story Cloze Test data was not a fast or simple task for crowd workers, \citeA{gururanganAnnotationArtifactsNatural2018} suggest that such biases can come from crowd workers' adoption of predictable annotation strategies and heuristics to quickly generate data. These strategies have been revealed for several textual entailment benchmarks which consist of pairs of short sentences. For example, on the entailment task in SemEval 2014 \cite{marelliSemEval2014TaskEvaluation2014} as part of the SICK benchmark \cite{marelliSICKCureEvaluation2014}, \citeA{laiIllinoisLHDenotationalDistributional2014} found that the presence of negation in an example was associated with the appearance of the contradiction class label. Their trained classifier using this feature alone was able to achieve 61\% accuracy. Later, \citeA{poliakHypothesisOnlyBaselines2018} and \citeA{gururanganAnnotationArtifactsNatural2018} found the presence of particular words in the hypothesis sentence can bias the entailment prediction in several entailment benchmarks. For example, ``nobody'' in contradictory examples from SNLI \cite{bowmanLargeAnnotatedCorpus2015} was found to be an indicator of contradiction, while generic words like ``animal'' and ``instrument'', as well as gender-neutral pronouns, were found to be indicators of entailment. \citeauthor{gururanganAnnotationArtifactsNatural2018} further find that a high sentence length is an indicator of a neutral entailment relationship, and suggest that crowd workers often remove words from the context sentences to create entailed hypothesis sentences. Using biases like these, a baseline approach by \citeauthor{poliakHypothesisOnlyBaselines2018} which only used the hypothesis sentence from entailment benchmarks was able to outperform a majority-class baseline in SNLI, JOCI \cite{zhangOrdinalCommonsenseInference2016}, SciTail \cite{khotSciTailTextualEntailment2018}, two tasks within Inference is Everything \cite{whiteInferenceEverythingRecasting2017}, and the MultiNLI task within GLUE \cite{williamsBroadCoverageChallengeCorpus2017,wangGLUEMultiTaskBenchmark2018}. 

\paragraph{Addressing superficial correlation bias.}
\textcolor{black} {Various approaches have been developed to address this data bias problem. For example, to recognize biases, a simple technique is to calculate the mutual information between words and prediction classes within benchmark examples. This was performed by researchers in discovering stylistic biases in entailment benchmarks \cite{gururanganAnnotationArtifactsNatural2018}. This type of analysis should be performed on any new benchmark data when it is created.}

\textcolor{black} {To avoid the biases, more involved techniques may be required. For example, in creating the SWAG benchmark \cite{zellersSWAGLargeScaleAdversarial2018}, a novel adversarial filtering process was introduced to ensure writing styles are consistent among ending choices, and the correct answer cannot be identified by exploitative stylistic classifiers.} \textcolor{black}{This is done by over-generating negative choices for each question with a language model, then using a collection of strong classifiers to filter out negative examples which are too easy to identify given an arbitrary train-test split. } 
\textcolor{black}{
When SWAG was released, it was challenging to state-of-the-art classifiers. 
But it did not take long before a new model solved SWAG with a high accuracy \cite{devlinBERTPretrainingDeep2018}. 
By incorporating this new model into the adversarial filtering, as well as using a newer language model to generate text,
\citeA{zellers2019hellaswag} created an updated version called HellaSWAG, on which the new state-of-the-art model struggled.}
\textcolor{black} {Variants of adversarial filtering were used in creating other benchmarks such as WinoGrande \cite{sakaguchi2019winogrande} and AlphaNLI \cite{ch2019abductive} to respectively remove statistical biases and select the most challenging pairs of choices from a pool of hypotheses, resulting in low performance from state-of-the-art baseline approaches.}
\textcolor{black}{As this iterative process of discovering stronger classifiers and in turn strengthening the adversarial filtering continues, an important question will be when all of the statistical biases are removed, or if it will ever be possible to remove all statistical biases.}
\textcolor{black}{Nonetheless, given these recent successes, adversarial filtering should continue to be used on new benchmarks to challenge the latest state-of-the-art models, and perhaps be used to strengthen existing benchmarks where possible.}

\textcolor{black}{Another technique to avoid these biases proposed by \citeA{belinkovDonTakePremise2019a}, particularly in textual entailment problems, is to train a model to predict the premise sentence given the hypothesis sentence and class label.} \textcolor{black}{This training setup ensures that the model ignores unexpected language artifacts in the premise, attempting to remove superficial statistical biases.}
\textcolor{black}{While the development of these paradigms to avoid data bias is promising, a continuous effort in finding and improving techniques to avoid these kinds of biases will be important as the state of the art continues to evolve.}

\subsubsection{Collection Methods}
Methods to collect benchmark data ideally should be cost-efficient and should result in high-quality and unbiased data. Both manual and automatic approaches have been applied. Manual curation of data can be done by experts or researchers, or through crowd workers, which comes with its own set of considerations. Automatic approaches often involve automatically generating data by applying language models or automatically extracting or mining data from existing resources. As detailed in Appendix~\ref{sec:appendix}, a benchmark is often created through a combination of these approaches. For the rest of this section, we summarize pros and cons for these different approaches.

\paragraph{Manual versus automatic generation.}
Until recently, many of the existing benchmarks were created manually by groups of experts. This may involve tedious processes like manually collecting data from other corpora or the Internet, e.g., the first RTE Challenge \cite{daganPASCALRecognisingTextual2005}, or authoring most data from scratch, e.g., the Winograd Schema Challenge \cite{levesqueWinogradSchemaChallenge2011,levesqueWinogradSchemaChallenge2012,morgensternWinogradSchemaChallenge2015,morgensternPlanningExecutingEvaluating2016}. This approach ensures data is high quality and thus typically requires little validation, however it is not scalable. These datasets are often small compared to those created with other approaches.

Recent advances in NLP make it possible to automatically generate textual data (e.g., natural language statements, questions, etc.) for benchmark tasks. Although this approach is scalable and efficient, the quality of data varies, and often depends directly on the language model used. For example, in bAbI \cite{westonAICompleteQuestionAnswering2016}, as agents interacted with objects in a virtual world and with each other, examples were automatically generated. This method ensures that produced data are sensible to the constraints of the physical world. However, as the questions and answers are written with simple structure, the data is easily understood by machines. Most of the dataset is solved with 100\% accuracy just by baseline systems. Consequently, the bAbI tasks are often considered as toy tasks. A more sophisticated rule-based method for probabilistically generating text data which encourages more diverse language without any added biases is presented by \citeA{manningRealworldVisualReasoning2018}. Though automatic natural language generation methods are improving, it is likely that such approaches will still require some manual validation.

\paragraph{Automatic generation versus text mining.}
As millions of natural language texts are publicly available on the Internet and in existing datasets, it is possible to build language benchmarks from these texts by automatically mining texts and extracting sentences. This process is most successful when the information source is created by experts and highly accurate. For example, in the CLOTH \cite{xieLargescaleClozeTest2017} benchmark, data instances were mined from fill-in-the-blank English tests created by human teachers. While other automatically generated cloze tasks like CBT \cite{hillGoldilocksPrincipleReading2015} choose missing words mostly randomly, CLOTH is more challenging as the missing word in each example was chosen by an expert. For many other benchmarks that are built by mining less reliable or consistent sources, there is often a need for automatic or human validation or filtering, e.g., in creating SWAG \cite{zellersSWAGLargeScaleAdversarial2018}, which was mined in part from other corpora, such as the Large Scale Movie Description Challenge \cite{rohrbachMovieDescription2017} and the captions in ActivityNet \cite{heilbronActivityNetLargescaleVideo2015}.

\paragraph{Crowdsourcing considerations.}
Acquiring language data directly from crowd workers 
has become more feasible in recent years due to the growth of crowdsourcing platforms, e.g., Amazon Mechanical Turk. Crowdsourcing has enabled researchers to create larger datasets than ever before, however it comes with a set of considerations relating to task complexity, worker qualification, data validation, and cost optimization.

\vspace{5pt}
\noindent
{\em Task complexity.}
When creating a crowdsourcing task, it is important to consider the difficulty level of what crowd workers will be expected to do. When given overly complicated instructions, the non-expert crowd workers may find it difficult understanding the instructions and keeping them in mind while working on the task. Easy crowdsourcing tasks typically involve quick pass/fail validation or relabeling of data, e.g., in validating SNLI \cite{bowmanLargeAnnotatedCorpus2015}. Difficult crowdsourcing tasks usually require crowd workers to write large texts, e.g., in creating ROCStories \cite{mostafazadehCorpusClozeEvaluation2016}, where workers wrote five-sentence stories following a fairly elaborate set of restrictions on story content to ensure stories were high-quality, focused, and well-organized. Such restrictions must be explained briefly and clearly, and it may take several pilot studies to ensure workers understand and follow the instructions correctly \cite{mostafazadehCorpusClozeEvaluation2016}.

\vspace{5pt}
\noindent
{\em Worker qualification.}
Regardless of task difficulty, efforts should be made to avoid workers submitting invalid data, whether they are trolls or unable to follow directions. This may be done through some sort of qualification task, perhaps requiring a prospective worker to recognize examples of acceptable submissions \cite{mostafazadehCorpusClozeEvaluation2016}, or testing a prospective worker's grammar \cite{richardsonMCTestChallengeDataset2013}. It may also be worthwhile to identify excellent workers, and reward them and/or recruit them for more work \cite{mostafazadehCorpusClozeEvaluation2016}. In our own experience with crowdsourcing, we have found that if a worker produces one invalid submission, all of the worker's submissions will likely be invalid, and thus the worker should be rejected and potentially banned from the task. On the other hand, if a worker produces an excellent submission, all of the worker's submissions will likely be excellent.

\vspace{5pt}
\noindent
{\em Data validation.}
Even though crowdsourced data is produced by non-experts, data can just as easily be validated by non-experts. In creating ROCStories, \citeA{mostafazadehCorpusClozeEvaluation2016} employ several novel methods of crowdsourced validation of crowdsourced data. Involved validation is especially necessary for difficult crowdsourcing tasks like the authoring of ROCStories, which required crowd workers to write long texts following strict guidelines. Crowdsourced data validation typically just requires a separate group of workers to review generated data and identify any bad examples, e.g., the validation of DNC benchmark data \cite{poliak2018emnlp-DNC}. For labeling tasks, multiple crowd workers can label the same examples, and agreement can be measured from this to estimate data quality, e.g., in creating the JOCI benchmark \cite{zhangOrdinalCommonsenseInference2016}.

For complex writing tasks where both data authoring and validation are highly involved, it may be advantageous to present crowdsourced tasks to two interacting workers at once. \citeA{reddyCoQAConversationalQuestion2018}, for example, do this to create CoQA from actual human conversations about provided passages on Amazon Mechanical Turk, and achieve high data quality with minimal validation. In creating the data, the two interacting workers validate each others' work, and can even report workers who do not follow instructions, reducing the burden of worker qualification.

\vspace{5pt}
\noindent
{\em Cost optimization.}
Though hiring crowd workers is typically cheaper than hiring permanent workers, the cost may still be limiting, especially if the creator wishes to properly evaluate the quality of generated data through verification by even more crowd workers. For example, ROCStories \cite{mostafazadehCorpusClozeEvaluation2016}, consisting of about 50,000 well-evaluated five-sentence stories and 13,500 test cases, cost an average of 26 cents per story and an extra 10 cents per test case, resulting in a total cost near 15,000 USD to generate the dataset. If the cost of such thorough validation is an issue, validating a random sample of produced data, e.g., in validating SNLI \cite{bowmanLargeAnnotatedCorpus2015}, can serve as an indicator of the overall quality of benchmark data.

Ultimately, each method of data collection has its own advantages and drawbacks. While manual authoring results in high-quality, expert-verified data, it is slow and unscalable. Meanwhile, the quality of automatically authored data is highly dependent on the language model used, and though it is faster, it may require manual verification. If using text mining rather than generating from scratch, data is more likely to be representative of human language, however manual verification may still be necessary depending on the source which data are extracted from. And lastly, crowdsourcing is a quick and convenient way to collect human-authored data following any set of criteria or restrictions, but it comes with special considerations that address the difficulty of work, qualification of workers, data validation, and cost optimization. When developing a new benchmark, the above trade-offs will need to be carefully considered.

\subsubsection{Benchmark Complexity}\label{sec:complexity}
\textcolor{black}{Given the large variety of available benchmarks for natural language inference, some understanding of the complexity or difficulty of these benchmarks would be helpful.
However, as demonstrated earlier in this section, there are many variations among these benchmarks. Consequently, it is not feasible to have a fair and direct comparison. 
As such, we do not attempt to address whether one benchmark is more complex than another. 
But rather we discuss various dimensions that may affect the perception of the complexity of a benchmark, and consider their possible implications.\footnote{In this analysis, data was collected from papers referenced in Section~\ref{sec:existing benchmarks} and leaderboards linked to in Section~\ref{sec:approaches} at time of writing (October 2019). Additional data collected from the leaderboards for AlphaNLI (found at \url{https://leaderboard.allenai.org/anli/submissions/public}), the Story Cloze Test (found at \url{https://competitions.codalab.org/competitions/15333\#results}), and SocialIQA (found at \url{https://leaderboard.allenai.org/socialiqa/submissions/public}), which are not linked to in Section~\ref{sec:approaches}.}}

\begin{figure}
  \centering
  \includegraphics[width=1\textwidth]{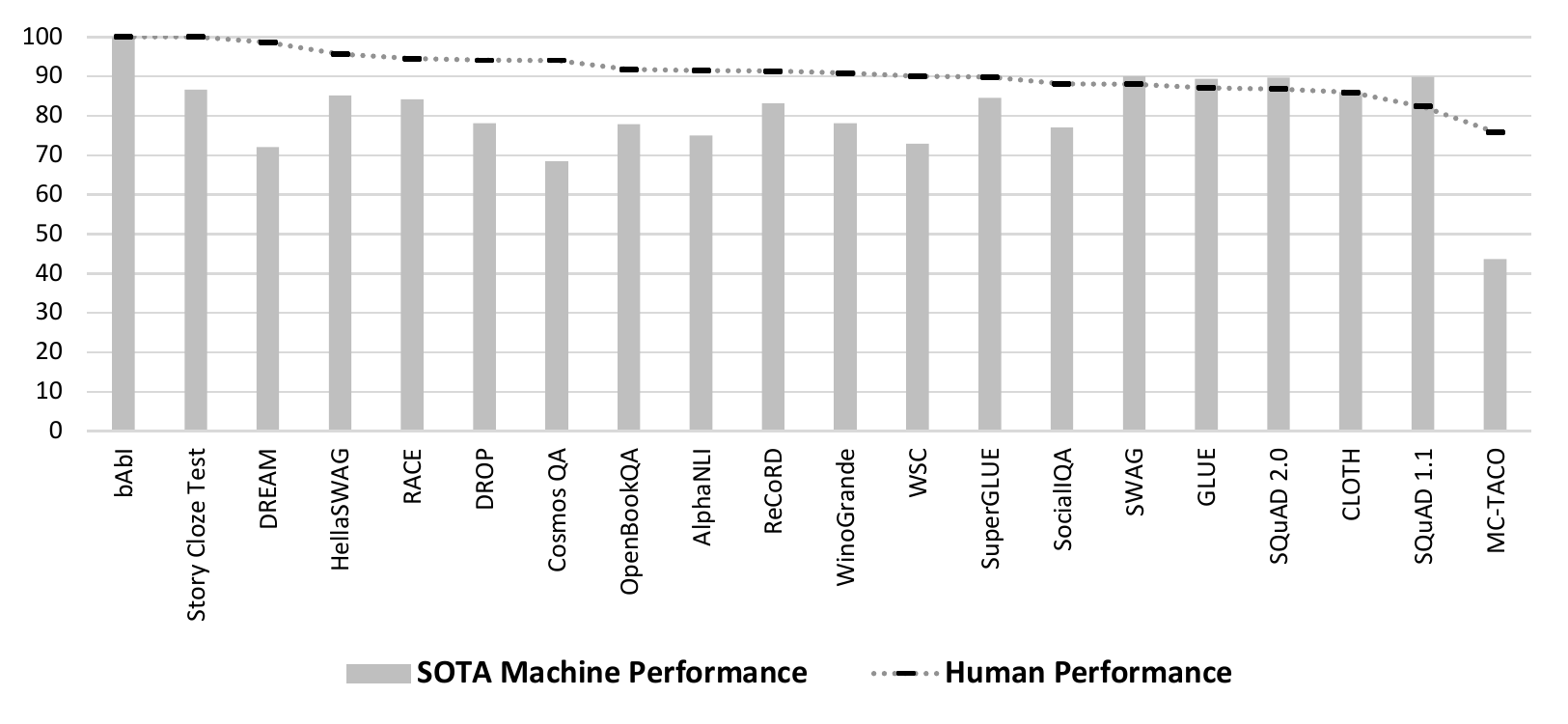}
  \caption{Comparison of state-of-the-art accuracy to human accuracy on selected benchmarks where human performance is reported. }
  \label{fig:difficulty hp}
\end{figure}

\paragraph{Perception of complexity.}
\textcolor{black}{One natural way to think about the complexity of a task is to consider the performance of both machines and humans on the task; low performance may suggest high complexity, and vice versa. Figure~\ref{fig:difficulty hp} shows human performance and state-of-the-art machine performance for selected benchmarks. This figure seems to suggest that bAbl \cite{westonAICompleteQuestionAnswering2016} is easiest for both the human and the machine. This may not be surprising, as the bAbl benchmark data is synthetically generated with simple structure, and these simpler language patterns can be well captured by deep learning models. For the majority of benchmarks, human performance is between 80-90\% accuracy. Machine performance has a much larger variance among these benchmarks. For some benchmarks, e.g., SQuAD 1.1 \cite{rajpurkarSQuAD1000002016} and SWAG \cite{zellersSWAGLargeScaleAdversarial2018}, machine performance actually exceeds human performance. Does that mean these problems are easy and solved? On the other hand, those benchmarks where machine performance exceeds human performance have relatively low human performance. Does this suggest that the problems are actually difficult, since they seem difficult for humans? It is important to note that these performance measures, whether human or machine, may allow us to form some perception of task complexity, but they are not perfect measures for several reasons, some of which are discussed below.}

\vspace{5pt}
\noindent
{\em Considerations for human performance measurements.}
\textcolor{black}{Benchmark design choices, such as the evaluation framework, can affect human performance measurements. If a benchmark has high variability in possible responses, human performance is likely to be lower, but this does not necessarily mean the benchmark is extraordinarily difficult. An example of this is SQuAD 1.1, which takes a span of a passage as an answer, and has a human accuracy of 77.0\% \cite{rajpurkarSQuAD1000002016}. While this may suggest that the task is difficult, the state-of-the-art performance has actually far exceeded human performance, suggesting the opposite. When analyzing human responses, the authors found that errors typically occurred when humans omitted unnecessary words or phrases from their answer spans. Thus, the low human performance may have largely resulted from limitations of the evaluation framework rather than the difficulty of the task.}

\textcolor{black}{The method of collecting human performance should also be considered. Human performance may be lower than expected if the measurement is crowdsourced, which is typical for recent benchmarks. Crowd workers want to complete tasks quickly to maximize their earnings, and thus may not spend as much time reasoning through the examples as they would in an everyday setting. Because of this, as discussed in Section~\ref{sec:eval schemes}, an additional human performance measurement collected from authors of the dataset or other experts is sometimes provided as a ceiling. However, the difference between performance of crowd workers and experts can be high, e.g., 73.5\% versus 94.5\% accuracy on RACE~\cite{laiRACELargescaleReAding2017}. An ensemble of crowd workers may also be used to generate a higher human performance measure. Again, though, the disparity between the performance of one worker and an ensemble of workers can be high, e.g., 82.8\% to 88.0\% accuracy on SWAG~\cite{zellersSWAGLargeScaleAdversarial2018} from one worker to an ensemble of five. This uncertainty raises questions about the naturalness of these performance measurement strategies, and these factors must be considered when interpreting human performance where crowdsourcing is involved.}

\vspace{5pt}
\noindent
{\em Considerations for machine performance measurements.}
\textcolor{black}{Some tasks may actually be more difficult for humans than for machines based on their individual capabilities, which should be considered when comparing their performance. An example of this is CBT, whose language modeling sub-tasks require predicting random words in a sentence with or without a preceding context paragraph \cite{hillGoldilocksPrincipleReading2015}. Predicting some types of words, such as prepositions, is difficult for humans (67.6\% accuracy without context), perhaps because humans excel at inferring meaning, and many prepositions have similar meanings. Consequently, neural language models, more capable of capturing patterns, easily beat humans in this sub-task of CBT. While the task is actually difficult for humans, machines are already more capable of performing it, making it less interesting of a problem to study.} 

\textcolor{black}{On the other hand, a high machine performance does not always mean that the machine can truly perform better than humans on these tasks. As discussed in Section~\ref{sec:bias}, the presence of statistical biases in benchmark data has made such a measure questionable. Further, as state-of-the-art neural approaches lack interpretability (and benchmarks do not require it), there is no way to prove this. As work progresses on eliminating these biases and providing greater explanation ability to state-of-the-art models, machine performance measurements will become more trustworthy.}

\paragraph{Factors affecting perception of complexity.}
\textcolor{black}{Beyond those limitations which must be considered when interpreting machine performance, some external factors 
can also affect the perceived complexity of the tasks. Apart from variations in the specific techniques or approaches, which will be discussed in Section~\ref{sec:approaches}, we find the {\em training data size} and the {\em benchmark age and popularity} most notable.} 

\begin{figure}
  \includegraphics[width=1\textwidth]{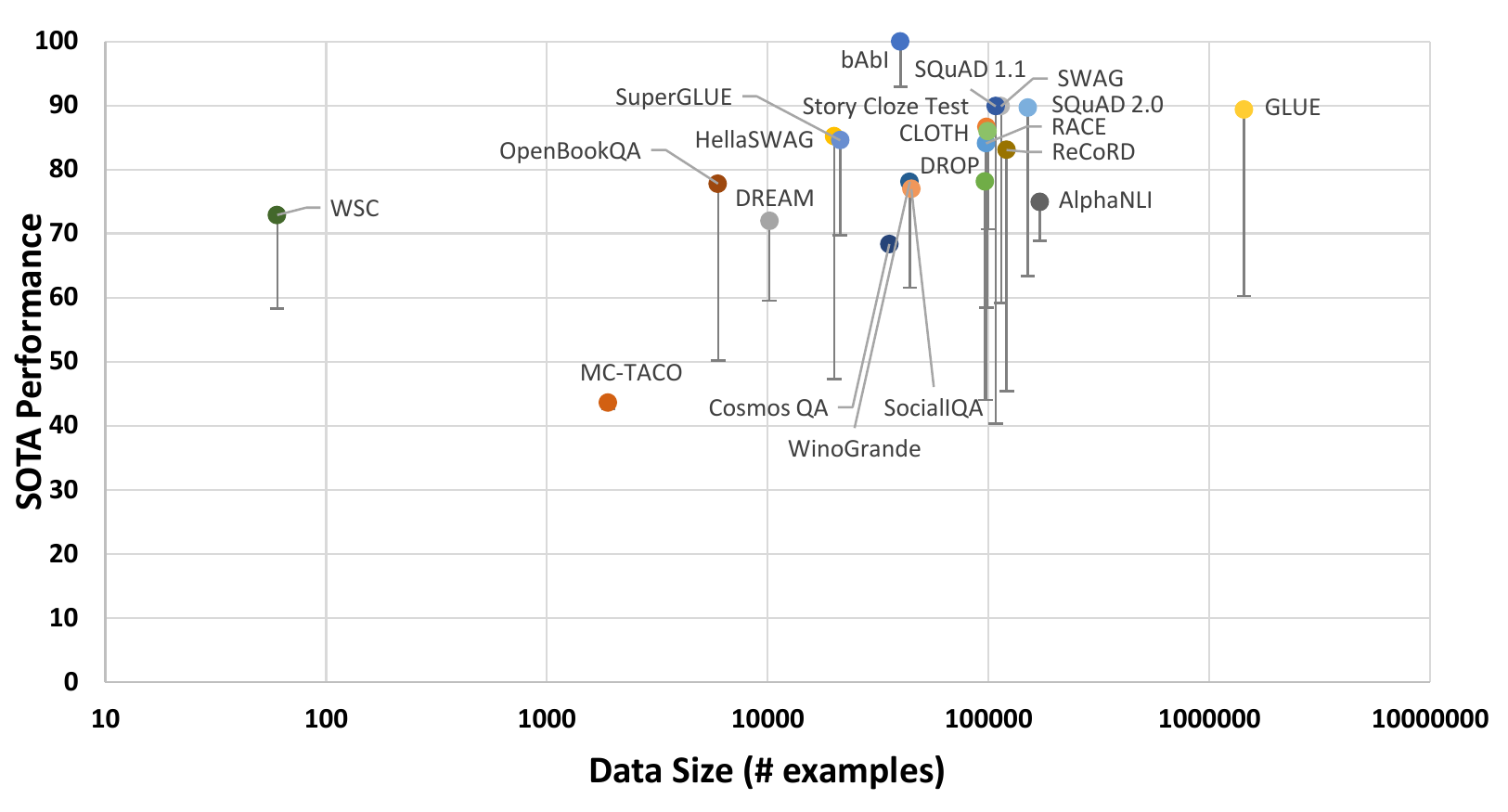}
  \caption{Data size versus state-of-the-art accuracy for selected benchmarks. Error bars indicate improvement from the original state-of-the-art accuracy reported by baselines in paper.}
  \label{fig:difficulty datasize}
\end{figure}

\vspace{5pt}
\noindent
{\em Training data size.}
\textcolor{black}{
Benchmarks with a high magnitude of training data are well-suited for deep neural networks, which can learn various patterns in language, especially in datasets with statistical biases as discussed in Section~\ref{sec:bias}. Figure~\ref{fig:difficulty datasize} compares overall benchmark data size with state-of-the-art accuracy on several benchmarks, including the improvement of the state of the art since each benchmark's release.}

\textcolor{black}{In the graph, performing a linear regression indicates no statistically significant correlation, however we can qualitatively interpret the data. With the exception of a few data points, we see a slightly positive trend in both the absolute state-of-the-art performance and the difference from the baseline performance. It appears that the highest performance and most improvement has come in benchmarks with greater than 10,000 examples. On small benchmarks with less than 1,000 examples, such as the Winograd Schema Challenge \cite{davisFirstWinogradSchema2017}, we have seen limited improvement, perhaps because neural networks struggle to converge on such small data. Conversely, on large benchmarks with over 100,000 examples, such as SQuAD 1.1 \cite{rajpurkarSQuAD1000002016}, SWAG \cite{zellersSWAGLargeScaleAdversarial2018}, and GLUE \cite{wangGLUEMultiTaskBenchmark2018}, we see the highest state-of-the-art performance, and some of the biggest improvements in state-of-the-art performance since release.}

\textcolor{black}{While this is intuitive, we do see some anomalies. For example, while bAbI \cite{westonAICompleteQuestionAnswering2016} has a relatively small magnitude of data, state-of-the-art accuracy has reached 100\%. This is likely because the data in this benchmark is synthetically generated and predictably structured, another potential factor that can make a benchmark easier. We also see large benchmarks where performance or improvement has been limited, e.g., the Story Cloze Test~\cite{mostafazadehCorpusClozeEvaluation2016} and Cosmos QA~\cite{huang2019cosmos}, and we predict that the age or popularity of a benchmark may factor into this.}

\begin{figure}
  \centering
  \includegraphics[width=1\textwidth]{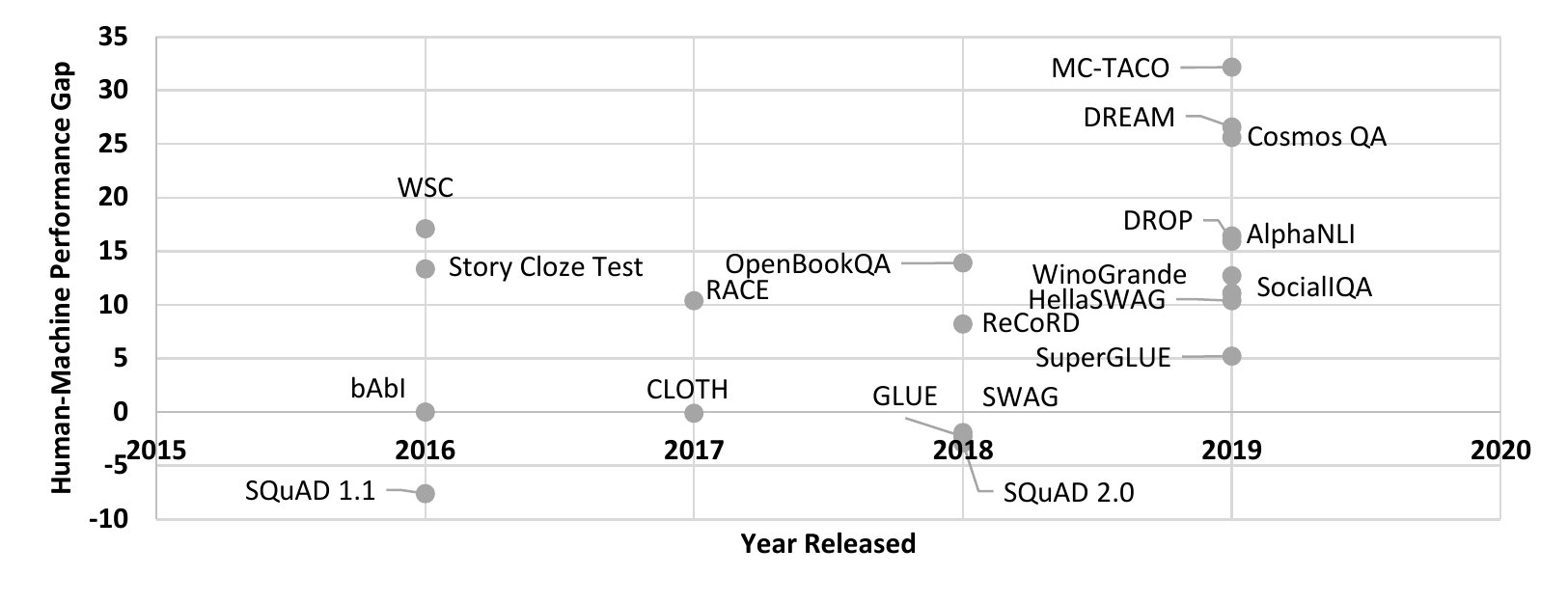}
  \caption{Year released versus gap between state-of-the-art machine accuracy and human accuracy for selected benchmarks where human performance is reported. }
  \label{fig:difficulty popularity}
\end{figure}

\vspace{5pt}
\noindent
{\em Benchmark age.}
\textcolor{black}{How long a benchmark has been released certainly affects how much the benchmark has been worked on, and therefore must affect the state-of-the-art machine performance. Figure~\ref{fig:difficulty popularity} shows the difference between the state-of-the-art and human performance with respect to the year when the benchmark was released.}
\textcolor{black}{While again we cannot find a statistically significant trend, many of the data points qualitatively support that we have made more progress on older benchmarks. This does not mean that the newer benchmarks are more difficult than those which we have made more progress on, rather there may have not yet been enough time or enough people working on them to achieve similar progress. However, some interesting anomalies can be identified in the graph, such as the Winograd Schema Challenge \cite{davisFirstWinogradSchema2017} and the Story Cloze Test \cite{mostafazadehCorpusClozeEvaluation2016}. While the former can likely be explained by very small data size, as discussed earlier, the latter has a relatively high gap from human performance despite being available for a few years and having a large magnitude of training data. This may be due to the winning entry from \citeA{schwartzEffectDifferentWriting2017} in a shared task showing that the benchmark has human writing style biases which enable artificially high performance on the benchmark \cite{mostafazadehLSDSem2017Shared2017}, as discussed in Section~\ref{sec:bias}. The last leaderboard\footnote{\url{https://competitions.codalab.org/competitions/15333\#results}} submission for this benchmark occurred over a year ago at time of writing (October 2019), so it seems researchers may have lost interest in it. Meanwhile, on other large benchmarks from the same year like bAbI \cite{westonAICompleteQuestionAnswering2016} and SQuAD 1.1 \cite{rajpurkarSQuAD1000002016}, human performance has been met or exceeded.}

\textcolor{black}{More recent benchmarks generally have a higher gap, and this may just be because they are newer. This may also be partly explained by the recent trend of adversarial construction of data and avoiding data biases, e.g., in benchmarks like AlphaNLI~\cite{ch2019abductive} and WinoGrande~\cite{sakaguchi2019winogrande}, which used adversarial methods to remove some statistical biases from the data. On the other hand, benchmarks created before this research trend are likely to be solved easily by state-of-the-art models, and thus have a lower human-machine performance gap. There are exceptions, however. For example, the Winograd Schema Challenge~\cite{davisFirstWinogradSchema2017} has a higher gap from human performance than the adversarial WinoGrande~\cite{sakaguchi2019winogrande}. Again, this could be because the former benchmark does not have much data, and also because it was constructed by several experts, perhaps making it much harder for neural models to find and exploit biases in writing style.}

\begin{figure}
  \centering
  \includegraphics[width=1\textwidth]{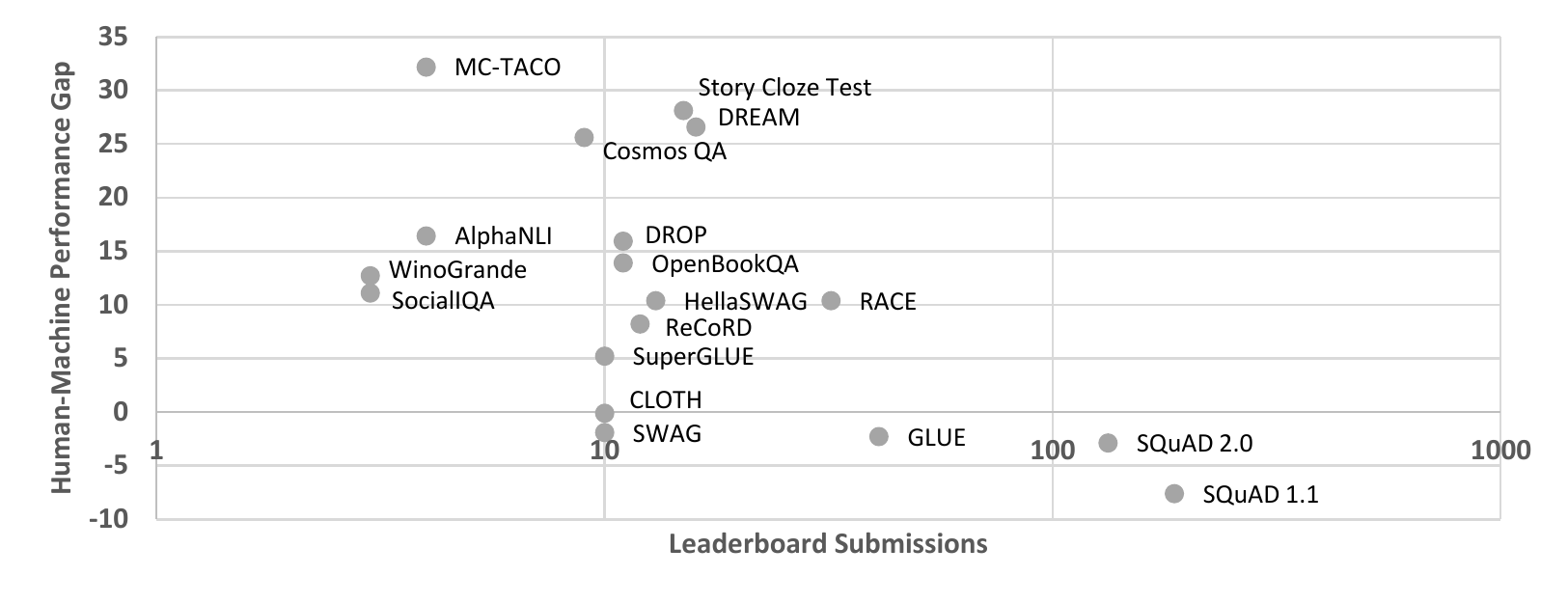}
  \caption{Number of leaderboard submissions versus gap between state-of-the-art machine accuracy and human accuracy for selected benchmarks where human performance is reported and public leaderboards are maintained. }
  \label{fig:difficulty popularity 2}
\end{figure}

\vspace{5pt}
\noindent
{\em Benchmark popularity.}
\textcolor{black}{The popularity of a benchmark can also affect how many researchers are working on it, and consequently how much progress can be made on it and thus how difficult it may be perceived to be.} 
\textcolor{black}{
If a benchmark is more popular, we may expect it to get more submissions on a leaderboard, so the gap between state-of-the-art and human performance will close rapidly. On the other hand, if a benchmark is less popular, it will receive fewer submissions, and the state-of-the-art performance will not change much, perhaps causing the benchmark to be perceived as more difficult. For selected benchmarks with online leaderboards, Figure~\ref{fig:difficulty popularity 2} compares the number of unique, public submissions to the human-machine performance gap. }

\textcolor{black}{For benchmarks with a low number of submissions, there is high variability, suggesting no relationship between popularity and progress on the benchmark. However, for the few benchmarks with a much higher number of submissions, we see an increasingly negative human-machine performance gap, indicating that human performance has been exceeded. The most popular benchmarks, i.e., SQuAD 1.1 \cite{rajpurkarSQuAD1000002016} and 2.0 \cite{rajpurkarKnowWhatYou2018}, have seen the most progress, with machine performance far surpassing human performance. }

\textcolor{black}{In conclusion, when it comes to the complexity of benchmark tasks, there is not a straightforward way to analyze this. While human and machine performance on a benchmark can shed some light on task complexity, we must consider various limitations of these performance measures themselves, as well as other factors that may hinder or drive progress in machine performance, such as benchmark data size, age, and popularity. All of these issues can affect how we perceive the complexity of benchmarks, so results must be interpreted accordingly.}

%% file: 3-resources.tex
\section{Knowledge Resources}\label{sec:resources}

It is estimated that typical human adults possess up to 100,000,000 different axioms of commonsense knowledge \cite{chklovskiLearnerSystemAcquiring2003}. The lack of this knowledge in NLI applications is currently a major bottleneck. \textcolor{black}{In order to remove this bottleneck, decades of efforts in AI have been spent on developing knowledge representations and various knowledge resources, from early works on semantic networks \cite{woodsWhatLinkFoundations1975,quillianSemanticMemory1966} to the recent efforts of compiling large-scale commonsense knowledge graphs \cite{speerConceptNetOpenMultilingual2017,sapATOMICAtlasMachine2019}.} The acquired knowledge is often represented in various forms such as propositions, taxonomies, ontologies, and semantic networks.
In this section, we start with an introduction to several existing knowledge resources, then discuss the main issues involved in building and leveraging these resources.

\subsection{An Overview of Existing Knowledge Resources}
As outlined in Section~\ref{sec:knowledgereasoning}, to understand human language, it is important to have {\em linguistic knowledge} resources that allow computers to identify syntactic and semantic structures from language. These structures often need to be augmented with {\em common knowledge} and {\em commonsense knowledge} to reach a full understanding.

\subsubsection{Linguistic Knowledge Resources}\label{representation linguistic}
Linguistic resources have been pivotal in pushing the NLP field forward in the last thirty years. Resources have been developed where annotations for syntactic, semantic, and discourse structures are provided for training machine learning models. Several knowledge bases, particularly for lexical semantics, have also been made available to facilitate semantic processing. Also toward semantic processing, various vector embeddings of word meanings have been created. In this section, we summarize these annotated corpora, frame semantics resources, lexical resources, and pre-trained semantic vectors for linguistic knowledge.

\paragraph{Annotated linguistic corpora.}
Widely used linguistic resources include the Penn Treebank \cite{marcusBuildingLargeAnnotated1993} and several derivatives of it. The Penn Treebank is perhaps the first annotated corpus that drove the development of earlier machine learning approaches in the 1990s. It started with POS tags and syntactic structures based on context-free grammar. 
The Penn Discourse Treebank (PDTB) is built upon it, adding annotated discourse structures \cite{miltsakakiPennDiscourseTreebank2004}. OntoNotes revises information in the Wall Street Journal portion of these treebanks, integrating them with word sense, proper name, coreference, and ontological annotations, as well as including some Chinese linguistic annotations \cite{pradhanOntoNotesUnifiedRelational2007}. The Abstract Meaning Representation (AMR) corpus extends the Penn Treebank into a sentence-level semantic formalism \cite{banarescuAbstractMeaningRepresentation2013}. All of these linguistic corpora can be downloaded by members of the Linguistic Data Consortium at \url{https://www.ldc.upenn.edu/}.

\paragraph{Lexical resources.}
A widely used lexical resource for commonly used nouns, verbs, adjectives and adverbs is WordNet~\footnote{\url{https://wordnet.princeton.edu/}}  \cite{millerWordNetLexicalDatabase1995}. Different from a traditional online dictionary, WordNet organizes words in terms of concepts (i.e., a list of synonyms) and their semantic relations to other words (e.g., antonymy, hyponymy/hypernymy, entailment, etc.).
There are also resources specifically for verbs. VerbNet~\footnote{\url{https://verbs.colorado.edu/~mpalmer/projects/verbnet.html}} by \citeA{schuler2005verbnet} is a hierarchical English verb lexicon that is created based on the verb classes from the English Verb Classes and Alternations (EVCA) resource by \citeA{levinEnglishVerbClasses1993}. VerbNet defines many classes of verbs and their argument structures, selectional restrictions on the arguments, and syntactic descriptions. Other resources for verbs include VerbOcean~\footnote{\url{https://demo.patrickpantel.com/demos/verbocean/}}\cite{chklovskiVerbOceanMiningWeb2004} which captures a network of finer-grained relations among a smaller set of common verbs, and VerbCorner~\footnote{\url{https://archive.gameswithwords.org/VerbCorner/about.php}}~\cite{hartshorne2013verbcorner} which provides crowd-sourced validation for VerbNet.

\paragraph{Frame semantics.}
\textcolor{black}{Beyond annotations over large corpora or lexical hierarchies, verb semantics may also be captured by frames. Frames were defined by \citeA{minskyFrameworkRepresentingKnowledge1974} as cognitive data structures for information about prototypical events and situations. After further development in works by \citeA{fillmoreNeedFrameSemantics1976} and \citeA{fillmoreScenesandFramesSemantics1977}, the FrameNet project began annotating large texts to be used for such semantic generalization \cite{bakerBerkeleyFrameNetProject1998}. This resulted in the FrameNet resource~\footnote{\url{https://framenet.icsi.berkeley.edu/fndrupal/framenet_data}}~\cite{fillmoreFrameNetDatabaseSoftware2002}, which provides a database of semantic frames describing situations, information about them (e.g., the expected participants), relations between them, and sentences annotated for elements of the frames. It additionally includes a lexical database of nouns, verbs, and adjectives which are paired with the frames. This important resource inspired the influential and well-studied task of semantic role labeling \cite{gildeaAutomaticLabelingSemantic2000,punyakanokSemanticRoleLabeling2004}, where models are trained to annotate sentences for types of participants in events.
PropBank~\footnote{\url{https://propbank.github.io/}} by \citeA{kingsburyAddingSemanticAnnotation2002} uses frames to provide the annotation of similar predicate-argument structures over the Wall Street Journal portion of the Penn Treebank \cite{taylorPennTreebankOverview2003}. 
Other structures for common events have been proposed, such as plans \cite{millerPlansStructureBehavior1960} and scripts \cite{schankScriptsPlansGoals1988}, which have inspired popular language tasks like the narrative cloze task \cite{chambersUnsupervisedLearningNarrative2008}. }

\paragraph{Pre-trained semantic vectors.}
\textcolor{black}{
Recent years have seen a surge of using continuous numeric vector representations to capture semantics of words. 
A detailed review of such resources is provided by \citeA{camacho-colladosWordSenseEmbeddings2018}. } 
\textcolor{black}{Early approaches evolved from first using counts of co-occurring n-grams for statistical language modeling \cite{brownClassBasedNgramModels1992}, then using these counts to generate spare semantic vectors \cite{lundProducingHighdimensionalSemantic1996}.}
\textcolor{black}{The most recent works, e.g., word2vec \cite{mikolovEfficientEstimationWord2013}, train neural networks on word co-occurrence classification tasks generated from large texts. The learned weights from these neural networks are then used as dense vector representations of words, which can later be used as a resource to represent words in the input to neural networks for other tasks. More word embedding models have been proposed since word2vec, and the pre-trained semantic vectors from them are often made available. GloVe\footnote{\url{https://nlp.stanford.edu/projects/glove/}} \cite{penningtonGloveGlobalVectors2014} is a highly popular example, which improves efficiency and performance by training on only nonzero co-occurrence measures. Newer works have leveraged sub-word information in order to better handle rare words for which limited training data is available \cite{bojanowskiEnrichingWordVectors2017}. This can be done by splitting training text into characters when training semantic vectors, as in FastText\footnote{\url{https://github.com/facebookresearch/fastText}} by \citeA{bojanowskiEnrichingWordVectors2017}, an extension of the word2vec models. This has also been explored by training semantic vectors for flexibly-sized pieces of words, as in the wordpiece embeddings\footnote{See \url{https://github.com/google-research/bert} for pre-trained model, or the similar SentencePiece project at \url{https://github.com/google/sentencepiece} to train a new model.} by \citeA{wuGoogleNeuralMachine2016}.}

\textcolor{black}{There are many advantages to using these pre-trained semantic vectors in language processing tasks. As similar words co-occur in text and they are trained from co-occurrence, vectors for similar words appear close together in the semantic vector space. Further, they can be shown to capture word meaning to some degree. For example, in the word2vec model, it was found that when subtracting the vector for \textit{man} from the vector for \textit{king}, then adding the vector for \textit{woman}, a vector approximately equal to the pre-trained vector for \textit{queen} is produced \cite{mikolovEfficientEstimationWord2013}. A major limitation of these vectors, however, is that they are context-independent, so the vector representation of a word is the same regardless of context and word sense. Most recently, contextual word representations, where representations for a word depend on where the word appears in a context, have been heavily used. More information on such representations is given in Section~\ref{sec:transfer learning}.}

\subsubsection{Common Knowledge Resources}\label{representation common}
Common knowledge refers to well-known facts about the world that are often explicitly stated \cite{cambriaIsanetteCommonCommon2011a}. Since these facts are so often stated, they can be mined from the Web with relative ease to create knowledge bases. In this section, we summarize several such resources for common knowledge.

\paragraph{YAGO.}
Wikipedia is a large and open source of common knowledge. Yet Another Great Ontology (YAGO) by \citeA{suchanekYAGOCoreSemantic2007} augments WordNet \cite{millerWordNetLexicalDatabase1995} with common knowledge facts extracted from Wikipedia, converting WordNet from a primarily linguistic resource to a common knowledge base. YAGO originally consisted of more than 1 million entities and 5 million facts describing relationships between these entities. YAGO2 grounded entities, facts, and events in time and space, contained 446 million facts about 9.8 million entities \cite{hoffartYAGO2SpatiallyTemporally2012}, while YAGO3 added about 1 million more entities from non-English Wikipedia articles \cite{mahdisoltaniYAGO3KnowledgeBase2013}. YAGO is available for free download at \url{https://www.mpi-inf.mpg.de/departments/databases-and-information-systems/research/yago-naga/yago/downloads/}.

\paragraph{DBpedia.}
DBpedia by \citeA{auerDBpediaNucleusWeb2007} is another Wikipedia-based knowledge base originally consisting of structured knowledge from more than 1.95 million Wikipedia articles. At its creation, DBpedia included around 103 million Resource Description Framework (RDF) triples\footnote{\url{https://www.w3.org/TR/rdf-concepts/\#section-triples}}, which are triples of subjects, predicates, and objects which describe semantic relationships. These triples included descriptions of concepts within articles, information about people, links between articles, and category labels from YAGO. The latest version, available for free at \url{https://wiki.dbpedia.org/develop/datasets}, consists of 6.6 million entities, 5.5 million resources classified in the DBpedia ontology, and over 23 billion RDF triples. 

\paragraph{WikiTaxonomy.}
Yet another Wikipedia-based resource is WikiTaxonomy by \citeA{ponzettoDerivingLargeScale2007}, which consists of about 105,000 well-evaluated semantic links between categories in Wikipedia articles. Categories and relationships are labeled using the connectivity of the conceptual network formed by the categories. The authors demonstrate that this resource can be used to calculate semantic similarity of words, which may be useful in textual entailment and other NLI tasks. WikiTaxonomy is available for free download at \url{https://www.h-its.org/en/research/nlp/wikitaxonomy/}.

\paragraph{Freebase.}
Freebase by \citeA{bollackerFreebaseCollaborativelyCreated2008} was a knowledge graph which originally contained 125 million RDF triples of general human knowledge about 4,000 types of entities and 7,000 properties of entities. The data of Freebase was absorbed into the Google Knowledge Graph for intelligent web searching, and even migrated to Wikidata (introduced later), however the last release is still available for free download at \url{https://developers.google.com/freebase/}. It contains more than 1.9 billion triples.

\paragraph{NELL.}
As common knowledge is not static, it is advantageous for knowledge bases to grow over time. This is typically done through new releases, however the Never-Ending Language Learner (NELL) by \citeA{carlsonArchitectureNeverEndingLanguage2010} continually grows by automatically mining structured beliefs of varying confidence from the web daily. It originally contained 242,000 beliefs about properties of entities, but now contains over 50 million beliefs, with almost 3 million of these having high confidence. This version can be downloaded for free at \url{https://rtw.ml.cmu.edu/rtw/}.

\paragraph{Probase.}
Probase by \citeA{wuProbabilisticTaxonomyMany2011} is different from previous common knowledge taxonomies in that relationships are probabilistic rather than concrete. Probase consists of 2.7 million concepts extracted from 1.6 billion web pages. Relationships between concepts are described in 20.8 million is-a and is-instance-of pairs, and probabilistic interpretation is possible through provided similarity values between 0 and 1 for each pair of concepts in the knowledge base. Though the original resource is no longer available, the Microsoft Concept Graph which was built upon Probase can be downloaded for free at \url{https://concept.research.microsoft.com/Home/Download}.

\paragraph{Wikidata.}
\textcolor{black}{Wikidata\footnote{No paper has been published for Wikidata, however all information can be found at the homepage linked to at the end of this paragraph.} is an open, collaborative, multilingual knowledge graph launched in 2012 which can be edited by anyone. At time of writing, it consisted of nearly 59 million items, which are typically commonly known entities such as famous people, artistic works, and buildings. Each item consists of various property relations linking them to other items, such as the example given on their homepage: the \textit{highest point} property for the item \textit{Earth} has the value \textit{Mount Everest}. Information for using Wikidata can be found at \url{https://www.wikidata.org/}.}

\subsubsection{Commonsense Knowledge Resources}\label{representation commonsense}
Commonsense knowledge, on the other hand, is considered obvious to most humans, and not likely to be explicitly stated \cite{cambriaIsanetteCommonCommon2011a}. There has been a long effort in capturing and encoding commonsense knowledge \cite{davisCommonsenseReasoningCommonsense2015}, and various knowledge bases have been developed for this. Note that as commonsense knowledge and common knowledge are not always distinguished from each other in past work, and because some of these resources were gathered from the Web, a source of primarily common knowledge, many knowledge bases we describe here also contain common knowledge. We present several resources which focus on commonsense knowledge, but may also include common knowledge.

\paragraph{Cyc.}
A well-known project toward encoding commonsense knowledge is Cyc by \citeA{lenatBuildingLargeKnowledgeBased1989}, a knowledge base of rules expressing ontological relationships between objects encoded in the formal logic-based CycL language. The types of objects in Cyc include entities, collections, functions, and truth functions. Cyc also includes a powerful inference engine. ResearchCyc, a release of Cyc for the research community, can be licensed for free at \url{https://www.cyc.com/researchcyc/}. According to this site, the latest release of ResearchCyc contains over 7 million commonsense knowledge assertions. More recently, there have been efforts to map Cyc to Wikipedia articles in an attempt to connect it to other resources such as DBpedia and Freebase \cite{medelyanIntegratingCycWikipedia2008,pohlClassifyingWikipediaArticles2012}.

\paragraph{ConceptNet}
Another popular knowledge base is ConceptNet from \citeA{liuConceptNetPracticalCommonsense2004}, a product of the Open Mind Common Sense project by \citeA{singhPublicAcquisitionCommonsense2002}, which collected free text common and commonsense knowledge from online users. This semantic network originally contained over 1.6 million assertions of commonsense knowledge represented as links between 300,000 nodes representing entities, but subsequent releases have expanded it and added more features. The latest release, ConceptNet 5.5 \cite{speerConceptNetOpenMultilingual2017}, contains over 21 million links between over 8 million nodes, having been augmented by several additional resources including Cyc \cite{lenatBuildingLargeKnowledgeBased1989}, WordNet \cite{fellbaumWordNetElectronicLexical1999}, and DBpedia \cite{auerDBpediaNucleusWeb2007}. It includes knowledge from multilingual resources, and links to knowledge from other knowledge graphs. ConceptNet has been successfully applied in NLI systems, some of which are described in Section~\ref{sec:knowledge base approaches}. ConceptNet is open; information for using or downloading it can be found at \url{https://conceptnet.io/}.

\paragraph{SenticNet.}
SenticNet by \citeA{cambriaSenticNetPubliclyAvailable2010} was originally only a commonsense knowledge base, but later versions incorporated common knowledge as well \cite{cambriaSenticNetCommonCommonSense2014}. Though the knowledge base is intended for sentiment analysis, it may be useful in NLI tasks which require understanding sentiment. SenticNet is available for free at \url{https://sentic.net/downloads/}.

\paragraph{Isanette and IsaCore.}
Isanette by \citeA{cambriaIsanetteCommonCommon2011a} was a semantic network of both common and commonsense knowledge created by combining ProBase \cite{wuProbabilisticTaxonomyMany2011} and ConceptNet 3 \cite{havasiConceptNetFlexibleMultilingual2007} into a set of "is a" relationships and confidences. This work was later cleaned and optimized into IsaCore \cite{cambriaSemanticMultidimensionalScaling2014}, and demonstrated to be effective for sentiment analysis. IsaCore may also be a useful resource for other tasks. It is available for free download at \url{https://sentic.net/downloads/}.

\paragraph{COGBASE.}
COGBASE by \citeA{olsherSemanticallybasedPriorsNuanced2014a} uses a novel formalism to represent 2.7 million concepts and 10 million commonsense facts about them. It makes up the core of SenticNet 3 \cite{cambriaSenticNetCommonCommonSense2014}. Data from COGBASE can currently be accessed via an online interface and an API available at \url{https://cogview.com/cogbase/}. 

\paragraph{WebChild.}
WebChild by \citeA{tandonWebChildHarvestingOrganizing2014} was originally a commonsense knowledge base of general noun-adjective relations extracted from Web content and other resources, consisting of about 78,000 distinct noun senses, 5,600 distinct adjective senses, and 4.6 million assertions between them. These assertions captured fine-grained relations among the noun and adjective senses. Unlike other resources collected from the Web, WebChild consists primarily of commonsense knowledge, as it consists of generalized, fine-grained relationships between nouns and adjectives collected from various corpora rather than structured common knowledge taken directly from a single open source like Wikipedia. WebChild 2.0 \cite{tandonWebChildFineGrainedCommonsense2017} was later released to include over 2 million concepts and activities, and over 18 million such assertions. Data from WebChild can be browsed and downloaded online at \url{https://www.mpi-inf.mpg.de/departments/databases-and-information-systems/research/yago-naga/webchild/}.

\paragraph{LocatedNear.}
\citeA{xuAutomaticExtractionCommonsense2018} claim that objects which tend to be near each other (e.g., silverware, a plate, and a glass) is a type of commonsense knowledge lacking in previous knowledge bases like ConceptNet 5.5 \cite{speerConceptNetOpenMultilingual2017}. They refer to this property as LocatedNear, and to address this issue, they create two datasets which we refer to jointly as LocatedNear. The first consists of 5,000 sentences describing scenes of two objects labeled for whether the objects tend to occur near each other, which can serve as a benchmark task similar to those introduced in Section~\ref{sec:benchmarks}. The second consists of 500 pairs of objects with human-produced confidence scores for how likely the objects are to appear near each other. These resources can be downloaded from \url{https://github.com/adapt-sjtu/commonsense-locatednear}.

\paragraph{ATOMIC.}
The Atlas of Machine Commonsense (ATOMIC) by \citeA{sapATOMICAtlasMachine2019} is a crowdsourced knowledge graph consisting of about 300,000 nodes corresponding to short textual descriptions of events, and about 877,000 "if-event-then" triples representing nine types of if-then relations between everyday events. Rather than taxonomic or ontological knowledge, this graph contains easily-accessed inferential knowledge. \citeauthor{sapATOMICAtlasMachine2019} demonstrate that neural models can learn simple reasoning skills from ATOMIC which can be used to make inferences about previously unseen events. ATOMIC can be browsed and downloaded for free at \url{https://homes.cs.washington.edu/~msap/atomic/}.

\paragraph{ASER.}
Similar to ATOMIC, the Activities, States, Events, and Relations knowledge graph by \citeA{zhangASERLargescaleEventuality2019} consists of everyday events and relations between them which can be used for inference. It is on a much larger scale, though, containing 194 million events, 64 million relations, and fifteen types of relations. Further, these relations are automatically extracted from 11 billion tokens of unstructured text data rather than crowdsourced. More information about ASER and a free download link can be found at \url{https://hkust-knowcomp.github.io/ASER/}.

\subsection{Approaches to Creating Knowledge Resources}

Similar to creating the benchmarks described in Section~\ref{sec:benchmarks}, various approaches have been applied to create knowledge resources. These approaches range from manual encoding by experts, to text mining from web documents, and collection through crowdsourcing. A detailed description of these approaches is provided by~\citeA{davisCommonsenseReasoningCommonsense2015}. Here, we give a brief discussion about pros and cons of these approaches.

\paragraph{Manual encoding.}
Early knowledge bases were often manually created. The classic example of this is Cyc, which is produced by knowledge engineers who hand-code commonsense knowledge into the CycL formalism \cite{lenatBuildingLargeKnowledgeBased1989}. Cyc has been going through continuous development over the last 35 years, since its first release in 1984. The cost of this manual encoding is high, with a total estimated cost of \$120M~\cite{paulheimHowMuchTriple2018}. As a consequence,  
Cyc is small relative to other resources, and growing very slowly. On the other hand, this expert-based approach ensures high quality of data.

\paragraph{Text mining.}
Text mining and information extraction tools are often applied to automatically generate knowledge graphs and taxonomies from information sources on the Web. One popular information source is Wikipedia, which was drawn from in creating common knowledge bases such as YAGO \cite{suchanekYAGOCoreSemantic2007}, DBpedia \cite{auerDBpediaNucleusWeb2007}, WikiTaxonomy \cite{ponzettoDerivingLargeScale2007}. Other knowledge bases are generated from crawling the Web, e.g., NELL \cite{carlsonArchitectureNeverEndingLanguage2010}, from benchmark datasets, e.g., ATOMIC \cite{sapATOMICAtlasMachine2019}, or even from other knowledge bases, e.g., IsaCore \cite{cambriaSemanticMultidimensionalScaling2014}. \textcolor{black}{KnowItAll \cite{etzioniUnsupervisedNamedentityExtraction2005} and TextRunner \cite{etzioniOpenInformationExtraction2008a} have been popular approaches for extracting this common knowledge from online sources.}

\textcolor{black}{However, as commonsense knowledge is rarely explicitly stated, it is more difficult to extract from text sources. Additional processes are often required to obtain knowledge beyond the text. For example, processes have been used to extract unstated inference and textual entailment rules, e.g., from Web text \cite{schoenmackersLearningFirstorderHorn2010,gordonDiscoveringCommonsenseEntailment2011}, and knowledge graphs \cite{berantGlobalLearningTyped2011}. Such rules could prove useful for many of the benchmarks introduced in Section~\ref{sec:benchmarks}. Several processes have been proposed for extracting temporal commonsense knowledge of events from text, particularly useful for benchmarks like MC-TACO \cite{zhou2019going}, which requires this knowledge. The expected durations of events can be gathered through the processes proposed by \citeA{kozarevaLearningTemporalInformation2011} and 
\citeA{samardzicAspectbasedLearningEvent2016}, and expected frequencies of events can be learned through work by \citeA{gordonUsingTextualPatterns2012}. Information about the order of events and their participants can be extracted through processes proposed by \citeA{chambersUnsupervisedLearningNarrative2008} and \citeA{wangIntegratingOrderInformation2017}, which may be useful in gathering commonsense knowledge about which events typically follow each other. Commonsense physical and spatial relations between objects can be inferred through processes proposed by \citeA{forbesVerbPhysicsRelative2017}, \citeA{collellAcquiringCommonSense2018}, and \citeA{yangExtractingCommonsenseProperties2018}. Relations and trends between numbers and language can also be identified and predicted through processes by  \citeA{chagantyHowMuch1312016} and \citeA{spithourakisNumeracyLanguageModels2018}. General comparative commonsense relations between entities, e.g., the knowledge that bears tend to be more dangerous than dogs, can be gathered through a process by \citeA{tandonAcquiringComparativeCommonsense2014}.}

One key advantage of these text mining approaches is cost efficiency. According to~\citeA{paulheimHowMuchTriple2018}, creating a statement in the Wikipedia-extracted DBpedia and YAGO respectively costs 1.85 cents and 0.83 cents (USD), which are hundreds of folds less than the cost of manually encoding a statement in Cyc (which was estimated at about \$5.71 per statement). This makes text mining approaches easily scale up to create large knowledge bases. However, the drawback is that the acquired knowledge can be noisy and inconsistent if extracted from open online data. There exist some automatic error correction processes, e.g., a process by \citeA{spithourakisNumericallyGroundedLanguage2016} which uses language models to identify and correct textual misinterpretations of numbers. However, human validation is still likely to be required when automatically generating knowledge resources in this way \cite{gordonEvaluationCommonsenseKnowledge2010}.

\paragraph{Crowdsourcing.}
Another highly popular approach to creating knowledge bases is crowdsourcing. 
The Open Mind Common Sense project responsible for producing ConceptNet \cite{liuConceptNetPracticalCommonsense2004} used a competitive online game to accept statements from humans in free text \cite{singhPublicAcquisitionCommonsense2002}. 
Later, researchers converted the knowledge within collected statements into a knowledge graph by automatic processes. \textcolor{black}{This method of using games to attract users to perform human intelligence tasks for free is typically referred to as Games with a Purpose (GWAP). Other GWAPs include the one used to collect annotations for VerbCorner \cite{hartshorne2013verbcorner}, and the Robot Trainer knowledge acquisition game by \citeA{rodosthenousHybridApproachCommonsense2016}, where players must teach human knowledge to a virtual robot.}
The cost of crowdsourcing effort is difficult to assess. It ranges from literally getting it for free (e.g., GWAPs like Open Mind Common Sense) to an estimate of \$2.25 per statement~\cite{paulheimHowMuchTriple2018} for paid crowdsourcing efforts. Though the gaming approach may be cheaper in the long run, developing such a game platform is inevitably more time-consuming. Another challenge of crowdsourcing, as pointed out by \citeA{davisCommonsenseReasoningCommonsense2015}, is that typical crowd workers may not be able to follow the theories and representations of knowledge that engineers have worked out. As a result, knowledge acquired by crowdsourcing can be somewhat messy, which again often needs human expert validation.

\vspace{10pt}

Each of these methods has its own advantages and drawbacks in terms of the trade-offs between the cost and the quality of the acquired knowledge. Most of these knowledge resources are developed from a bottom-up fashion. The goal is to create general knowledge bases to provide inductive bias for a variety of learning and reasoning tasks. Nevertheless, it is not clear whether such a goal is met and to what extent these knowledge resources are applied to natural language inference in practice. A systematic study, as suggested by \citeA{davisCommonsenseReasoningCommonsense2015}, for Cyc and other resources would be useful.

\subsection{Leveraging Incomplete Resources}\label{sec:kb completion}
\textcolor{black}{Due to the high magnitude of knowledge possessed by humans, no single knowledge resource is complete, and even the largest resources are missing essential knowledge to perform well on NLI benchmarks. This was discovered, for example, of ConceptNet \cite{liuConceptNetPracticalCommonsense2004}, which is missing many required relations for the intuitive psychology-based Event2Mind benchmark \cite{rashkinEvent2MindCommonsenseInference2018}. To address this problem, several methods have been proposed for leveraging incomplete knowledge bases, from dimensionality reduction techniques to extractive and automatic knowledge base completion techniques.}

\paragraph{Dimensionality reduction.}
\textcolor{black}{One method is through dimensionality reduction. An example of this is AnalogySpace \cite{speerAnalogySpaceReducingDimensionality2008}, an algorithm packaged with the latest release of ConceptNet \cite{speerConceptNetOpenMultilingual2017}. AnalogySpace uses principle component analysis to make analogies along dimensions such as goodness and difficulty, and smooth missing and noisy relations. \citeA{kuoBridgingCommonSense2010} use a similar approach to compute the similarity between knowledge graph structures and draw analogies across them, producing new relations of which up to 77.6\% are judged as accurate by crowd workers. The drawback of such a method is that it cannot create new terms, rather it can only generate confidence scores between any existing terms in a knowledge graph \cite{li-etal-2016-commonsense}.}

\paragraph{Extractive knowledge base completion.}
\textcolor{black}{More recently, work has aimed to complete these knowledge bases by generating new relations. \citeA{li-etal-2016-commonsense} train a neural network to distinguish relations in ConceptNet from false relations, which can then be used to score the truth of proposed new relations. Further, they use underlying crowdsourced text from Open Mind Common Sense \citeA{singhPublicAcquisitionCommonsense2002} to train a word embedding which maps natural language queries to rigidly-worded relations in ConceptNet. \citeA{jastrzebski-etal-2018-commonsense} use such a mapping to mine novel commonsense knowledge relations from text, while \citeA{saito-etal-2018-commonsense} train a model to jointly generate and complete knowledge bases from language queries to improve accuracy of proposed relations.}

\paragraph{Automatic knowledge base completion.}
\textcolor{black}{Rather than extracting new relations from text, new relations may be inferred from existing relations. \citeA{angeliNaturalLINaturalLogic2014} use a relaxed logical framework to infer novel relations from existing relations in a large knowledge base, and demonstrate that the new relations can be successfully used for inference in a textual entailment task. More recently, neural networks are used to generate new relations. One example is memory comparison networks~\cite{andradeLeveragingKnowledgeBases2018}, which generalize over existing temporal relations in knowledge graphs, i.e., the typical ordering of events, in order to acquire new temporal relations. A similar recent work by \citeA{bosselutCOMETCommonsenseTransformers2019} trains a transformer on existing relations in order to infer novel relations in knowledge graphs. When this model was trained on ConceptNet \cite{speerConceptNetOpenMultilingual2017}, human workers rated up to 92.1\% of generated relations as accurate, demonstrating the potential effectiveness of this technique for already large but incomplete knowledge bases. While current efforts are promising, future work will need to continue to find and improve solutions to handle the long-tail phenomenon observed in these knowledge resources~\cite{davisCommonsenseReasoningCommonsense2015}.}

%% file: 4-approaches.tex
\section{Learning and Inference Approaches}\label{sec:approaches}
To solve the benchmark tasks described in Section~\ref{sec:benchmarks}, a variety of approaches have been developed. These range from earlier symbolic and statistical approaches to recent approaches that apply deep learning and neural networks. This section gives a brief overview of symbolic and early statistical approaches to the surveyed benchmarks, followed by a more detailed description of representative neural approaches, which are the current state of the art on all of the benchmarks.

\subsection{Symbolic Approaches}
\textcolor{black}{\textit{Symbolic approaches} use logical forms and processes to make inferences. The study of human reasoning through logic has continued through history, from Aristotle's theories of logic and deductive reasoning \cite{aristotlePriorAnalytics1989}, to the reasoning and intellect theories of philosophers like Alfarabi, Avicenna, and Averroes \cite{davidsonAlfarabiAvicennaAverroes1992}. These works were a foundation for modern mathematical logic, developed in influential works like the formal logic frameworks by \citeA{morganFormalLogicCalculus1847} and \citeA{booleInvestigationLawsThought1854}, and the symbolic logic theories by \citeA{hilbertPrinciplesMathematicalLogic1999} and \citeA{lewisSymbolicLogic1959}. Later work by \citeA{selmanAbductiveDefaultReasoning1990}, \citeA{pooleMethodologyUsingDefault1990}, and \citeA{HOBBS199369} further developed this methodology. During this evolution, \citeA{Peirce1883} proposed the process of logical abduction, i.e., the process of making a conclusion using a limited set of observations and a minimal number of assumptions, similar to plausible inference as defined by \citeA{davisCommonsenseReasoningCommonsense2015} for language problems. Bayesian networks have also been used for probabilistically reasoning with logical variables \cite{dechterReasoningProbabilisticDeterministic2013}.}

\textcolor{black}{Meanwhile, theories of logic found applications in AI, e.g., in the early work on commonsense for machines by \citeA{mccarthyProgramsCommonSense1968}, and linguistics, e.g., in the theory of natural logic toward a semantic representation of language by \citeA{lakoffLinguisticsNaturalLogic1970}. Fuzzy logic, which maps linguistic descriptions of numeric variables to probability distributions over the numeric variables, was an important development to handle inexactness in human language. For example, the word \textit{young} can be used to describe someone's age, and we may specify or calculate a likelihood of this word being used to describe someone over a range of ages \cite{zadehConceptLinguisticVariable1975}. \citeA{mooreRoleLogicKnowledge1982} and \citeA{nunbergPositionPaperCommonsense1987} further motivated the role of logic in commonsense knowledge and reasoning in AI and semantic processing.}
\textcolor{black}{By the early 1990s, symbolic approaches were dominant for knowledge representation and semantic processing of language \cite{birnbaumRigorMortisResponse1991}. Consequently, these approaches were heavily used for early NLI problems. A detailed review of these approaches is provided by \citeA{davisLogicalFormalizationsCommonsense2017}.}

\textcolor{black}{Among the surveyed benchmarks, symbolic approaches have primarily been applied in the early RTE Challenges. One example by \citeA{rainaRobustTextualInference2005} parsed sentences into a logical form, then performed abduction over them using learned assumptions and likelihoods in order to determine if a high-likelihood set of assumptions can be used to prove that a sentence entails another. \citeA{giampiccoloFourthPASCALRecognizing2008} used outside semantic knowledge from resources like Wikipedia, WordNet \cite{millerWordNetLexicalDatabase1995}, and VerbOcean \cite{chklovskiVerbOceanMiningWeb2004} to augment information in the hypothesis sentence, then attempted to map this to words in the premise sentence using manually authored logic rules. A later approach developed natural logic \cite{lakoffLinguisticsNaturalLogic1970} into a formalism for NLI, parsing the premise and hypothesis into a natural logic form and using a decision tree to compare their features and make a decision \cite{maccartneyNaturalLogicTextual2007}. They show that combining this approach with an existing statistical RTE model exceeds the state-of-the-art performance on the third RTE Challenge dataset \cite{giampiccoloThirdPASCALRecognizing2007}. More recently, the baseline approach to the Triangle-COPA benchmark achieved 91\% accuracy by creating a comprehensive set of manually authored logic and commonsense rules to use with given mappings from the natural language in the benchmark data to logical forms \cite{gordonCommonsenseInterpretationTriangle2016}. While manually authoring logical rules and mappings from language to logical forms has been demonstrated to be highly effective in some tasks, this is not scalable for recent, larger-magnitude datasets where variation in required knowledge and language and semantic phenomena is much higher.} 
\textcolor{black}{To facilitate more practical logic-based reasoning, another body of work focuses on developing approaches to automatically mapping natural language text to a logical form \cite{kamathSurveySemanticParsing2018}, coined as {\em semantic parsing}. There has been an extensive amount of work on this topic, and we refer readers to the survey by \citeA{kamathSurveySemanticParsing2018} for a more detailed review of this work.}

\subsection{Early Statistical Approaches}

\textcolor{black}{Statistical approaches dominated the NLP field from the mid 1990s until the early 2010s. These earlier statistical approaches relied on engineered features to train various types of statistical models based on training data for a variety of early NLP benchmarks, e.g., question answering benchmarks like the classic Remedia reading comprehension dataset \cite{hirschmanDeepReadReading1999} and the TREC datasets \cite{voorheesTREC8QuestionAnswering2000}, which were largely based on linguistic context and thus outside of the scope of this survey. Notable early approaches often combined various word matching and other lexical features with handcrafted deterministic rules \citeA{charniakReadingComprehensionPrograms2000}, or with traditional statistical models like decision trees \cite{ngMachineLearningApproach2000}. Similar approaches were also applied to some of the benchmarks we survey here,}
beginning with the RTE Challenges \cite{daganPASCALRecognisingTextual2005}. 
Lexical features, for example, based on bag-of-words and word matching were commonly used in the first two RTE Challenges \cite{daganPASCALRecognisingTextual2005,bar-haimSecondPASCALRecognising2006}, but often achieved results only slightly better than random guessing \cite{bar-haimSecondPASCALRecognising2006}. More competitive systems have used more linguistic features to make predictions, such as semantic dependencies and paraphrases \cite{hicklRecognizingTextualEntailment2006}, synonym, antonym, and hypernym relationships derived from training data, and hidden correlation biases in benchmark data \cite{laiIllinoisLHDenotationalDistributional2014}.

External knowledge and the Web were often used to complement features derived from the training data. For example, the best system in the first RTE Challenge \cite{daganPASCALRecognisingTextual2005} used a na{\"i}ve Bayes classifier with features from the co-occurrences of word from an online search engine \cite{glickmanAppliedTextualEntailment2006}. A similar approach was also applied in the top system from the seventh RTE Challenge \cite{bentivogliSeventhPASCALRecognizing2011}, which utilized some knowledge resources from Section~\ref{sec:resources}, acronyms extracted from the training data, and linguistic knowledge to calculate a statistical measure of entailment between sentences \cite{tsuchidaIKOMATAC2011Method2011}. While the use of some external knowledge provides an advantage over models which only use linguistic features extracted from training data, these earlier statistical models have still not been competitive in recent benchmarks of large data size. Nonetheless, such models may serve as useful baselines for new benchmarks, as demonstrated for JOCI \cite{zhangOrdinalCommonsenseInference2016}.

\subsection{Neural Approaches}\label{sec:neural approaches}

\begin{figure}
  \centering
  \includegraphics[width=0.85\textwidth]{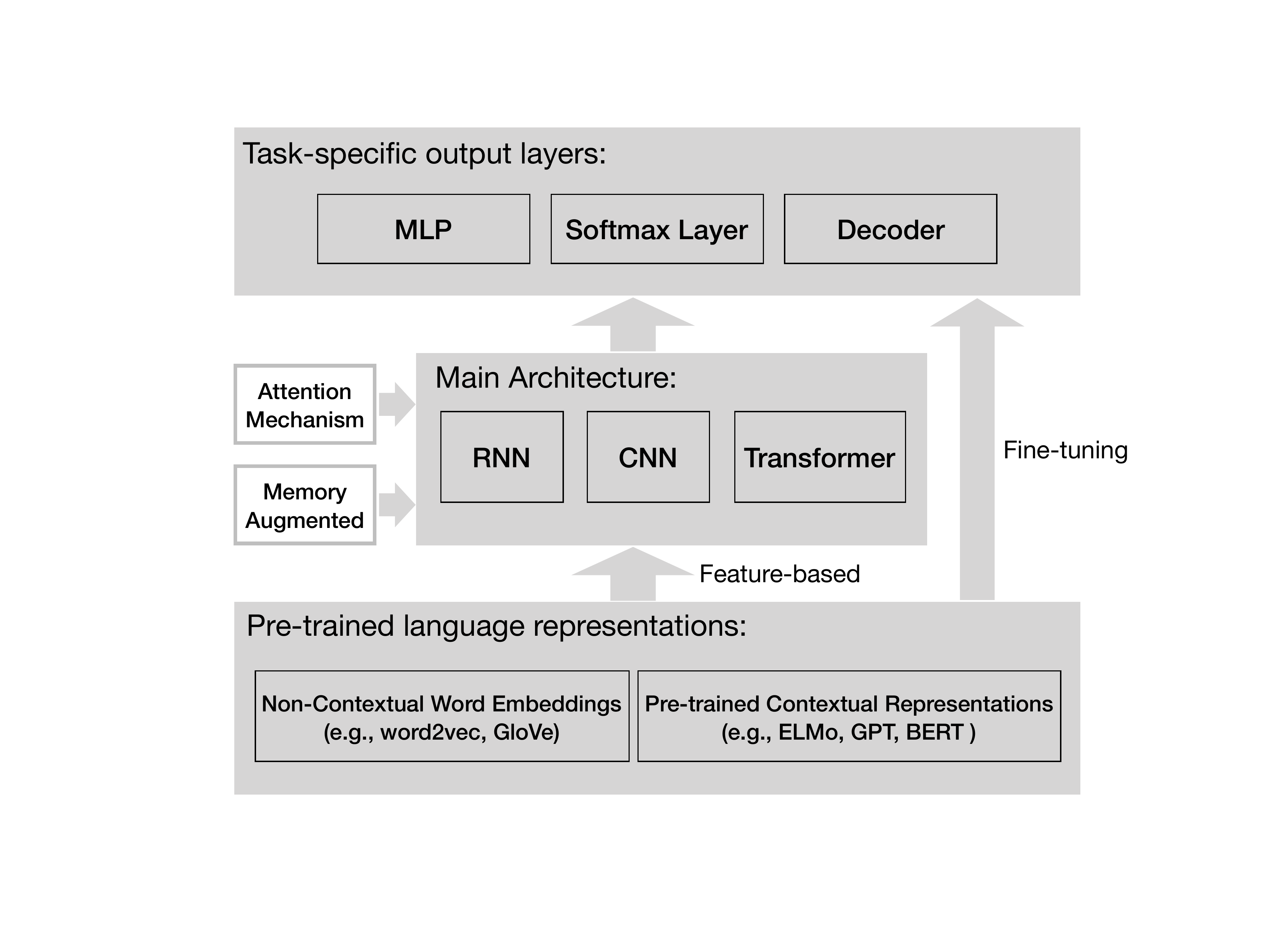}
  \caption{Common components in neural approaches to natural language inference tasks.}
  \label{fig:neural_model}
\end{figure}

\textcolor{black}{\textit{Neural approaches} evolved from earlier statistical approaches, but use various kinds of neural network architectures
to discover useful features in the data, rather than manually specifying all of the features. The increasingly large amount of data available for recent benchmarks makes it possible to train larger and deeper neural models. Today, these approaches top virtually all leaderboards for NLI benchmarks surveyed here.} Figure~\ref{fig:neural_model} shows some common components in neural models. First of all, distributional representation of words is fundamental where word vectors or embeddings are usually trained using neural networks on large-scale text corpora. 
In traditional word embedding models like word2vec \cite{mikolovEfficientEstimationWord2013} or GloVe \cite{penningtonGloveGlobalVectors2014}, the embedding vectors are context independent. No matter what context the target word appears in, once trained, its embedding vector is always the same. Consequently, these embeddings lack the capability of modeling different word senses in different contexts, although this phenomenon is prevalent in language. To address this problem, recent works have developed contextual word representation models, e.g., Embeddings from Language Models (\textsc{ELMo}) by \citeA{petersDeepContextualizedWord2018}
and Bidirectional Encoder Representations from Transformers (\textsc{BERT}) by \citeA{devlinBERTPretrainingDeep2018}. These models give words different embedding vectors based on the context in which they appear. These pre-trained word representations can be used as features or fine-tuned for downstream tasks. For example, the Generative Pre-trained Transformer (\textsc{GPT}) by \citeA{radfordImprovingLanguageUnderstanding2018a} and \textsc{BERT} \cite{devlinBERTPretrainingDeep2018} introduce minimal
task-specific parameters, and can be easily fine-tuned on the downstream tasks with modified final layers and loss functions. 

On top of the word embedding layers, task-specific network architectures are designed for different downstream applications. These architectures often adopt recurrent neural networks (RNNs) such as LSTMs \cite{Hochreiter:1997:LSM:1246443.1246450} and GRUs \cite{choPropertiesNeuralMachine2014}, convolutional neural networks (CNNs), or more recently, transformers \cite{vaswaniAttentionAllYou2017} to solve specific tasks. Output layers of these networks are chosen based on the task formulation. For example, a linear layer and softmax are often used for classification, while a language decoder will be used for language generation. Because of the sequential nature of language, RNN-based architectures are widely applied and are often implemented in both baseline approaches \cite{bowmanLargeAnnotatedCorpus2015,rashkinEvent2MindCommonsenseInference2018} and state-of-the-art approaches \cite{kimSemanticSentenceMatching2019,chenHFLRCSystemSemEval20182018,henaffTrackingWorldState2017}. Given different architectures, neural models also benefit from techniques like attention mechanisms and memory augmentation. For tasks that require reasoning based on multiple supporting facts, e.g., bAbI \cite{westonAICompleteQuestionAnswering2016}, memory-augmented networks like memory networks \cite{westonMemoryNetworks2015} and recurrent entity networks \cite{henaffTrackingWorldState2017} have been shown effective. And for tasks that require alignment between input and output, e.g., textual entailment tasks like SNLI \cite{bowmanLargeAnnotatedCorpus2015}, or capturing long-term dependencies, it is often beneficial to adopt attention mechanisms to models.

Next, we introduce several past and current state-of-the-art neural models for the surveyed benchmarks, particularly focusing on three aspects: attention mechanism, memory augmentation, and contextual models and representations.

\subsubsection{Attention Mechanism}
Since the first application of attention mechanism for neural machine translation~\cite{bahdanauNeuralMachineTranslation2015}, attention has been used widely in NLP tasks, especially to capture the alignment between an input (encoder) and an output (decoder). Modeling attention has several advantages. It allows the decoder to directly go to and focus on certain parts of the input. It alleviates the vanishing gradient problem by providing a way to account for states far away in the input sequence. Another advantage is that the attention distribution learned by the model automatically provides an alignment between inputs and outputs which allows some understanding of their relations. Because of these advantages, attention mechanisms have been successfully applied to many NLI benchmark tasks.

\paragraph{Attention in RNN/CNN.}

Adding an attention mechanism to RNNs,  LSTMs, CNNs, and more 
has been shown to improve performance on various tasks compared to their vanilla models \cite{kimSemanticSentenceMatching2019}. It is particularly successful for tasks which require alignment between input and output, such as various textual entailment tasks which require the modeling of context and hypothesis, and reading comprehension questions which refer directly to an accompanying passage, like MCScript \cite{ostermannMCScriptNovelDataset2018}.

For example, the official leaderboard\footnote{\url{https://nlp.stanford.edu/projects/snli/}} for the SNLI task \cite{bowmanLargeAnnotatedCorpus2015} reports that one of the best performing systems \cite{kimSemanticSentenceMatching2019},
partly inspired by \textsc{DenseNet} \cite{huangDenselyConnectedConvolutional2016}, uses a densely connected RNN while concatenating features from an attention mechanism to recurrent features in the network. As discussed by \citeA{kimSemanticSentenceMatching2019}, the attentive weights resulting from this alignment help the system make accurate entailment and contradiction decisions for highly similar pairs of sentences. One such example given is the context sentence "Several men in front of a white building" compared to the hypothesis sentence "Several people in front of a gray building."
For the MCScript task~\citeA{ostermannMCScriptNovelDataset2018} used in SemEval 2018, online results\footnote{\url{https://competitions.codalab.org/competitions/17184\#results}} indicate that the best performer~\cite{chenHFLRCSystemSemEval20182018} achieved an accuracy of 84.13\% using a bidirectional LSTM-based approach with an attention layer.

RNNs with attention also have limitations particularly when the alignment between inputs and outputs is not straightforward. For example, 
\citeA{chenHFLRCSystemSemEval20182018} found that yes/no questions were particularly challenging in MCScript, as they require a special handling of negation and deeper understanding of the question. Further, since the multiple choices to answer questions in MCScript are human-authored instead of extracted directly from the accompanying passage like other QA benchmarks, this caused some difficulties in connecting words back to the passage that stemming alone could not fix.

\paragraph{Self-attention in transformers.}
In lieu of adding attention mechanisms to a typical neural model such as an RNN, LSTM, or CNN, the recently proposed transformer architecture is composed entirely of attention mechanisms \cite{vaswaniAttentionAllYou2017}.~\footnote{An excellent post on the implementation of the transformer can be found at \url{https://nlp.seas.harvard.edu/2018/04/03/attention.html}.} One key difference is the self-attention layer in both the encoder and the decoder. For each word position in an input sequence, self-attention allows it to attend to all positions in the sequence to better encode the word. It provides a method to potentially capture long-range dependencies between words, such as syntactic, semantic, and coreference relations. Furthermore, instead of performing a single attention function, the transformer performs multi-head attention in the sense that it applies the attention function multiple times with different linear projections, and therefore allows the model to jointly capture different attentions from different subspaces, e.g., jointly attend to information that might indicate both coreference and syntactic relations.

Another big benefit of the transformer is its suitability for parallel computing. The sequence models such as RNN and LSTM by their sequential nature make it difficult for parallelization. The transformer, which uses attention to capture global dependencies between inputs and outputs, maximizes the amount of parallelizable computations. Empirical results on NLP tasks such as machine translation and constituency parsing have shown impressive performance gains with a significant reduction in training costs \cite{vaswaniAttentionAllYou2017}. Transformers have recently been used in pre-trained contextual models like \textsc{GPT}~\cite{radfordImprovingLanguageUnderstanding2018a} and \textsc{BERT}~\cite{devlinBERTPretrainingDeep2018} to achieve state-of-the-art performance on many of the surveyed language benchmarks.

\subsubsection{Memory Augmentation}\label{sec:memory}
Mentioned earlier, a popular type of approach for tasks which require comprehending passages with several state changes or supporting facts, such as bAbI \cite{westonAICompleteQuestionAnswering2016} or ProPara \cite{mishraTrackingStateChanges2018}, involves augmenting systems with a dynamic memory which may be maintained over time to represent the changing state of the world. \textcolor{black}{Attention mechanisms are often applied to input text to identify relevant locations or entities in memory to read or update.} We discuss several recent architectures which adopt this strategy to highlight key characteristics of such an approach.

\paragraph{Memory networks.}
Memory networks by \citeA{westonMemoryNetworks2015}, introduced as high-performing baseline approaches to both bAbI \cite{westonAICompleteQuestionAnswering2016} and CBT \cite{hillGoldilocksPrincipleReading2015}, track the world state by adding a long-term memory component to the typical network architecture. A memory network consists of a memory array, an input feature map, a generalization module which updates the memory array given new input, an output feature map, and a response module which converts output to the appropriate response or action. The networks can take characters, words, or sentences as input. Each component of the network can take different forms, but a common implementation is for them to be neural networks, in which case the network is called a \textsc{MemNN}. \textcolor{black}{An end-to-end variant of memory networks, which can be used for a higher variety of tasks, e.g., language modeling, has also been proposed \cite{sukhbaatarEndToEndMemoryNetworks2015}.}

The ability to maintain a long-term memory provides more involved tracking of the world state and context. On the CBT cloze task \cite{hillGoldilocksPrincipleReading2015}, it is demonstrated that memory networks can outperform primarily RNN- and LSTM-based approaches in predicting missing named entities and common nouns, and this is because memory networks can efficiently leverage a wider context than these approaches in making inferences. When tested on bAbI, \textsc{MemNN}s also achieved high performance and outperformed LSTM baselines, and on some tasks were able to achieve high performance with fewer training examples than provided \cite{westonAICompleteQuestionAnswering2016}.

\paragraph{Recurrent entity networks and variants.}
The recurrent entity network (\textsc{EntNet}) by \citeA{henaffTrackingWorldState2017} is composed of several dynamic memory cells, where each cell learns to represent the state or properties concerning entities mentioned in the input. Each cell is a gated RNN which only updates its content when new information relevant to the particular entity is received. Further, \textsc{EntNet}'s memory cells run in parallel, allowing multiple locations of memory to be updated at the same time.

\textsc{EntNet} is one of the first models to pass all twenty subtasks (i.e., achieve at least 95\% accuracy on all twenty subtasks) in bAbI \cite{westonAICompleteQuestionAnswering2016}, achieving 99.5\% accuracy over all questions. It also achieves impressive results on CBT \cite{hillGoldilocksPrincipleReading2015}, outperforming memory network baselines on both benchmarks. \textsc{EntNet} is also used as a baseline in the Story Commonsense benchmark \cite{rashkinModelingNaivePsychology2018} in an attempt to track the motivations and emotions of characters in stories from ROCStories \cite{mostafazadehCorpusClozeEvaluation2016} with some success. An advantage of \textsc{EntNet} is that it maintains and updates the state of the world as it reads the text, unlike memory networks, which can only perform reasoning when the entire supporting text and the question are processed and loaded to the memory. For example, given a supporting text with multiple questions, \textsc{EntNet} does not need to process the input text multiple times to answer these questions, while memory networks need to re-process the whole input for each question.

While \textsc{EntNet} achieved state-of-the-art performance on bAbI, it does not perform so well on ProPara \cite{mishraTrackingStateChanges2018}, another benchmark which requires tracking the world state. According to \citeA{dasBuildingDynamicKnowledge2019}, a drawback of \textsc{EntNet} is that while it maintains memory registers for entities, it has no separate embedding for individual states of entities over time. They further explain that \textsc{EntNet} does not explicitly update coreferences in memory, which can certainly cause errors when reading human-authored text which is rich in coreference, as opposed to the simply-structured, automatically-generated bAbI data.

\textcolor{black}{A similar model to \textsc{EntNet} is the query reduction network (QRN) by \citeA{seoQueryReductionNetworksQuestion2017}, which also uses several recurrent units as memory cells controlled by an RNN to track entities in a procedural text. Like \textsc{EntNet}, the memory cells run in parallel, eliminating the vanishing gradient problem which makes it difficult for entirely RNN-based models to track long-term dependencies in procedural text. The entity representations held by the model can be queried at any point while processing the procedural text to predict states and locations of objects, not only at the end of the text as required by typical procedural text benchmarks, e.g., bAbI \cite{westonAICompleteQuestionAnswering2016}. QRNs achieve comparable results to \textsc{EntNet} on bAbI.}

\textcolor{black}{Another similar variant, the original Differentiable Neural Computer (DNC) model proposed by \citeA{gravesHybridComputingUsing2016} uses several recurrent units as memory cells. Unlike the other models, DNC's memory cells are controlled by a central LSTM, whereas nearly all of \textsc{EntNet}'s computation occurs in the memory cells \cite{henaffTrackingWorldState2017}. While \textsc{EntNet} achieves higher performance on the bAbI benchmark \cite{westonAICompleteQuestionAnswering2016} than the original DNC, an optimized, bidirectional form of the DNC proposed by \citeA{frankeRobustScalableDifferentiable2018} exceeds the maximum performance on bAbI over all previous models, achieving a response word error rate of about 0.4. 
This version solves bAbI so close to perfectly that we are unlikely to see any new models attempt the benchmark.}

\paragraph{Memory, attention, and composition networks.}
\textcolor{black}{Similar models which may be useful toward NLI have been developed in other areas like visual question answering (VQA). One such example is the memory, attention, and composition (MAC) network \cite{hudsonCompositionalAttentionNetworks2018}. The MAC network consists of chained memory cells each consisting of a separate memory and attentive controller. The controller attends upon the input question to update a hidden control state which determines the type of reasoning required for the question, while read and write units extract information from a knowledge source (e.g., an image or paragraph of text) guided by the controller and write it to memory. While originally applied to a VQA task, it exceeded the performance of similar network architectures which have been applied to language problems, e.g., DNC \cite{gravesHybridComputingUsing2016}. As such, MAC networks may also be useful for tracking the world state in procedural text.}

\paragraph{Neural process networks.}
\textcolor{black}{Closely related to these models are neural process networks (NPNs) introduced by \citeA{bosselutSimulatingActionDynamics2017}, which aim to simulate the effects of actions on entities in procedural text. It embeds input sentences in a GRU, then uses a multi-layer perceptron (MLP) to predict actions which occur in each sentence, and sentence-level and recurrent attention mechanisms to identify entities affected by actions. Selected actions and entities are fed into a simulation module which uses a recurrent unit and external knowledge of state changes caused by actions to maintain embeddings for the states of all entities in the procedural text. Unlike previous similar models, neural process networks also learn functional operators to represent actions, allowing them to explain the properties of an object that are changed by each action in a procedural text. Their results on a dataset of cooking recipes \cite{kiddonGloballyCoherentText2016} demonstrate that learning these action embeddings allows the model to learn a more informative representation of procedural text. This representation not only allows the model to understand the text, but predict future actions in the text. Models with this suitability for learning action verb semantics may be useful for explaining predictions on NLI benchmarks requiring intuitive physics, a capability that many current state-of-the-art models do not have.}

\paragraph{\textsc{KG-MRC}.}
The Knowledge Graph-Machine Reading Comprehension (\textsc{KG-MRC}) system from \citeA{dasBuildingDynamicKnowledge2019}, maintains a dynamic memory similar to memory networks. However, this memory is in the form of knowledge graphs generated after every sentence of procedural text, leveraging research efforts from the area of information extraction. Generated knowledge graphs are bipartite, connecting entities in the paragraph with their locations. Connections between entities and locations are updated to generate a new graph after each sentence. According to the official ProPara leaderboard\footnote{\url{https://leaderboard.allenai.org/propara/submissions/public}}, the Knowledge Graph-Machine Reading Comprehension (\textsc{KG-MRC}) system from \citeA{dasBuildingDynamicKnowledge2019} achieves the second-highest F-measure on the benchmark, reported as 57.60 on the leaderboard. It provides advantages over \textsc{ProStruct}, the previous state of the art for ProPara \cite{tandonReasoningActionsState2018}. While \textsc{ProStruct} manually enforces hard and soft commonsense constraints, further investigation suggests that \textsc{KG-MRC} learns these constraints automatically, violating them less often than \textsc{ProStruct} \cite{dasBuildingDynamicKnowledge2019}. This shows that the use of the recurrent graph representation helps the model learn these constraints, perhaps better than can be manually enforced. Further, since \textsc{KG-MRC} includes a trained reading comprehension model, it can likely better track changes in coreference which often occur in these texts.

\paragraph{\textsc{NCET}.}
\textcolor{black}{The Neural Conditional Random Field (CRF) Entity Tracking (\textsc{NCET}) model by \citeA{guptaTrackingDiscreteContinuous2019} instead tracks each entity's state and location through per-entity bidirectional LSTMs. These LSTMs are trained using new, crowdsourced annotations for state and location changes of entities in the training paragraphs. A neural CRF maintains the global structure of all object states, enforcing physical laws such as their example that an object cannot move after being destroyed.}
\textcolor{black}{According to the ProPara leaderboard, \textsc{NCET} is the current state of the art on the benchmark, achieving an F-measure of 62.50, a fair improvement over \textsc{KG-MRC}. }

\subsubsection{Contextual Models and Representations}\label{sec:transfer learning}
One of the most exciting recent advances in NLP is the development of pre-trained models and embeddings that can be used as features or further fine-tuned for downstream tasks. These models are often trained based on a large amount of unsupervised textual data. The earlier pre-trained word embedding models such as word2vec \cite{mikolovEfficientEstimationWord2013} and GloVe \cite{penningtonGloveGlobalVectors2014} have been widely applied. However, these models are context-independent, meaning the same embedding is used in different contexts, and they therefore cannot capture different word senses. More recent work has addressed this problem by pre-training models that can provide a word embedding based on context. The most representative models are \textsc{ELMo}, \textsc{GPT}, \textsc{BERT} and its variants, and \textsc{XLNet}. Next, we give a brief overview of these models and summarize their performance on the selected benchmark tasks. Table~\ref{tbl:bert comparison} quantitatively compares the performance of these models on various benchmarks.

\paragraph{\textsc{ELMo}.}

The characteristic contribution of Embeddings from Language Models (\textsc{ELMo}) is its contextual word embeddings, which each rely on the entire input sentence they belong to \citeA{petersDeepContextualizedWord2018}. These embeddings are calculated from learned weights in a bidirectional LSTM which is pre-trained on the One Billion Word language modeling benchmark \cite{chelbaOneBillionWord2014}. Simply adding these embeddings to the input features of previous state-of-the-art systems improved performance, suggesting that they are indeed successful in representing word context. An investigation by the authors shows that \textsc{ELMo} embeddings make it possible to identify word sense and POS, further supporting this.

When the \textsc{ELMo} embedding system was originally released, it helped exceed the state of the art on several benchmarks in the areas of question answering, textual entailment, and sentiment analysis. These included SQuAD \cite{rajpurkarKnowWhatYou2018} and SNLI \cite{bowmanLargeAnnotatedCorpus2015}. All of these approaches have since been exceeded. ELMo still commonly appears in baseline approaches to benchmarks, e.g., to SWAG \cite{zellersSWAGLargeScaleAdversarial2018} and CommonsenseQA \cite{talmorCommonsenseQAQuestionAnswering2019}. It is often combined with the enhanced LSTM-based models such as the ESIM model \cite{chenEnhancedLSTMNatural2017}, or the CNN- and bidirectional GRU-based models such as the DocQA model \cite{clarkSimpleEffectiveMultiParagraph2018}.

\paragraph{\textsc{GPT}.}
The Generative Pre-trained Transformer (\textsc{GPT}) model by \citeA{radfordImprovingLanguageUnderstanding2018a} uses the transformer architecture originally proposed by \citeA{vaswaniAttentionAllYou2017}, in particular the decoder. This system is  pre-trained on a large amount of open online data unsupervised, then fine-tuned to various benchmark datasets. Unlike \textsc{ELMo}, \textsc{GPT} learns its contextual embeddings in an unsupervised setting, which allows it to learn features of language without restrictions. The creators found that this technique produced more discriminative features than supervised pre-training when applied to a large magnitude of clean data. The transformer architecture itself can then be easily fine-tuned in a supervised setting for downstream tasks, which \textsc{ELMo} is not as suitable for, and instead should be used for input features to a separate task-specific model.

When \textsc{GPT} was first released, it pushed the state of the art forward on 12 benchmarks in textual entailment, semantic similarity, sentiment analysis, and more. These included SNLI \cite{bowmanLargeAnnotatedCorpus2015}, MultiNLI \cite{williamsBroadCoverageChallengeCorpus2017}, SciTail \cite{khotSciTailTextualEntailment2018}, the Story Cloze Test \cite{mostafazadehCorpusClozeEvaluation2016}, COPA \cite{roemmeleChoicePlausibleAlternatives2011}, and GLUE \cite{wangGLUEMultiTaskBenchmark2018}. 
\textsc{GPT} holds high positions on several other NLI benchmark leaderboards as well. It is often commonly used as a baseline for new benchmarks, e.g., for CommonsenseQA \cite{talmorCommonsenseQAQuestionAnswering2019}.

\citeA{radfordImprovingLanguageUnderstanding2018a} identify several limitations of the model. First, the model has high computational requirements, which is undesirable for obvious reasons. Second, data from the Internet, which the model is pre-trained on, are incomplete and sometimes inaccurate. Lastly, like many deep learning NLP models, \textsc{GPT} shows some issues with generalizing over data with high lexical variation.

To improve the generalization ability and develop upon the use of unsupervised training settings, the larger \textsc{GPT 2.0} was later released \cite{radfordLanguageModelsAre2019}, which is highly similar to the original implementation, but with significantly more parameters and formulated as a language model. The expanded model has achieved new state-of-the-art results on several language modeling tasks, including CBT~\cite{hillGoldilocksPrincipleReading2015}, LAMBADA~\cite{papernoLAMBADADatasetWord2016}, and the 2016 Winograd Schema Challenge~\cite{davisFirstWinogradSchema2017}. Further, it exceeds three out of four baseline approaches to CoQA \cite{reddyCoQAConversationalQuestion2018} in an unsupervised setting, i.e., trained only on documents and questions, not answers. In a supervised training setting, the model would be fed the answers directly with the questions so that model parameters could be updated based upon correlations between them. The model instead learns to perform the task by observing natural language demonstrations of it without being told where the questions and answers are. This way, it is ensured that the model is not overfitting to superficial correlations between questions and answers. A qualitative investigation into model predictions suggests that some heuristics are indeed being learned to answer questions. For example, if asked a "who" question, the model has learned to return the name of a person mentioned in the passage that the question is posed on. This provides some evidence of the model performing genuine reasoning. 

\textcolor{black}{Upon reporting their results, the authors of \textsc{GPT 2.0} were hesitant to release their full pre-trained model, training datasets, or training code, claiming that the model's impressive text generation ability could be used for malicious purposes such as fake news articles \cite{makWhenTechnologyToo2019}. However, this makes it impossible to reproduce or study their results. Critics have argued that releasing such models is important for the community to develop defenses against such attacks \cite{zellersWhyWeReleased2019}. To demonstrate this, \citeA{zellersDefendingNeuralFake2019} released a highly similar text generation model called \textsc{Grover}, and showed that the best defense against it is actually itself. \textsc{Grover} can be used to discriminate fake news generated by itself from real news up to 92\% accuracy, while other state-of-the-art discriminative models can only achieve up to 73\% accuracy. The full pre-trained models and code were finally released in November 2019,\footnote{See \url{https://openai.com/blog/gpt-2-1-5b-release/}.} but this debate has started an important discussion about the potential dangers of and defenses against the current state-of-the-art language models.}

\paragraph{\textsc{BERT}.}

The recent Bidirectional Encoder Representations from Transformers (\textsc{BERT}) model proposed by \citeA{devlinBERTPretrainingDeep2018} provides several advantages over past state-of-the-art systems with pre-trained contextual embeddings. First, it is pre-trained on larger data than previous competitive systems like \textsc{GPT} \cite{radfordImprovingLanguageUnderstanding2018a}. Where \textsc{GPT} is pre-trained with a large text corpus, \textsc{BERT} is trained with two larger corpora of passages on two tasks: a cloze task where input tokens are randomly masked, and a sentence ordering task where given two sentences, the system must predict whether the second sentence could come directly after the first. This sort of transfer learning from large-scale supervised tasks has been demonstrated several times to be effective in NLP problems. Similar to \textsc{GPT}, \textsc{BERT} was further fine-tuned for each of the 11 commonsense benchmark tasks it originally attempted.

Second, \textsc{BERT} uses a bidirectional form of the transformer architecture \cite{vaswaniAttentionAllYou2017} for pre-training contextual embeddings. This better captures context, an advantage that previous competitive approaches like \textsc{GPT} \cite{radfordImprovingLanguageUnderstanding2018a} and ELMo \cite{petersDeepContextualizedWord2018} do not have. Instead, \textsc{GPT} uses a left-to-right transformer, while ELMo uses a concatenation of left-to-right and right-to-left LSTMs.

Lastly, the structure of \textsc{BERT}'s input embedding is advantageous for representing context, which beyond a traditional embedding for tokens, also incorporates a learned embedding for the sentence a token belongs to (i.e., in sentence pair tasks), and learned positional embeddings for tokens. 
Consequently, the embedding can capture each unique word and its context, perhaps in a more sophisticated way than previous systems. Further, it can flexibly represent a sentence or a pair of sentences, advantageous for solving a wide variety of language processing tasks in question answering, textual entailment, and more.

Upon its initial release, \textsc{BERT} exceeded the state-of-the-art accuracy on several benchmarks including GLUE \cite{wangGLUEMultiTaskBenchmark2018}, SQuAD 1.1 \cite{rajpurkarSQuAD1000002016}, and SWAG \cite{zellersSWAGLargeScaleAdversarial2018} benchmarks. 
\textcolor{black}{After this, it topped several other leaderboards, including those of CLOTH~\cite{xieLargescaleClozeTest2017},\footnote{\url{https://www.qizhexie.com/data/CLOTH_leaderboard}} DREAM~\cite{sunDREAMChallengeDataset2019},\footnote{\url{https://dataset.org/dream/}} OpenBookQA~\cite{mihaylovCanSuitArmor2018},\footnote{\url{https://leaderboard.allenai.org/open\_book\_qa/submissions/public}} CoQA~\cite{reddyCoQAConversationalQuestion2018}, \footnote{\url{https://stanfordnlp.github.io/coqa/}} ReCoRD~\cite{zhangReCoRDBridgingGap2018},\footnote{\url{https://sheng-z.github.io/ReCoRD-explorer/}} and SQuAD 2.0 \cite{rajpurkarKnowWhatYou2018}. \footnote{\url{https://rajpurkar.github.io/SQuAD-explorer/}}}  
\textcolor{black}{Now, variants of \textsc{BERT} have actually topped most of the leaderboards of our surveyed benchmarks at least once, usually several times, especially for the benchmarks originally attempted by the model, e.g., GLUE. Now, \textsc{BERT} is seen as a baseline approach for new benchmarks, such as AlphaNLI \cite{ch2019abductive} and SocialIQA \cite{sapSocialIQACommonsenseReasoning2019}. In the latter, \textsc{BERT} was also used for a transfer learning experiment, where the model is fine-tuned to SocialIQA before fine-tuning to a new task. From this, new state-of-the-art results were achieved on COPA \cite{roemmeleChoicePlausibleAlternatives2011} and the Winograd Schema Challenge \cite{davisFirstWinogradSchema2017}.}

\textcolor{black}{While variants of \textsc{BERT} are still the state of the art on many of the surveyed benchmarks, its deep, complex structure makes it unexplainable. According to \citeA{devlinBERTPretrainingDeep2018}, a goal of future work will be to determine whether \textsc{BERT} truly captures the intended semantic phenomena in benchmark datasets. Unfortunately, results from an early investigation by \citeA{nivenProbingNeuralNetwork2019} suggest that like many neural models for language tasks, \textsc{BERT} exploits superficial correlations in data to achieve high performance, particularly on the ARCT benchmark \cite{habernalArgumentReasoningComprehension2018}. More investigations like these for more state-of-the-art neural models on various benchmarks will be useful to gauge how much progress the models have actually made in performing genuine inference.}

\paragraph{\textsc{MT-DNN}.}
A variant of \textsc{BERT} called the Multi-Task Deep Neural Network (\textsc{MT-DNN}) by \citeA{liuMultiTaskDeepNeural2019} later achieved competitive performance on several leaderboards such as for SciTail~\footnote{\url{https://leaderboard.allenai.org/scitail/submissions/public}}, SNLI \cite{bowmanLargeAnnotatedCorpus2015}~\footnote{\url{https://nlp.stanford.edu/projects/snli/}}, and GLUE \cite{wangGLUEMultiTaskBenchmark2018},\footnote{https://gluebenchmark.com/leaderboard} beating the original implementation of \textsc{BERT}. It appears that the performance gain from the \textsc{MT-DNN} model can mostly be attributed to adding multi-task learning during fine-tuning procedures. This is done through task-specific layers which generate representations for specific tasks, e.g., text similarity and sentence pair classification. While pre-training the bi-directional transformer helps \textsc{BERT} learn universal word representations which are applicable across several tasks, multi-task learning prevents the model from overfitting to particular tasks during fine-tuning, thus allowing it to leverage more cross-task data.

\paragraph{\textsc{XLNet}.}
\textcolor{black}{A more recent model, \textsc{XLNet} \cite{yangXLNetGeneralizedAutoregressive2019}, exceeded the performance of \textsc{BERT} on several benchmarks upon its release, including GLUE \cite{wangGLUEMultiTaskBenchmark2018}, SQuAD 1.1 \cite{rajpurkarSQuAD1000002016}, and SQuAD 2.0 \cite{rajpurkarKnowWhatYou2018}, and DREAM \cite{sunDREAMChallengeDataset2019}, exceeding human performance on the former two. This is done primarily by using a new pre-training method. A key limitation that \citeauthor{yangXLNetGeneralizedAutoregressive2019} identify in \textsc{BERT} is that since tokens are randomly masked for pre-training tasks, there is a discrepancy between language data in pre-training and fine-tuning. Further, as \textsc{BERT} assumes independence of these masked predicted tokens, it may miss important relations between these masked tokens, e.g., if the tokens masked are \textit{New} and \textit{York} in the phrase ``New York'' \cite{yangXLNetGeneralizedAutoregressive2019}. To resolve this issue, \citeauthor{yangXLNetGeneralizedAutoregressive2019} use an autoregressive language model, which can be trained from left and right context. Instead of a bidirectional transformer, they use the newer Transformer-XL, which models long-term dependencies beyond a fixed context, as the main architecture \cite{daiTransformerXLAttentiveLanguage2019}. While they lose the benefit of training on the text in both directions jointly, they propose a new permutation technique which maximizes the likelihood of a sequence of tokens and all of its permutations. This allows the model to jointly train on context from both the left and right, capturing some advantages of both an autoregressive model and a bidirectional model, and resolving the discrepancy between pre-training and fine-tuning data. Further, the permutation technique allows \textsc{XLNet} to capture dependencies between predicted tokens, and the authors show that the model generally captures more dependencies between tokens than previous models.}

\paragraph{\textsc{RoBERTa}.}
\textcolor{black}{The original implementation and architecture of \textsc{BERT} has been outperformed by several variants and other transformer-based models, some of which have been discussed earlier in this section. However, an updated version with only an optimized pre-training approach, called the Robustly Optimized BERT Approach (\textsc{RoBERTa}), has recently become the state of the art for some benchmarks \cite{liuRoBERTaRobustlyOptimized2019}, returning to the top of the leaderboards for RACE~\cite{laiRACELargescaleReAding2017},\footnote{\url{http://www.qizhexie.com/data/RACE_leaderboard.html}} SWAG \cite{zellersSWAGLargeScaleAdversarial2018}, and GLUE \cite{wangGLUEMultiTaskBenchmark2018}, exceeding human performance on the latter two. \textsc{RoBERTa} has also topped other leaderboards since then, such as for WinoGrande \cite{sakaguchi2019winogrande}.\footnote{\url{https://leaderboard.allenai.org/winogrande/submissions/get-started}} Some of the primary changes made to the pre-training approach include randomizing masked tokens in the cloze pre-training task for each epoch rather than keeping them the same over epochs, and adding an additional task to the next sentence prediction pre-training task, where the model must predict whether a candidate next sentence comes from the same document or not. These changes along with greater attention to hyperparameter choices allowed them to pre-train \textsc{BERT} to a higher potential than previously done.}

\paragraph{\textsc{ALBERT}.}
\textcolor{black}{Many of the improvements in transformer-based models come from training more parameters, but this trend becomes more difficult over time due to hardware limitations \cite{lan2019albert}. A Lite BERT (\textsc{ALBERT}) by \citeA{lan2019albert} implements several novel parameter reduction techniques to increase the training speed and efficiency of \textsc{BERT}, allowing the model to scale up much deeper than the original large form of \textsc{BERT} while having fewer parameters. Further, in the sentence ordering pre-training task, they use a new self-supervised loss based upon objectives for discourse coherence. These improvements lead to superior performance on several leaderboards, including GLUE~\cite{wangGLUEMultiTaskBenchmark2018}, SQuAD 2.0~\cite{rajpurkarKnowWhatYou2018}, RACE~\cite{laiRACELargescaleReAding2017}, and DROP~\cite{duaDROPReadingComprehension2019}.\footnote{\url{https://leaderboard.allenai.org/drop/submissions/public}}}

\begin{table}[H]\scriptsize
\centering
\begin{tabular}{P{1.7cm}P{1cm}P{0.9cm}P{0.9cm}P{1.1cm}P{1.3cm}P{1.1cm}P{1.1cm}P{0.9cm}P{1cm}}\toprule
\textbf{Benchmark}  & \textbf{\thead{Simple\\Baseline}} & \textbf{ELMo}  & \textbf{\textsc{GPT}}   & \textbf{\textsc{BERT}} & \textbf{\textsc{MT-DNN}} & \textbf{\textsc{XLNet}} & \textbf{\textsc{RoBERTa}} & \textbf{\textsc{ALBERT}} & \textbf{Human} \\ \toprule

\href{https://www.qizhexie.com/data/CLOTH_leaderboard}{\textbf{CLOTH}}      & 25.0              & 70.7  & --         & \textbf{86.0}         & --           & --          & --      & -- & 85.9  \\ \midrule
\href{https://wilburone.github.io/cosmos/}{\textbf{Cosmos QA}} & -- & -- & 54.5 & 67.1 & -- & -- & -- & -- & 94.0 \\\midrule
\href{https://dataset.org/dream/}{\textbf{DREAM}}      & 33.4            & 59.5  & 55.5        & 66.8       & --           & \textbf{72.0}          & --     & -- & 95.5  \\ \midrule
\href{https://gluebenchmark.com/leaderboard}{\textbf{GLUE}}       & --              & 70.0  & --       & 80.5  & 87.6         & 88.4        & 88.5    & \textbf{89.4} & 87.1  \\ \midrule
\href{https://rowanzellers.com/hellaswag/#leaderboard}{\textbf{HellaSWAG}}  & 25.0              & 33.3  & 41.7        & 47.3       & --           & --          & \textbf{85.2}    & & 95.6  \\ \midrule
\href{https://leaderboard.allenai.org/mctaco/submissions/public}{\textbf{MC-TACO}} & 17.4 & 26.4 & -- & 42.7 & -- & -- & \textbf{43.6} & -- & 75.8 \\\midrule
\href{https://www.qizhexie.com/data/RACE_leaderboard}{\textbf{RACE}}       & 24.9            & --    & 59.0          & 72.0         & --           & 81.8        & 83.2    & \textbf{89.4} & 94.5  \\\midrule
\href{https://leaderboard.allenai.org/scitail/submissions/public}{\textbf{SciTail}}    & 60.3            & --  & 88.3       & --         & 94.1    & --          & --     & -- & --    \\\midrule
\href{https://rajpurkar.github.io/SQuAD-explorer/}{\textbf{SQuAD 1.1}}  & 1.3             & 81.0    & --         & 87.4  & --           & \textbf{89.9}        & --     & -- & 82.3  \\\midrule
\href{https://rajpurkar.github.io/SQuAD-explorer/}{\textbf{SQuAD 2.0}}  & 48.9            & 63.4  & --          & 80.8  & --           & 86.3   & 86.8   & \textbf{89.7} & 86.9  \\\midrule
\href{https://super.gluebenchmark.com/leaderboard}{\textbf{SuperGLUE}}  & 47.1            & --    & --          & 69.0    & --           & --          & \textbf{84.6}    & --  & 89.8  \\\midrule
\href{https://leaderboard.allenai.org/swag/submissions/public}{\textbf{SWAG}}       & 25.0              & 59.1  & 78.0      & 86.3      & 87.1        & --          & \textbf{89.9}  & -- & 88.0    \\\bottomrule
 \normalsize
\end{tabular}
\caption{Comparison of exact-match accuracy achieved on selected benchmarks by a random or majority-choice baseline, various neural contextual embedding models, and humans. 
\textsc{ELMo} refers to the highest-performing listed approach using \textsc{ELMo} embeddings. Best system performance on each benchmark in bold. Information extracted from leaderboards (linked to in the first column) at time of writing (October 2019), and original papers for benchmarks introduced in Section~\ref{sec:benchmarks}.}
\label{tbl:bert comparison}
\end{table}

\paragraph{When to fine-tune.}
These new pre-trained contextual models are applied to benchmark tasks in different ways. Particularly, while \textsc{ELMo} has been traditionally used to generate input features for a separate task-specific model, \textsc{BERT}-based models are typically fine-tuned on various tasks and applied to them directly. Understanding why these choices were made is important for the further development of these models.

\citeA{petersDeepContextualizedWord2018} investigate this difference in training the two models and compare their performance both when just extracting their output as features for another model, against when fine-tuning them to be used directly on various tasks. Their results show that \textsc{ELMo}'s LSTM architecture can actually be fine-tuned and applied directly to downstream tasks like \textsc{BERT} can with some success, although it is more difficult to perform this fine-tuning on \textsc{ELMo}. Further, performance on sentence pair classification tasks like MultiNLI \cite{williamsBroadCoverageChallengeCorpus2017} and SICK \cite{marelliSICKCureEvaluation2014} is shown to be better when the contextual embeddings generated by \textsc{ELMo} are instead used as input features to a separate task-specific architecture. They infer that this may be because the LSTM architecture of \textsc{ELMo} must consider tokens sequentially, rather than being able to compare all tokens to each other across sentence pairs like \textsc{BERT}'s transformer architecture can. \textsc{BERT}'s output can also be used as features for a task-specific model with some success, and it actually outperforms \textsc{ELMo} in most of the studied tasks when used in this way, likely for the same reason. However, they find that performance on sentence similarity tasks like the Microsoft Research Paraphrase Corpus \cite{dolanAutomaticallyConstructingCorpus2005} is significantly better when the model is fine-tuned to the task.

\textcolor{black}{\citeA{liu-etal-2019-linguistic} further investigate the transferability of embeddings produced by these models, and find that even linear models trained with them perform comparably to state-of-the-art task-specific models. They also find that features from the lowest layer of LSTM models like \textsc{ELMo} are most transferable, while higher layers are more task-specific. Meanwhile, features from the middle layers of transformer models like \textsc{BERT} are most transferable, and embeddings are not observed to become more task-specific through layers. This may shed some more light on performance differences of these models when used for pre-trained embeddings compared to fine-tuning them on a particular task. More investigation into these contextual models will be invaluable to the research community as they continue to gain popularity.}

\subsection{Incorporating External Knowledge}\label{sec:knowledge base approaches}

\textcolor{black}{One challenge of the current trend of work is the disconnect between knowledge resources and approaches taken to tackle those benchmark tasks. As outlined in Section~\ref{sec:benchmarks}, most of our surveyed benchmarks require a large amount of external knowledge to be solved by humans. Surprisingly, many of the recent approaches, especially neural approaches, have relied on only the benchmark training data and some linguistic resources, usually pre-trained word embeddings, to build models for reasoning and inference. Despite the availability of common and commonsense knowledge resources discussed in Section~\ref{sec:resources}, none of them are actually applied to achieve state-of-the-art performance on the benchmark tasks, and only a few of them are applied in any recent approaches. } 

\paragraph{Use of linguistic resources.}
\textcolor{black}{
WordNet \cite{millerWordNetLexicalDatabase1995} is perhaps the most widely applied lexical resource, and its word relations are particularly useful for textual entailment problems. As such, WordNet has made an appearance in early statistical approaches all throughout the RTE Challenges \cite{daganPASCALRecognisingTextual2005,hicklRecognizingTextualEntailment2006,giampiccoloFourthPASCALRecognizing2008,ifteneUAICParticipationRTE42008,bentivogliSeventhPASCALRecognizing2011,tsuchidaIKOMATAC2011Method2011}. WordNet has also been shown to improve performance of neural models on more recent benchmarks. For example, \citeA{bauerCommonsenseGenerativeMultiHop2018} use relations from WordNet to help answer questions requiring external knowledge in NarrativeQA \cite{kociskyNarrativeQAReadingComprehension2018}. The Knowledge-Based Inference Model (KIM), a more involved approach by \citeA{chenNeuralNaturalLanguage2018}, improves performance on the SNLI \cite{bowmanLargeAnnotatedCorpus2015} and MultiNLI \cite{williamsBroadCoverageChallengeCorpus2017} benchmarks by using WordNet throughout. Pairs of words in the premise and hypothesis texts are aligned using knowledge-enriched co-attention, where if a relation exists in WordNet between two words in a pair, they have a higher attention score. Later, they use the content of these relations to help infer a class label of entailment, contradiction, or neutral. FrameNet \cite{fillmoreFrameNetDatabaseSoftware2002} has also been a useful resource for external knowledge, particularly about the semantics of common events. For example, \citeA{botschenFrameEntityBasedKnowledge2018} used FrameNet to embed events recognized in the given context, concatenating this embedding to a traditional word embedding input to improve the performance of a baseline model for ARCT \cite{habernalArgumentReasoningComprehension2018}. }

\paragraph{Use of common knowledge resources.}

\textcolor{black}{Popular common knowledge resources such as DBpedia \cite{auerDBpediaNucleusWeb2007} and YAGO \cite{suchanekYAGOCoreSemantic2007} have been used in creating benchmarks \cite{morgensternPlanningExecutingEvaluating2016,choiQuACQuestionAnswering2018}, but have not been directly applied to solve any benchmarks. A possible exception is the Enhanced Language Representation with Informative Entities (\textsc{ERNIE}) model, which used the common knowledge in Wikidata\footnote{\url{https://www.wikidata.org/}} to create knowledge-enhanced embeddings for entities mentioned in text, but was not applied to our surveyed benchmarks \cite{zhangERNIEEnhancedLanguage2019}. Another related example comes from \citeA{emamiKnowledgeHuntingFramework2018}, which uses information from search engines to aid in solving the Winograd Schema Challenge \cite{davisFirstWinogradSchema2017}.}

\textcolor{black}{Most approaches gain relevant knowledge through benchmark training data and
large pre-training texts only. While it is possible that these large pre-training texts are helpful for acquiring some linguistic and common knowledge, models likely still are lacking commonsense knowledge, which is typically unstated~\cite{cambriaIsanetteCommonCommon2011a}, and thus unlikely to appear in pre-training text, which is largely acquired from the Web.}

\paragraph{Use of commonsense knowledge resources.} 
\textcolor{black}{Cyc \cite{lenatBuildingLargeKnowledgeBased1989} and ConceptNet \cite{liuConceptNetPracticalCommonsense2004} are by far the most talked-about commonsense knowledge resources available. However, Cyc does not appear in any of our surveyed approaches, while ConceptNet has only been occasionally applied. }
\textcolor{black}{ConceptNet is used in a neural baseline approach to OpenBookQA \cite{mihaylovCanSuitArmor2018}, where questions explicitly require outside common and commonsense knowledge. Interestingly, they find that their adding of facts from ConceptNet causes distraction which reduces performance, suggesting that the technique for selecting the appropriate relations is important to reduce distraction. \citeA{bauerCommonsenseGenerativeMultiHop2018} use a more careful technique in a knowledge-enhanced neural approach to NarrativeQA \cite{kociskyNarrativeQAReadingComprehension2018}. For each concept in a question, they use ConceptNet relations to build candidate paths which span to a concept in the given story context, then to another concept in the context, then to an outside concept, which facilitates multi-step reasoning over the context and with outside knowledge. They then use mutual information and term-frequency measures to prune these paths and select the appropriate knowledge to perform this reasoning. ConceptNet can also be used to create knowledge-enhanced word embeddings. A neural model applied to COPA leverages commonsense knowledge from ConceptNet \cite{roemmeleEncoderdecoderApproachPredicting2018} through ConceptNet-based embeddings by \citeA{liStoryGenerationCrowdsourced2013}, which were generated by applying the word2vec skip-gram model \cite{mikolovEfficientEstimationWord2013} to commonsense knowledge tuples in ConceptNet. A recent approach to the Winograd Schema Challenge from \cite{liuCombingContextCommonsense2017} uses a similar technique, as well as a baseline approach to SWAG \cite{zellersSWAGLargeScaleAdversarial2018}. Future work here will require the development of new techniques for selecting the appropriate knowledge from noisy knowledge resources, as well as the use of a greater breadth of knowledge resources. }

\textcolor{black}{Section~\ref{sec:discussion} discusses future directions that put more emphasis on incorporating external knowledge in creating both benchmarks and models. }

%% file: 5-otherbenchmark.tex
\section{Other Related Benchmarks}
While this paper intends to cover language understanding tasks for which some external knowledge or advanced reasoning beyond linguistic context is required, many related benchmarks have not been covered. First of all, nearly all language understanding benchmarks developed throughout the last couple decades could benefit from commonsense knowledge and reasoning. Second, as language communication is integral to other perception and reasoning systems, recent years have also seen an increasing number of benchmark tasks that combine language and vision.

\paragraph{Language-related benchmarks.}
Many early corpora for classical NLP tasks such as semantic role labeling, relation extraction, and paraphrase may also require commonsense knowledge and reasoning, though this was not emphasized or investigated at the time. For example, in creating the Microsoft Research Paraphrase Corpus, \citeA{dolanAutomaticallyConstructingCorpus2005} found that the task of annotating text pairs was difficult to streamline because this often required commonsense, suggesting that the paraphrases within the corpus require commonsense knowledge and reasoning to identify. Some such tasks are actually included in the multi-task benchmarks like Inference is Everything \cite{whiteInferenceEverythingRecasting2017}, GLUE \cite{wangGLUEMultiTaskBenchmark2018}, and DNC \cite{poliak2018emnlp-DNC}. 

More recent related tasks include challenging textual conversational benchmarks. One example is QuAC \cite{choiQuACQuestionAnswering2018}, which asks questions about Wikipedia articles through a discourse for robust contextual question answering, but does not require commonsense in the way that the similar CoQA benchmark does \cite{reddyCoQAConversationalQuestion2018}.\footnote{\citeA{yatskarQualitativeComparisonCoQA2018} give a detailed account of further differences between CoQA and QuAC.} Another related conversation dataset created by \citeA{xuCommonsenseKnowledgeAware2018} explores the use of commonsense knowledge from ConceptNet in producing higher quality and more relevant responses for chatbots.

Besides English, there are also benchmarks in other languages. 
For example, there have been RTE datasets created in Italian\footnote{See \url{https://www.evalita.it/2009/tasks/te}.} and Portuguese\footnote{See \url{https://nilc.icmc.usp.br/assin/}.}, and cross-lingual RTE datasets have appeared in several SemEval shared tasks over the years \cite{negriSemeval2012TaskCrosslingual2012,negriSemeval2013TaskCrosslingual2013,cerSemEval2017TaskSemantic2017} to encourage progress in machine translation and content synchronization. There also exist various cross-lingual knowledge resources, including the latest version of ConceptNet \cite{speerConceptNetOpenMultilingual2017}, which contains relations from several multilingual resources.

\paragraph{Visual benchmarks.}
External knowledge and reasoning play an important role in integrating language and vision, for example, grounding language to perception~\cite{gaoPhysicalCausalityAction2016}, language-based justification for action recognition~\cite{yangCommonsenseJustificationAction2018}, and visual question answering~\cite{kafleVisualQuestionAnswering2017}.
Visual benchmarks requiring external knowledge include VQA benchmarks like the original VQA \cite{agrawalVQAVisualQuestion2015}, other similar VQA datasets like Visual7W \cite{zhuVisual7WGroundedQuestion2016}, and the newer GQA based on scene graphs of images \cite{hudson2018gqa}. There also exist similar datasets with synthetic images, such as CLEVR \cite{johnsonCLEVRDiagnosticDataset2017} and the work by \citeA{suhrCorpusNaturalLanguage2017}. They also include the tasks of commonsense action recognition and justification, which are found in the dataset by 
\citeA{Fouhey18}, and Visual Commonsense Reasoning (VCR) by \citeA{zellersRecognitionCognitionVisual2019}. \textcolor{black}{Some visual datasets are geared toward specific reasoning processes, e.g., the dataset by \cite{winnLighterCanStill2018} for learning the directional changes associated with comparative adjectives for colors;  
and the verb causality dataset by \citeA{gaoWhatActionCauses2018}, which provides both natural language descriptions and images for the effects of actions upon objects.} 

These are all image-based, but we are also beginning to see similar video-based datasets, such as Something Something \cite{goyalSomethingSomethingVideo2017}, which aims to evaluate visual commonsense through over 100,000 videos portraying everyday actions.
We have also seen the vision-and-language navigation (VLN) task such as Room-to-Room (R2R) by \citeA{andersonVisionandLanguageNavigationInterpreting2018}. Such benchmarks are important to promote progress in physically grounded inference. 

%% file: 6-discussion.tex
\section{Conclusion}\label{sec:discussion}

\textcolor{black}{The availability of data and computing resources and the rise of new learning and inference methods make this an unprecedentedly exciting time for research on natural language understanding and inference. 
As more benchmarks become available and performance on the benchmarks keeps growing, one central question is whether the technologies developed are in fact pushing the state-of-the-art in natural language inference. To address this question, in addition to continuous efforts on making bias-free benchmarks (as discussed in Section~\ref{sec:bias}), here are a few directions we think are important to pursue in the future.  }

\vspace{10pt}
\noindent
{\bf Need greater emphasis on external knowledge acquisition and incorporation.}

\noindent
\textcolor{black}{As discussed in Section~\ref{sec:knowledge base approaches}, most approaches to NLI rely on a large amount of pre-training and training data to learn the model. This is far from practical, which brings up some important questions as to how to properly incorporate external knowledge in modern approaches, and how to best acquire relevant external knowledge for the tasks at hand. This will be challenging, since as discussed in Section~\ref{sec:kb completion}, most knowledge resources are incomplete, and do not contain all required information for solving the benchmarks. Beyond attempting to complete or generalize over these resources during inference as discussed in Section~\ref{sec:kb completion}, another potential avenue to address the disconnect between benchmarks and knowledge resources is to jointly develop benchmark datasets and construct knowledge bases. Recently, we are seeing the creation of knowledge graphs geared toward particular benchmarks, like the ATOMIC knowledge graph by \citeA{sapATOMICAtlasMachine2019} which expands upon data in the Event2Mind benchmark \cite{rashkinEvent2MindCommonsenseInference2018}, and ideally provides the required relations that ConceptNet is missing for solving Event2Mind. We are also seeing the creation of benchmarks geared toward particular knowledge graphs, like CommonsenseQA \cite{talmorCommonsenseQAQuestionAnswering2019}, where questions were drawn from subgraphs of ConceptNet \cite{liuConceptNetPracticalCommonsense2004}, thus encouraging the use of ConceptNet in approaching the benchmark tasks. We have also seen benchmarks and knowledge resources created and released together. We have seen benchmarks being released with unstructured corpora containing some required knowledge, such as ARC \cite{clarkThinkYouHave2018} and OpenBookQA \cite{mihaylovCanSuitArmor2018}. A more recent example is SherLIiC \cite{schmitt-schutze-2019-sherliic}, which uses typed relations extracted from Freebase \cite{bollackerFreebaseCollaborativelyCreated2008} to form sentence pairs in a textual entailment problem, and is made available alongside the original relations. A tighter coupling of benchmark tasks and knowledge resources will help understand and formalize the scope of knowledge needed, and facilitate the development and evaluation of approaches that can incorporate external knowledge.
Shared tasks, such as those at the Commonsense Inference in NLP workshop at EMNLP 2019,\footnote{\url{https://coinnlp.github.io/}} have already begun to encourage the use of external knowledge in inference by providing several knowledge resources.}

\textcolor{black}{Since commonsense knowledge is so intuitive for humans, it is difficult for researchers to even identify and formalize the needed knowledge. It may be worthwhile to explore new task formulations beyond text that involve artificial agents (in either a simulated world or the real physical world) which can use language to communicate, to perceive, and to act. Some example tasks can be formed in the context of interactive task learning~\cite{chaiLanguageActionInteractive2018} or embodied question answering~\cite{dasEmbodiedQuestionAnswering2018,gordonIQAVisualQuestion2018}. \citeA{ortizWhyWeNeed2016} further motivates this idea, and provides suggestions of ways to implement a physically embodied Turing Test. Working with agents and observing their abilities and limitations in understanding language and grounding language to their own sensorimotor skills will allow researchers to better understand the space of commonsense knowledge from a practical standpoint, and tackle the problem of knowledge acquisition and language inference jointly.}

\vspace{10pt}
\noindent
{\bf Need greater emphasis on reasoning.}

\noindent
\textcolor{black}{Some recent results have questioned whether state-of-the-art approaches actually perform genuine inference and reasoning for those benchmark tasks. For example, it has shown that models which exploit statistical biases in benchmark data perform poorly when the biases are removed \cite{nivenProbingNeuralNetwork2019,zellers2019hellaswag}. For another example, \citeA{jiaAdversarialExamplesEvaluating2017} propose an adversarial evaluation scheme for SQuAD \cite{rajpurkarSQuAD1000002016} which randomly inserts distractor sentences into passages which do not change the meaning of the passage, and shows that high-performing models on the benchmark drop significantly in performance. \citeA{marasovicNLPGeneralizationProblem2018} highlights several more similar situations where modern NLI systems can break down due to small, inconsequential changes in inputs. All these findings point to an important question: how do we develop and evaluate models that can actually perform reasoning?}

\textcolor{black}{
\citeA{davisCommonsenseReasoningCommonsense2015} point out that in order for machines to perform comprehensive commonsense reasoning, as in the ``piggy bank'' example discussed in Section~\ref{intro}, there is a critical need for methods that can automatically integrate many types of reasoning such as temporal reasoning, plausible reasoning, and analogy. 
Thus, it is important to develop a more thorough understanding of this body of reasoning skills required in NLI. However, this is typically not given much attention in NLI benchmarks. 
Benchmark tasks often only evaluate the end result of inference (e.g., whether a question is correctly answered) without regard to the underlying process. Further, except for a few benchmarks, such as bAbI, which specifies sub-tasks by types of reasoning intended \cite{westonAICompleteQuestionAnswering2016},
and Event2Mind for prediction of specific psychological states surrounding events \cite{rashkinEvent2MindCommonsenseInference2018}, most benchmarks do not differentiate between types of reasoning. 
Useful analyses of required skills are occasionally done after benchmark creation, e.g., as done by \citeA{sugawaraPrerequisiteSkillsReading2017}, \citeA{chuBroadContextLanguage2017}, and \citeA{yatskarQualitativeComparisonCoQA2018} for selected benchmarks. Nonetheless, the majority of benchmarks do not support a systematic evaluation of reasoning abilities. As such, for future benchmarks, it would be helpful to explicitly address different types of reasoning capabilities through data curation, which can provide a better understanding of machines' commonsense reasoning ability and allow for targeted improvement in different areas.}

\vspace{10pt}
\noindent
{\bf Need stronger justification and better understanding on design choices of models.}

\noindent
\textcolor{black}{New neural models are constantly being developed. Very often, the design choices for various models are neither justified nor understood.
For example, design choices like parameter tuning strategies are often overlooked in favor of more significant or interesting model improvements, but these small choices can actually have a significant effect on performance. 
We have seen in other AI subfields that more complex models may lead to better performance on a particular benchmark, but simpler models with better parameter tuning may later lead to comparable results. For example, in image classification, a study by \citeA{brendelApproximatingCNNsBagoflocalFeatures2019} shows that nearly all improvements of recent deep neural networks over earlier bag-of-features classifiers come from better fine-tuning rather than improvements in decision processes. } 

\textcolor{black}{NLI models are similarly vulnerable to these kinds of issues. For example, we have recently seen that the \textsc{BERT} model \cite{devlinBERTPretrainingDeep2018}, was exceeded by several variants and other transformer-based models which made major changes to loss functions \cite{liuMultiTaskDeepNeural2019}, model architecture, pre-training objectives \cite{yangXLNetGeneralizedAutoregressive2019}, and model complexity, i.e., the number of parameters trained \cite{radfordLanguageModelsAre2019}. Even beyond these more substantial changes to the model, smaller tweaks to various aspects of the model have resulted in hundreds of entries on leaderboards (e.g., those linked to in Section~\ref{sec:transfer learning} and Table~\ref{tbl:bert comparison}) leading only to marginal improvements. Recently, though, we saw \textsc{RoBERTa} exceed all of these variants and become the state of the art on several benchmarks using the same architecture as \textsc{BERT}, but making small changes to its pre-training tasks and more carefully selecting hyperparameters \cite{liuRoBERTaRobustlyOptimized2019}. This demonstrates the importance of all model design choices, and puts into question the benefits of changes made in the earlier variants. 
More efforts on theoretical understanding and motivation for model design, pre-training techniques, and parameter tuning would be beneficial to better direct research efforts.}

\vspace{10pt}
\noindent
{\bf Need broader and multidimensional metrics for evaluation.}

\noindent
\textcolor{black}{Evaluation, particularly through leaderboards, has been a critical component in NLI research. The metrics typically used on these leaderboards, e.g., accuracy and others discussed in Section~\ref{sec:eval schemes}, measure \textit{task competence}, i.e., how well the learned model solves the benchmark task. Researchers are motivated to develop models that top the leaderboard by these kinds of performance measures. This opens up many questions about whether this competence-centric evaluation is an effective research practice. We argue that broader and multidimensional evaluation metrics should be considered in the future. }

\vspace{10pt}
\noindent
\textcolor{black}{
{\em Task competence.} A competence-centric evaluation, while important for pushing the state of the art, can also lead to a less productive path if not treated carefully. A significant amount of effort has been spent on tuning models or parameters to surpass other models or human performance without a good understanding of model behaviors, even on benchmarks with obvious drawbacks.    
For example, in Figure~\ref{fig:difficulty popularity 2}, we saw that SQuAD 1.1~\cite{rajpurkarSQuAD1000002016} is clearly the most popular benchmark. It has seen hundreds of leaderboard submissions,\footnote{\url{https://rajpurkar.github.io/SQuAD-explorer/}} and machine performance has surpassed human performance. However, there are known issues with the benchmark. First, SQuAD 1.1 has been considered too simple, as answers to questions can be extracted directly from the text. As such, it was replaced with SQuAD 2.0 \cite{rajpurkarKnowWhatYou2018}, a more interesting version with unanswerable questions. Further, state-of-the-art systems for SQuAD 1.1 have been shown to lack generalization to an adversarial modification of the benchmark data \cite{jiaAdversarialExamplesEvaluating2017}, suggesting that there are statistical biases in the dataset which models are overfitting to. Because of these issues, continuously improving the task competence measure on the leaderboard does not lead to anything insightful, and does not bring us closer to solving the broader task of question answering. 
Rather than submitting to popular leaderboards like SQuAD 1.1 hundreds of times to marginally surpass other models, researchers should be encouraged to work on some new, interesting, and/or adversarially constructed benchmarks, e.g., WinoGrande \cite{sakaguchi2019winogrande} and SocialIQA \cite{sapSocialIQACommonsenseReasoning2019}.
In addition, other evaluation metrics should also be considered in order to provide a better understanding of the progress on natural language inference, as discussed next. }

\vspace{10pt}
\noindent
\textcolor{black}
{ {\em Efficiency.}
Modern neural networks are becoming more and more complex and requiring more and more pre-training data and computational resources, with the cost to train models generally doubling every few months \cite{schwartzGreenAI2019}.}
\textcolor{black}{As outlined by \citeA{rogersHowTransformersBroke2019}, online leaderboards and shared tasks which rank approaches only by task competence measures exacerbate this problem, encouraging researchers with means to either use more data in pre-training or ``throw a deeper network at it'' to achieve state-of-the-art results. This trend discourages those researchers without the means to do this, especially academic research teams or those from emerging economies \cite{schwartzGreenAI2019}, from competing on leaderboards and in shared tasks. 
Factoring computational efficiency into evaluation, however, will reward approaches which make better use of available training data rather than continually augmenting pre-training data, and allow greater participation of researchers with limited computational resources. Additionally, this would encourage greener practices in AI \cite{schwartzGreenAI2019}, as training these models has a substantial carbon footprint \cite{strubellEnergyPolicyConsiderations2019}.}

\textcolor{black}{In fact, we are recently beginning to see an emphasis on model efficiency in NLI benchmarks and approaches. Some leaderboards, such as for SocialIQA \cite{sapSocialIQACommonsenseReasoning2019}\footnote{\url{https://leaderboard.allenai.org/socialiqa/submissions/public}} and AlphaNLI \cite{ch2019abductive}, \footnote{\url{https://leaderboard.allenai.org/anli/submissions/public}} are now reporting runtime during prediction, which is a step in this direction. More efficient models for NLI are also being developed. One example is \textsc{RoBERTa}, where an optimized pre-training approach is developed to better utilize the parameters of the original \textsc{BERT} model~\cite{devlinBERTPretrainingDeep2018} and achieve higher performance \cite{liuRoBERTaRobustlyOptimized2019}. Another example, motivated by hardware limitations, is \textsc{ALBERT}, where a parameter reduction technique allows the use of fewer parameters to achieve comparable results to previous models, and often better results when the model is scaled deeper \cite{lan2019albert}. Lastly, \textsc{DistilBERT} by \citeA{sanhDistilBERTDistilledVersion2019} compresses \textsc{BERT} with a small performance reduction by using model distillation, a model compression technique where a smaller model is trained to mimic the predictions of a larger model. Through this, they reduce model parameters by 40\% and increase prediction speed by 60\%, all while retaining 97\% of its performance on GLUE \cite{wangGLUEMultiTaskBenchmark2018}. Efforts like the latter, purely toward model efficiency at the cost of performance, are not rewarded by the current evaluation criteria of leaderboards. Going forward, we hope such approaches will begin receiving greater recognition.}

\vspace{10pt}
\noindent
\textcolor{black}{
{\em Transparency.} As in many sub-fields of AI where deep learning approaches are heavily applied, transparency of the model and the ability to explain model behaviors are important.
The underlying reasoning process of state-of-the-art models, e.g., \textsc{BERT} ~\cite{devlinBERTPretrainingDeep2018}, cannot be interpreted due to their deep, bidirectional architectures, and thus it is quite opaque why certain conclusions are made by these models.}
\textcolor{black}{It is unclear if \textsc{BERT} is capturing semantic phenomena or again learning statistical biases in the NLI benchmarks, but evidence from probing studies and adversarial evaluations, e.g., the investigation by \citeA{nivenProbingNeuralNetwork2019}, is beginning to suggest that \textsc{BERT} is indeed learning and exploiting statistical biases in certain benchmarks.} 
\textcolor{black}{A better understanding of the behaviors of NLI systems, especially deep learning models that achieve high performance, is critical to confirm that systems are trustworthy and capable of generalization. As such, evaluating NLI systems upon their ability to explain their underlying reasoning process and more thoroughly prove their understanding could be another direction to pursue. We have seen an analogue of this in visual question answering with the VCR benchmark \cite{zellersRecognitionCognitionVisual2019}, where models not only must answer a natural language question about an image, but must select an appropriate textual explanation for their answer from a set of choices. Although this setup may be similarly vulnerable to models overfitting on statistical artifacts of human language, it is a step in this direction.}

\vspace{10pt}
\noindent
\textcolor{black}{
{\em Generalization ability.} A system that can generalize should be able to perform inferences on new data and tasks with minimal training, as humans do. 
With better generalization, we would expect models to be able to acquire knowledge from any benchmark and apply it to other benchmarks. 
Domain adaptation, i.e., where the distribution of training data differs significantly from the distribution of test data, is a good step toward generalization to unseen data, and \citeA{marasovicNLPGeneralizationProblem2018} highlights recent successful approaches here.
In fact, state-of-the-art models like \textsc{BERT} \cite{devlinBERTPretrainingDeep2018} are getting somewhat closer to this capability, being pre-trained on large text corpora and capable of being fine-tuned to new problems with minimal effort or task-specific modifications. Some metrics that explicitly measure this type of adaptation and generalization will be useful. 
Nonetheless, the requirement of iteration over large amounts of training data in order to perform well on a benchmark is still a limitation for generalizing over multiple problems. 
This perhaps calls for a new practice for future NLI evaluation where no training data is provided, but rather a few examples to show what the evaluation would look like and to resolve engineering and interface issues. This may alleviate some problems caused by superficial data biases and dependence on large training data, and move toward models that can perform reasoning and generalization.}

%% file: 7-acknowledgements.tex
\section*{Acknowledgements}
We would like to thank the anonymous reviewers for their greatly helpful comments and suggestions.

%% file: 8-appendix.tex
\appendix
\section{Comparison of Benchmark Creation Methods}\label{sec:appendix}
The following table gives a chronological summary of methods used in creating benchmarks. It considers the steps of creating the context and/or question, the correct answer and alternatives, and any extra annotations (useful information beyond the answer which the benchmark evaluates upon), as well as validating the data. M refers to manual approaches by experts, A to automatic approaches through language generation, T to text mining, and C to crowdsourcing. Data size is included for comparison of methods used.

\begin{table}[H]\tiny
\vspace{-4em}
\begin{tabular}{P{5.3cm}P{1.2cm}P{1.6cm}P{1.2cm}P{1.1cm}P{1.1cm}P{1.1cm}}\toprule
\textbf{Benchmark (Reference)}                                                            & \textbf{\thead{Data Size}} & \textbf{\thead{Context/Question}} & \textbf{\thead{Answer}} & \textbf{\thead{Alternatives}} & \textbf{\thead{Annotation}} & \textbf{\thead{Validation}} \\ \toprule
\textbf{RTE-1} \cite{daganPASCALRecognisingTextual2005} & 1.37K               & M                & M      & --           & --         & M                    \\\midrule
\textbf{RTE-2} \cite{bar-haimSecondPASCALRecognising2006} & 1.60K             & M                & M      & --           & --         & M                    \\\midrule
\textbf{RTE-3} \cite{giampiccoloThirdPASCALRecognizing2007} & 1.60K           & M                & M      & --           & --         & M                    \\\midrule
\textbf{RTE-4} \cite{giampiccoloFourthPASCALRecognizing2008} & 1.00K          & M                & M      & --           & --         & M                    \\\midrule
\textbf{Conversational Entailment} \cite{zhang-chai-2009-know} & 875          & T, M                & M      & --           & --         & M                    \\\midrule
\textbf{RTE-5} \cite{bentivogliFifthPASCALRecognizing2009} & 1.20K        & M                & M      & --           & --         & M                    \\\midrule
\textbf{RTE-6} \cite{bentivogliSixthPASCALRecognizing2010} & 32.7K           & T                & M      & --           & --         & M                    \\\midrule
\textbf{RTE-7} \cite{bentivogliSeventhPASCALRecognizing2011} & 48.8K         & T                & M, T    & --           & --         & M                    \\\midrule
\textbf{COPA} \cite{roemmeleChoicePlausibleAlternatives2011} & 1.00K        & M                & M      & M            & --         & M                    \\\midrule
\textbf{MCTest} \cite{richardsonMCTestChallengeDataset2013} & 2.00K        & C                & C      & C            & --         & A                    \\\midrule
\textbf{SICK} \cite{marelliSICKCureEvaluation2014} & 9.84K & M, T & C & -- & -- & C \\\midrule 
\textbf{SNLI} \cite{bowmanLargeAnnotatedCorpus2015} & 570K &    T, C              & C      & --           & C         & C                    \\\midrule
\textbf{CBT} \cite{hillGoldilocksPrincipleReading2015} & 687K & T, A & A & A & -- & -- \\\midrule
\textbf{Triangle-COPA} \cite{gordonCommonsenseInterpretationTriangle2016} & 100  & M                & M      & M           & M         & M                    \\\midrule
\textbf{ROCStories} \cite{mostafazadehCorpusClozeEvaluation2016} & 98.2K          & C                & C      & C            & --         & C                    \\\midrule
\textbf{WSC} \cite{morgensternPlanningExecutingEvaluating2016} & 60         & M                & M      & --           & --         & M                    \\\midrule
\textbf{bAbI} \cite{westonAICompleteQuestionAnswering2016} & 40.0K          & A                & A      & --           & --         & --                   \\\midrule
\textbf{LAMBADA} \cite{papernoLAMBADADatasetWord2016} & 10.0K                     & T              & T      & --           & --         & A, C                    \\\midrule
\textbf{JOCI} \cite{zhangOrdinalCommonsenseInference2016} & 39.1K           & A, T              & C      & --           & --         & C                    \\\midrule
\textbf{RACE} \cite{laiRACELargescaleReAding2017} & 97.7K & T & T & T & C & -- \\\midrule
\textbf{CLOTH} \cite{xieLargescaleClozeTest2017} & 99.4K & T & T & T & C & -- \\\midrule
\textbf{IIE} \cite{whiteInferenceEverythingRecasting2017} & 313K & T, A & T, A & -- & -- & M \\\midrule
\textbf{NarrativeQA} \cite{kociskyNarrativeQAReadingComprehension2018} & 46.8K & T, C & C & -- & C & C \\\midrule
\textbf{SciTail} \cite{khotSciTailTextualEntailment2018} & 27.0K & T, C & C & -- & C & C \\\midrule
\textbf{ARC} \cite{clarkThinkYouHave2018} & 7.79K                           & T                & T      & T            & --         & --                   \\\midrule
\textbf{MCScript} \cite{ostermannMCScriptNovelDataset2018} & 13.9K & M, T, C & C & C & C & C \\\midrule
\textbf{Story Commonsense} \cite{rashkinModelingNaivePsychology2018} & 161K	& T & C & -- & C & C \\\midrule
\textbf{Event2Mind} \cite{rashkinEvent2MindCommonsenseInference2018} & 57.1K & T & C & -- & -- & C \\\midrule
\textbf{ProPara} \cite{mishraTrackingStateChanges2018} & 488                    & C                & C     & --           & C          & C                    \\\midrule
\textbf{MultiRC} \cite{KCRUR18} & 9.87K                    & T, C                & C     & C           & --          & C                    \\\midrule
\textbf{ARCT} \cite{habernalArgumentReasoningComprehension2018} & 2.45K & T,C & C & C & C & C \\\midrule
\textbf{SQuAD 2.0} \cite{rajpurkarKnowWhatYou2018} & 151K                      & T, C                & C      & --           & --         & C                    \\\midrule
\textbf{CoQA} \cite{reddyCoQAConversationalQuestion2018} & 8.40K & T, C & C & -- & C & C \\\midrule
\textbf{QuAC} \cite{choiQuACQuestionAnswering2018} & 98.4K & T, C & C & -- & C & C \\\midrule
\textbf{GLUE} \cite{wangGLUEMultiTaskBenchmark2018} & 1.44M & T, A & T, A & -- & M & -- \\ \midrule
\textbf{DNC} \cite{poliak2018emnlp-DNC} & 570K & M, A, T & M, A, T & -- & -- & C \\\midrule  
\textbf{OpenBookQA} \cite{mihaylovCanSuitArmor2018} & 5.96K & C & C & C & C & C \\\midrule 
\textbf{SWAG} \cite{zellersSWAGLargeScaleAdversarial2018} & 114K         & T                & T, A    & A            & C          & C  \\ \midrule 
\textbf{ReCoRD} \cite{zhangReCoRDBridgingGap2018} & 121K         & A, T                & A    & A            & A          & A, C  \\ \midrule
\textbf{CommonsenseQA} \cite{talmorCommonsenseQAQuestionAnswering2019} & 9.40K & C & T & T & -- & C \\ \midrule 
\textbf{SocialIQA} \cite{sapSocialIQACommonsenseReasoning2019} & 44.8K & T, C & C & A, C & -- & C \\ \midrule
\textbf{DREAM} \cite{sunDREAMChallengeDataset2019} & 10.2K & T & T & T & -- & -- \\ \midrule
\textbf{DROP} \cite{duaDROPReadingComprehension2019} & 96.6K & T, C & C & -- & -- & C \\ \midrule
\textbf{SuperGLUE} \cite{wangSuperGLUEStickierBenchmark2019} & 21.3K & T, A & T, A & -- & C & -- \\ \midrule
\textbf{HellaSWAG} \cite{zellers2019hellaswag} & 70.0K         & T                & T, A    & A            & C          & C  \\ \midrule
\textbf{WinoGrande} \cite{sakaguchi2019winogrande} & 44.0K & C, A & C & -- & -- & C, A \\ \midrule
\textbf{SherLIiC} \cite{schmitt-schutze-2019-sherliic} & 3.99K & T, C & C & -- & A & C \\ \midrule
\textbf{AlphaNLI} \cite{ch2019abductive} & 171K & T & C, A & C, A & -- & A \\ \midrule
\textbf{Cosmos QA} \cite{huang2019cosmos} & 35.6K & T, C & C & C & -- & C \\ \midrule 
\textbf{MC-TACO} \cite{zhou2019going} & 1.89K & T, C & C & C & -- & C  \\\bottomrule \normalsize 
\end{tabular}
\label{tbl:benchmark collection}
\end{table}